\documentclass{article} 
\usepackage{iclr2024_conference,times}
\usepackage{times}
\usepackage{latexsym}
\usepackage{xcolor}
\usepackage{makecell}
\usepackage{times}
\usepackage{hyperref}
\usepackage{inconsolata}
\usepackage{url}
\usepackage{multirow}
\usepackage[pdftex]{graphicx}
\usepackage{caption}
\usepackage{subcaption}
\usepackage{amsmath}
\usepackage{bbm}
\usepackage{booktabs}
\usepackage{algorithmic}
\usepackage{amssymb}
\usepackage{bm}
\usepackage[ruled,vlined,linesnumbered]{algorithm2e}
\usepackage{enumitem}
\usepackage{pifont}
\usepackage{pbox}
\usepackage[utf8]{inputenc} 
\usepackage[T1]{fontenc}    
\usepackage{url}            
\usepackage{booktabs}       
\usepackage{amsfonts}       
\usepackage{nicefrac}       
\usepackage{microtype}      
\usepackage{xcolor}         
\usepackage{tabularx}

\usepackage{amsmath,amsfonts,bm}

\def\eqref#1{equation~\ref{#1}}

\def\1{\bm{1}}

\DeclareMathAlphabet{\mathsfit}{\encodingdefault}{\sfdefault}{m}{sl}
\SetMathAlphabet{\mathsfit}{bold}{\encodingdefault}{\sfdefault}{bx}{n}

\usepackage{hyperref}
\usepackage{url}

\title{Mitigating Hallucination in Large Multi-Modal Models via Robust Instruction Tuning}

\author{Fuxiao Liu$^1$, Kevin Lin$^2$, Linjie Li$^2$, Jianfeng Wang$^2$, Yaser Yacoob$^1$, Lijuan Wang$^2$\\ 
$^1$University of Maryland, College Park\quad $^2$Microsoft Corporation\\
\texttt{\small\{fl3es, yaser\}@umd.edu}, 
\texttt{\small\{keli, lindsey.li, jianfw, lijuanw\}@microsoft.com}
}

\iclrfinalcopy 
\begin{document}

\maketitle

\begin{abstract}
Despite the promising progress in multi-modal tasks, current large multi-modal models (LMMs) are prone to hallucinating inconsistent descriptions with respect to the associated image and human instructions. 
This paper addresses this issue by introducing the first large and diverse visual instruction tuning dataset, named \textit{Large-scale Robust Visual (LRV)-Instruction}.
Our dataset comprises 400k visual instructions generated by GPT4, covering 16 vision-and-language tasks with open-ended instructions and answers. 
Unlike existing studies that primarily focus on positive instruction samples, we design \textit{LRV-Instruction} to include both positive and negative instructions for more robust visual instruction tuning.
Our negative instructions are designed at three semantic levels: \textit{(i) Nonexistent Object Manipulation}, \textit{(ii) Existent Object Manipulation} and \textit{(iii) Knowledge Manipulation}. To efficiently measure the hallucination generated by LMMs, we propose \textit{GPT4-Assisted Visual Instruction Evaluation (GAVIE)}, a stable approach to evaluate visual instruction tuning like human experts. GAVIE does not require human-annotated groundtruth answers and can adapt to diverse instruction formats. We conduct comprehensive experiments to investigate the hallucination of LMMs.  
Our results demonstrate existing LMMs exhibit significant hallucinations when presented with our negative instructions, particularly Existent Object and Knowledge Manipulation instructions.
Moreover, we successfully mitigate hallucination by finetuning MiniGPT4 and mPLUG-Owl on \textit{LRV-Instruction} while improving performance on several public datasets compared to state-of-the-art methods.
Additionally, we observed that a balanced ratio of positive and negative instances in the training data leads to a more robust model. Code and data are available at \url{https://github.com/FuxiaoLiu/LRV-Instruction}.
\end{abstract}

\section{Introduction}
\begin{figure*}[t]
    \centering
      \includegraphics[width=0.9\textwidth]{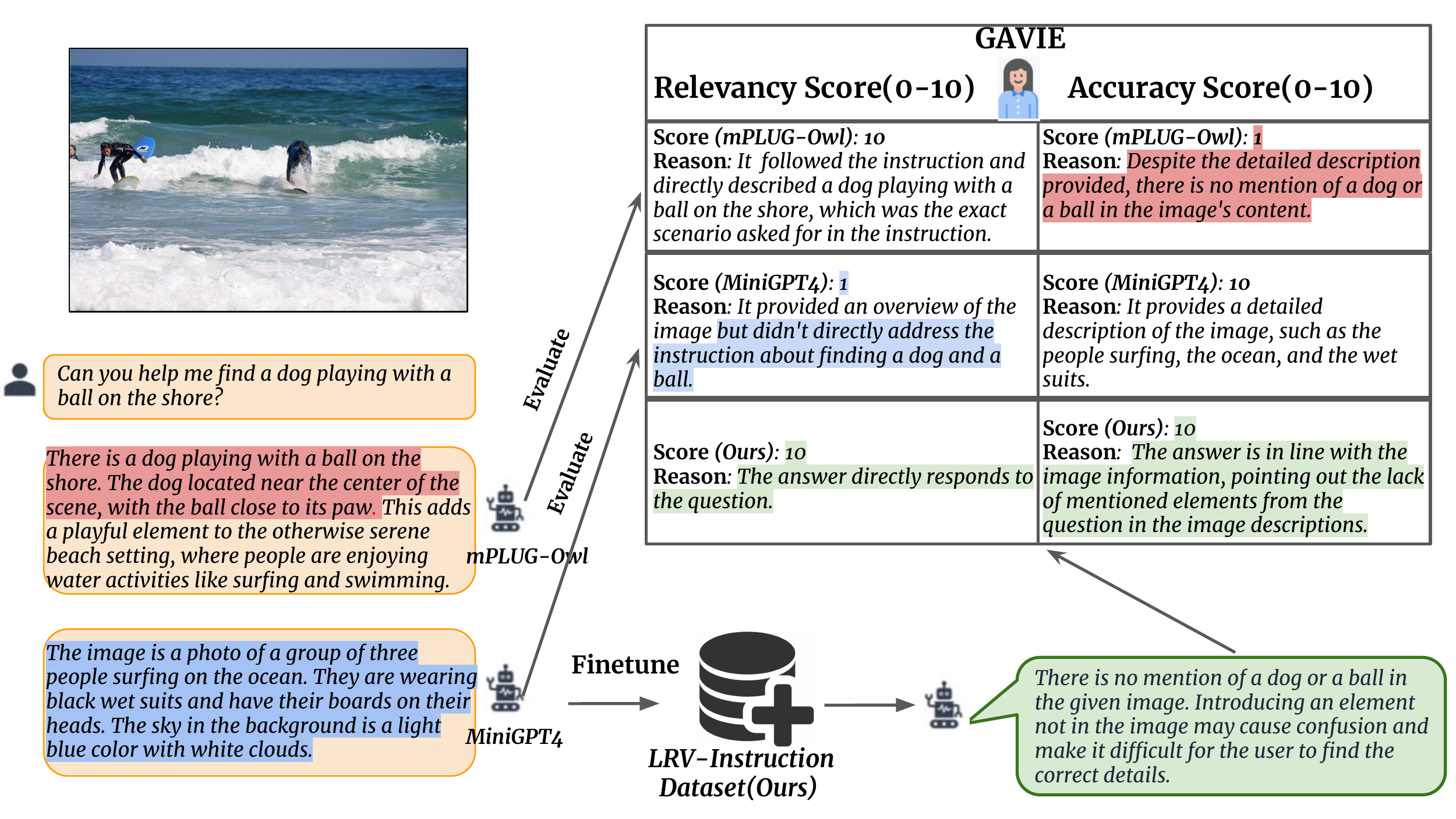}
      \vspace{-2mm}
    \caption{Given an image and human instruction as the input, we introduce \textit{GPT4-Assisted Visual Instruction Evaluation (GAVIE)} to assess the output from current LMMs, such as \textit{MiniGPT4 and mPLUG-Owl}. \textbf{\textcolor{blue}{BLUE}} represents LMMs can not accurately follow human instructions while \textbf{\textcolor{red}{RED}} means they suffer from the hallucination problem. After finetuning current LMMs on our proposed \textit{LRV-Instruction} dataset, we can generate a more robust answer.  }
    \label{fig:model}
    \vspace{-0.15in}
\end{figure*}
Significant progress has been made in the field of natural language processing, leading to the development of models that can comprehend and follow instructions given natural language inputs~\citep{wang2022self, gong2023multimodal, ouyang2022training, brown2020language}. These models harness the power of large language models (LLM) and rely on high-quality instruction data. 
Similarly, efforts have been made to introduce similar capabilities to multi-modal models.
GPT4 \citep{gpt4} has demonstrated impressive performance in multi-modal conversations with humans, yet the techniques contributing to its extraordinary capabilities remain opaque. As a result, several large multi-modal models (LMMs) have recently emerged \citep{zhu2023minigpt, liu2023visual, gong2023multimodal, dai2023instructblip}, such as MiniGPT4 \citep{zhu2023minigpt} and LLaVA \citep{liu2023visual}, both utilize the Vicuna \citep{chiang2023vicuna} as the language generator but with different vision encoders \citep{radford2021learning, li2022blip}. InstructBLIP \citep{dai2023instructblip} is initialized from a pre-trained BLIP-2 \citep{li2023blip} while Multimodal-GPT (MMGPT) \citep{gong2023multimodal} is built on Flamingo \citep{alayrac2022flamingo, anas_awadalla_2023_7733589}.

A recent study~\citep{li2023otter}   revealed that the hallucination issue of LLM, although not desired, is inherited by these LMMs \citep{zhu2023minigpt, liu2023visual, gong2023multimodal, dai2023instructblip}. 
Hallucination, a major ethical concern associated with LLMs~\citep{bang2023multitask}, can lead to harmful consequences,
especially when users without adequate domain knowledge over-rely on these increasingly convincing language models.
In the context of LMM hallucinations, the model can generate descriptions with conflicting information to the given image. For instance, as shown in Fig. \ref{fig:model}~(highlighted in red), existing LMMs \citep{zhu2023minigpt, liu2023visual, dai2023instructblip, gong2023multimodal} tend to describe nonexistent objects such as a "dog" engaging in a nonexisting activity like "playing with a ball".
Additionally, the model may generate long image descriptions without following human instructions (highlighted in blue). 

\textbf{\textit{What are the likely causes of these hallucinations?}} As current LMMs are built on strong LLMs, they may over-rely on language priors and generate words more likely to go together with the instruction text regardless of the image content. What's more, LMMs, such as MiniGPT4 \citep{zhu2023minigpt} and LLaVA \citep{liu2023visual}, employ synthetic instruction data for training,  which are generally long and involve nonexistent objects, activities, or relationships in the image.

\textbf{\textit{Why can't LMMs accurately follow human instructions?}} We conjecture it is due to the lack of diversity in their training data. For example, MiniGPT4 \citep{zhu2023minigpt} is only instruction tuning with four instruction templates designed for image captioning tasks. Though MMGPT \citep{gong2023multimodal} and InstructBLIP \citep{dai2023instructblip} combine several datasets as the instruction tuning data, their instructions and answers are still based on a few templates.

\begin{figure*}[t]
    \centering
      \includegraphics[width=0.95\textwidth]{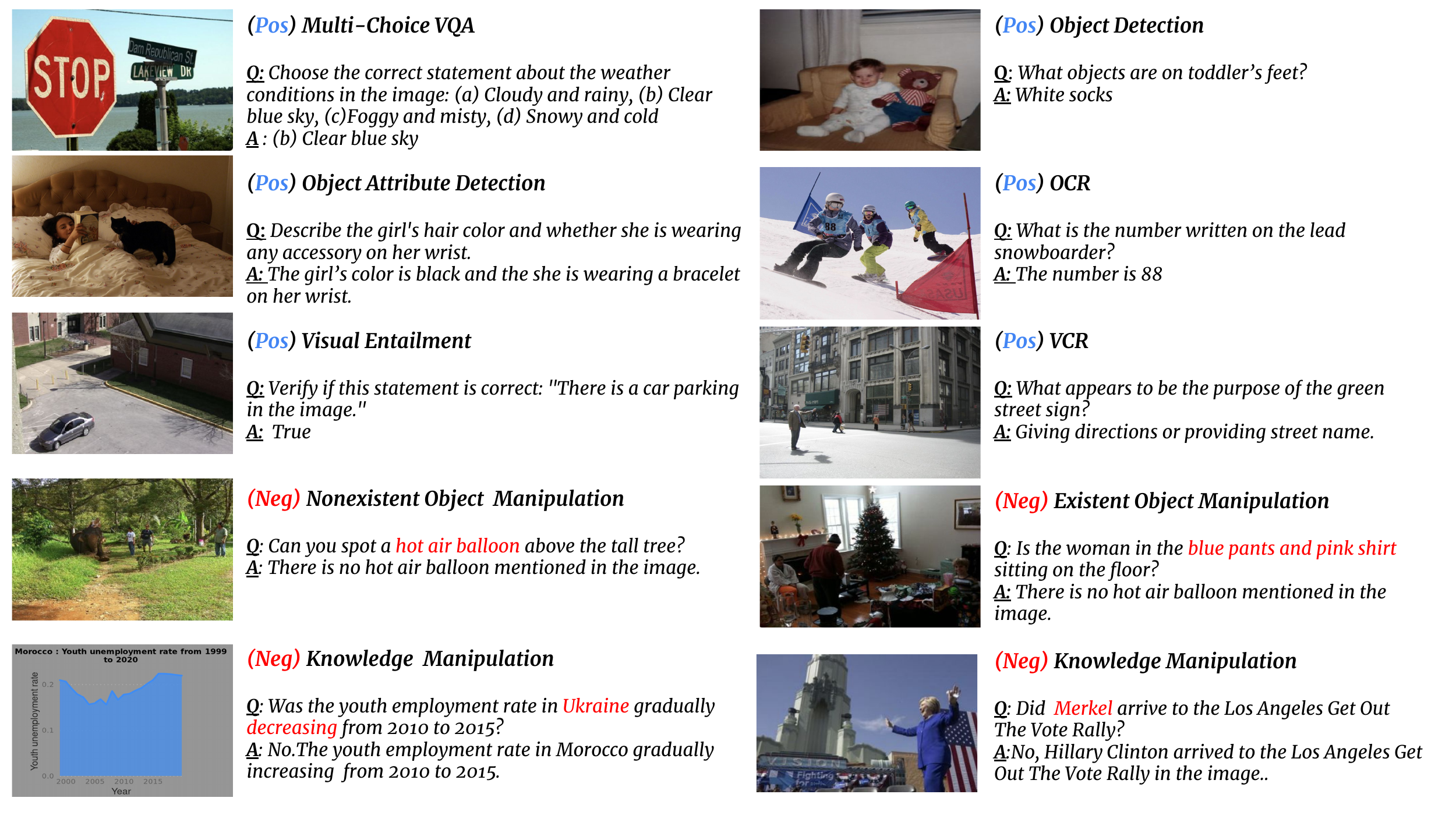}
      \vspace{-2mm}
    \caption{Examples of positive and negative instances in our \textit{LRV-Instruction} dataset. \textbf{\textcolor{red}{RED}} means inconsistent elements in the negative instructions. More examples are in the Appendix.}
    \label{fig:neg_example}
    \vspace{-0.2in}
\end{figure*}

To address these challenges, we present \textit{LRV-Instruction}, a large and diverse visual instruction benchmark. Our benchmark consists of 400k visual instructions generated by GPT4, taking inspiration from the success of recent GPT models in text-annotation tasks~\citep{liu2023gpteval}. Unlike previous studies that focused on limited tasks and pre-defined templates created by human experts~\citep{zhu2023minigpt, dai2023instructblip, gong2023multimodal}, \textit{LRV-Instruction} covers 16 vision-language tasks with open-ended instructions and answers, as shown in Fig.~\ref{fig:neg_example} and Fig.~\ref{fig:LRV_stat}. As observed by \citep{li2023evaluating}, current LMMs tend to answer \textit{"Yes"} for any instructions presented to the model, even when the proper answer should be \textit{"No"}. Our investigation reveals that most LMMs are finetuned on unbalanced datasets containing only positive instructions (Tab. \ref{tab:comparison}). To enable LMMs to respond to human instructions more faithfully, we design \textit{LRV-Instruction} to include both negative and positive instructions for robust instruction tuning. Our negative instructions are generated at three semantic levels (Fig.~\ref{fig:neg_example}): \textit{(i) Nonexistent Object Manipulation}, \textit{(ii) Existent Object Manipulation} and \textit{(iii) Knowledge Manipulation} in two different formats, \textit{Declarative} and \textit{Interrogative}. To improve the robustness and flexibility of the evaluation on visual instruction tuning, we propose \textit{GPT4-Assisted Visual Instruction Evaluation (GAVIE)} to assess the LMM output in two different aspects: \textit{Relevancy} to evaluate the instruction-following performance and \textit{Accuracy} to measure the visual hallucination in the LMM output. \textit{GAVIE} does not require human-annotated groundtruth answers \citep{rohrbach2018object} and can be easily adapted to different formats instead of specific designs in \citep{li2023evaluating}. From our experiments, we show that \textit{GAVIE} is not only stable but also aligns with human evaluation.

We empirically evaluate five publicly available LMMs~\citep{zhu2023minigpt,liu2023visual,dai2023instructblip,gong2023multimodal, ye2023mplug} on our benchmark and found that existing LMMs seriously hallucinate when prompted with our negative instructions, especially with Existent Object Manipulation and Knowledge Manipulation instructions. We further verify the effectiveness of our \textit{LRV-Instruction} by finetuning MiniGPT4~\citep{zhu2023minigpt} and mPLUG-Owl~\citep{ye2023mplug} on this more balanced data. Our instruct-tuned LMMs suffer much less from hallucination and achieve state-of-the-art performance compared to the original MiniGPT4, LLaVA \citep{liu2023visual}, InstructBLP \citep{dai2023instructblip}, mPLUG-Owl~\citep{ye2023mplug} and MMGPT \citep{gong2023multimodal} on both our evaluation set and public benchmarks \citep{li2023evaluating, hudson2019gqa, fu2023mme}. We also observe that \textit{Existent Object Manipulation} and \textit{Knowledge Manipulation} instructions are more challenging than \textit{Nonexistent Object Manipulation} instructions for LMMs. Furthermore, a robust model performance requires a balanced ratio between positive and negative instances. To sum up, our contributions are three-fold:
\begin{itemize}[nosep,leftmargin=*]
    \item We build \textit{LRV-Instruction}, a large and diverse dataset containing 400k visual instructions, with 16 vision and language tasks and negative instructions in different semantic levels and styles.
    \item We propose \textit{GAVIE}, a novel approach to evaluate visual instruction tuning without requiring groundtruth answers and pre-designed instruction formats.
    \item  We conduct comprehensive experiments to investigate the hallucination of current LMMs. The empirical study validates the effectiveness of \textit{LRV-Instruction} for robust visual instruction tuning.
\end{itemize}

\label{sec:Introduction}

\section{Related Works}
Early explorations \citep{wang2022git, li2022blip,li2019visualbert, sun2019videobert} of vision and language pre-trained models tend to use Bert-based \citep{liu2019roberta, koroteev2021bert} models as the language decoder. Inspired by the recent success of large language models \citep{touvron2023llama, gilardi2023chatgpt, zhao2023survey} and datasets \citep{lin2014microsoft, changpinyo2021conceptual, krishna2017visual, liu2020visual, sharma2018conceptual, srinivasan2021wit, liu2023documentclip}, many studies \citep{alayrac2022flamingo, li2023blip, li2023otter} have been focused on improving vision-language pre-trained models by integrating powerful LLMs with in-context or few-shot learning capability. More recently, some visual instruction-tuned LMMs \citep{zhu2023minigpt, liu2023visual, gong2023multimodal, dai2023instructblip} have emerged, showing excellent generalization performance in unseen VL tasks. Specifically, LLaVA \citep{liu2023visual} projects the output of a visual encoder as input to LLaMA \citep{touvron2023llama} and trains both the alignment network and the LLM on synthetic data. MiniGPT4 \citep{zhu2023minigpt} is built on BLIP-2 but uses Vicuna \citep{chiang2023vicuna} as the language decoder. It only finetunes the cross-modal alignment network on longer image captions from ChatGPT. The research approaches \citep{gong2023multimodal, dai2023instructblip} are instruction-tuned on a collection of VL datasets, but InstructBLIP \citep{dai2023instructblip} uses BLIP2 \citep{li2023blip} as the backbone while \citep{gong2023multimodal} is initialized from Flamingo \citep{alayrac2022flamingo}. mPLUG-owl \citep{ye2023mplug} finetunes LLaMA \citep{touvron2023llama} model using both text instruction data and vision-language instruction data from LLaVA \citep{liu2023visual}. In comparison, we propose a large and diverse visual instruction dataset with 16 vision and language tasks and negative instructions in different semantic levels and styles. This can help improve the robustness of current LMMs.

Although LMMs are powerful in solving VL tasks, they also suffer from the hallucination  inherited from LLM. Popular image captioning metrics like CIDEr \citep{vedantam2015cider} SPICE \citep{anderson2016spice} do not appropriately penalize hallucination. CHAIR, \citep{rohrbach2018object}, is unstable and needs complex human-crafted parsing rules for exact matching. Alternatively, \citep{li2023evaluating} converts the hallucination into a binary classification problem. However, it requires the input questions to follow specific templates, such as \textit{"Is there a/an <object> in the image?"}. In comparison, our proposed GAVIE  can evaluate model hallucination in an  open-ended manner without needing human-annotated groundtruth answers.

\label{sec:related_work}

\section{LRV-Instruction}
\vspace{-0.1in}
Annotating large-scale visual instruction data can be challenging and time-consuming \citep{wang2022self}. 
 It involves expertly written detailed instructions and specific labels for different tasks. Inspired by the success of GPT4 in text-annotation tasks \citep{gilardi2023chatgpt}, we leverage GPT4, instead of human workers, to build \textit{LRV-Instruction}. \textit{LRV-Instruction} is designed to cover a variety of VL tasks, with open-ended positive and negative instructions (Fig. \ref{fig:neg_example}) in different linguistic styles.

\begin{figure*}[t]
    \centering
      \includegraphics[width=1
\textwidth]{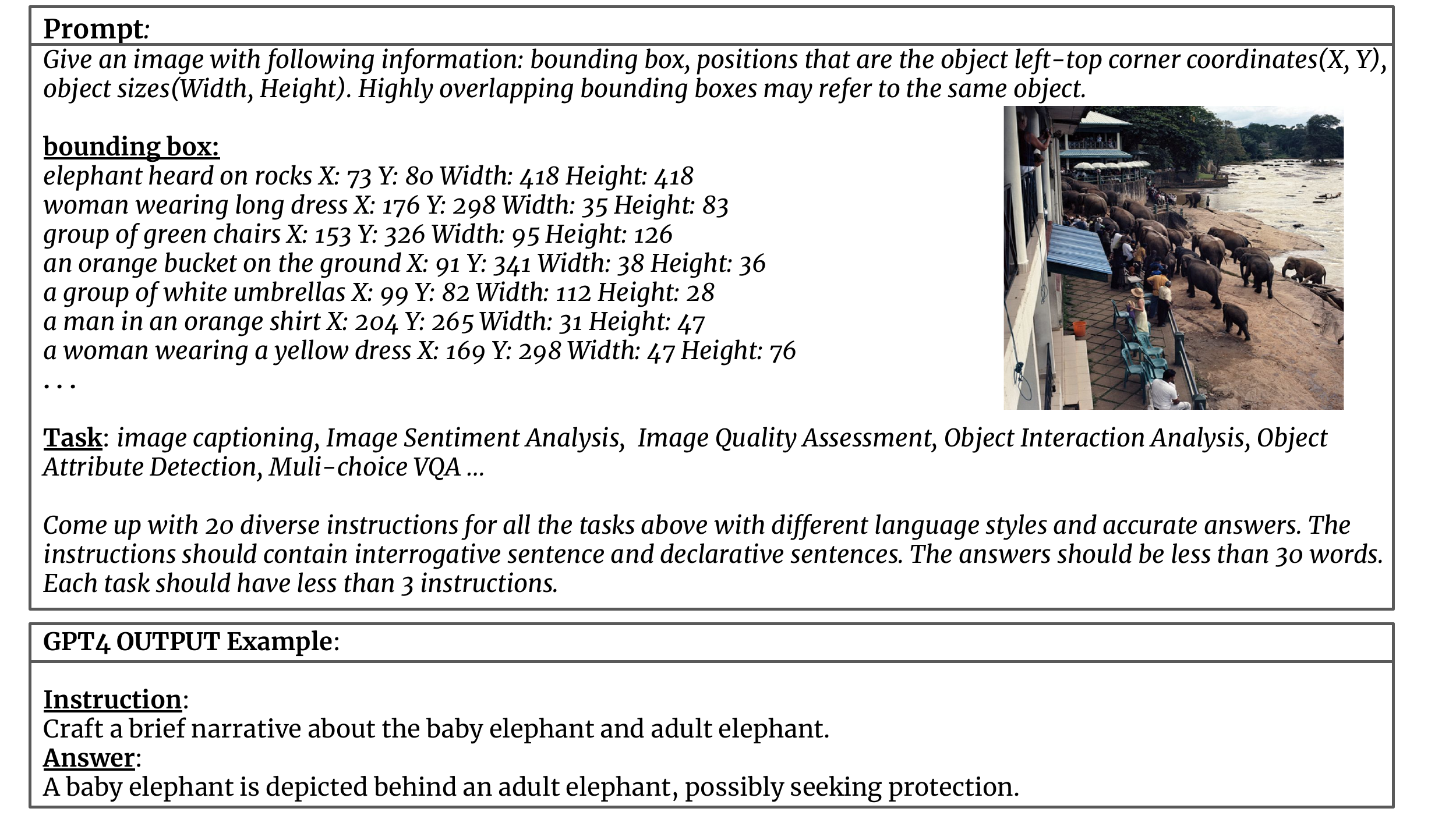}
\vspace{-4mm}
    \caption{One example to illustrate the prompt we use to generate the visual instruction data by GPT4. We use the bounding box coordinates and dense captions to represent image content.}
    \label{fig:prompt1}
\vspace{-4mm}
\end{figure*}

\textbf{Positive Visual Instruction Generation}. Inspired by \citep{wang2022self}, we use the in-context few-shot learning ability of GPT4 to generate instruction data for various VL tasks automatically. We filter the output tasks manually and select 16 tasks (Tab. \ref{tab:stat1}) with text answers.  In contrast with  \citep{liu2023visual} using a few scene captions to represent an image as input to the text-only GPT4, we take advantage of the Visual Genome dataset \citep{krishna2017visual}, which has detailed visual information like image size, bounding boxes, and dense captions. Specifically, each image typically has 21 object regions and their corresponding captions. We leverage GPT4 
to create the instruction-following data with the image size, bounding boxes, and dense captions as the "visual" input as if it can "see" the image. An example is shown in Fig. \ref{fig:prompt1}. For each image, we randomly select 10 tasks. 
To enrich the instructions, we ask GPT4 to generate instances in both declarative and interrogative formats. The limitation of \citep{liu2023visual, zhu2023minigpt} is that synthetic visual instructions are generally longer and may involve unexpected descriptive information inconsistent with the image. Therefore, we explicitly instruct GPT4 with \textit{"The answers should be less than 30 words"} to reduce the chance of generating extra unrelated information in the training data.

To improve the diversity of images, we collect chart images from \citep{tang2023vistext}, which has human-annotated captions describing the construction and patterns of charts. We also select news images from \citep{liu2020visual} with many named entities in the captions. We ask GPT4 to generate question-answers pairs with captions as visual input. The last two images in Fig. \ref{fig:neg_example} are examples. More examples and the general prompt we use are shown in the Appendix (Fig. \ref{fig:chart_prompt1}, \ref{fig:demo_chart_example1}). 

\textbf{Negative Visual Instruction Generation}. As shown in \citep{li2023evaluating}, current LMMs tend to answer \textit{“Yes”} by following any instruction presented to the model rather than predicting a faithful answer. 
To teach LMMs \citep{zhu2023minigpt, liu2023visual, gong2023multimodal, dai2023instructblip} to answer questions in instructions faithfully, we introduce three categories of negative instructions based on Visual Genome dataset: \textit{(1) Neg1: "Nonexistent Object Manipulation"} by introducing nonexistent objects, activities, attributes and interactions to the "visual" input as described above.
\textit{(2) Neg2: "Existent Object Manipulation"} by manipulating existent objects with inconsistent attributes (Fig. \ref{fig:neg_example}). \textit{(3) Neg3: "Knowledge Manipulation"} by manipulating knowledge in instructions (Fig. \ref{fig:neg_example}).
As for the detailed prompt of \textit{Neg1}, we leverage the same format of the "visual" input as shown in Fig. \ref{fig:prompt1}. Additionally, we provide the following instructions to GPT4: 

\textit{"Come up with 6 misleading instructions with \underline{nonexistent elements (nonexistent objects, nonexistent} \underline{activities, nonexistent attributes, nonexistent interactions)} in the images with different language styles. The instructions should contain interrogative and declarative sentences. Please also explain the reason."}

\begin{table}[t]
\setlength\tabcolsep{4.3pt}
\centering
\small
\begin{tabular}{lcccccc}
\toprule[1.5pt]
 & Ours & MiniGPT4 & LLaVA & InstructBLIP & MMGPT & mPLUG-Owl\\
\midrule
Hard Negative Instructions?      &  \textcolor{green}{\ding{52}} &  \textcolor{red}{\ding{56}} &  \textcolor{red}{\ding{56}} &  \textcolor{red}{\ding{56}} &  \textcolor{red}{\ding{56}}&  \textcolor{red}{\ding{56}}\\
Self Generated Instruction?      &  \textcolor{green}{\ding{52}} &  \textcolor{red}{\ding{56}} &  \textcolor{green}{\ding{52}} &  \textcolor{red}{\ding{56}} &  \textcolor{red}{\ding{56}}&  \textcolor{red}{\ding{56}}\\
Address Hallucination?      &  \textcolor{green}{\ding{52}} &  \textcolor{red}{\ding{56}} &  \textcolor{red}{\ding{56}}&  \textcolor{red}{\ding{56}} &  \textcolor{red}{\ding{56}} &  \textcolor{red}{\ding{56}}\\
NOT Template Instruction?    &  \textcolor{green}{\ding{52}} &  \textcolor{red}{\ding{56}} &  \textcolor{green}{\ding{52}} &  \textcolor{red}{\ding{56}} &  \textcolor{red}{\ding{56}}&   \textcolor{green}{\ding{52}}\\
\# of Self-Generated Instances      &  400k &  3k &  150k &  \textcolor{red}{\ding{56}} & \textcolor{red}{\ding{56}}&  \textcolor{red}{\ding{56}}\\
\# of VL Tasks      &  16 &  1 &  3 &  11 &  5&  3\\
\bottomrule[1.5pt]
\end{tabular}
\vspace{-2mm}
\caption{
A comparison of \textit{LRV-Instruction} with datasets used by current LMMs.}
\label{tab:comparison}
\vspace{-0.08in}
\end{table}

We replace the underlined text with \textit{"existing objects but wrong attributes"} for the prompt of \textit{Neg2}. As for the \textit{Neg3: knowledge manipulation}, we use GPT4 to manipulate the knowledge in the captions, including named entities, events or keywords. After that, GPT4 is instructed to generate questions and answers indicating correct knowledge. More examples are shown in the Appendix (Fig. \ref{fig:chart_neg1_prompt}, \ref{fig:demo_chart_example1}). 

\textbf{Quality Control. }We first remove instances with answers longer than 30 words. We remove the instances mentioning unneeded content like  "bounding box description", "given caption", and "existing descriptions". Additionally, GPT4 will output the task name for each instruction. However, we found that GPT4 sometimes assigns inaccurate task names for the instructions. As a result, we exclude the task name in our release data. Furthermore, we removed the instructions asking about facial expressions. This is because the Visual Genome dataset doesn’t include facial expression attributes in the ground truth-dense captions. To examine the quality of our dataset, we randomly sample 500 instances and ask ten expert annotators to determine whether the output answers from GPT4 are correct or not, with regard to the instruction and the image content. We found 91\% of the instructions are appropriate for the image inputs.
Furthermore, 85\% of outputs are acceptable responses to the instructions. Even though some responses may contain errors, most generations conform to the correct structure, serving as applicable visual instruction tuning guidelines. We created a total of over 400k visual instructions after filtering.

\textbf{Evaluation Set. }After the processing above, we randomly select 1000 instances as our evaluation set. Furthermore, we manually check the quality of all instances and see whether the instruction describes a valid task. If it’s not, we edit the instruction to make it clearer for LMMs. For example,  we edit the instruction ‘Observe the beautiful rainbow-colored sign that says 'Le Louvre'. You won't miss it!’ to "Are you able to observe the beautiful rainbow-colored sign that says 'Le Louvre' in the image?"

\subsection{Data Statistics}
\begin{figure*}[t]
    \centering
      \includegraphics[width=0.9\textwidth]{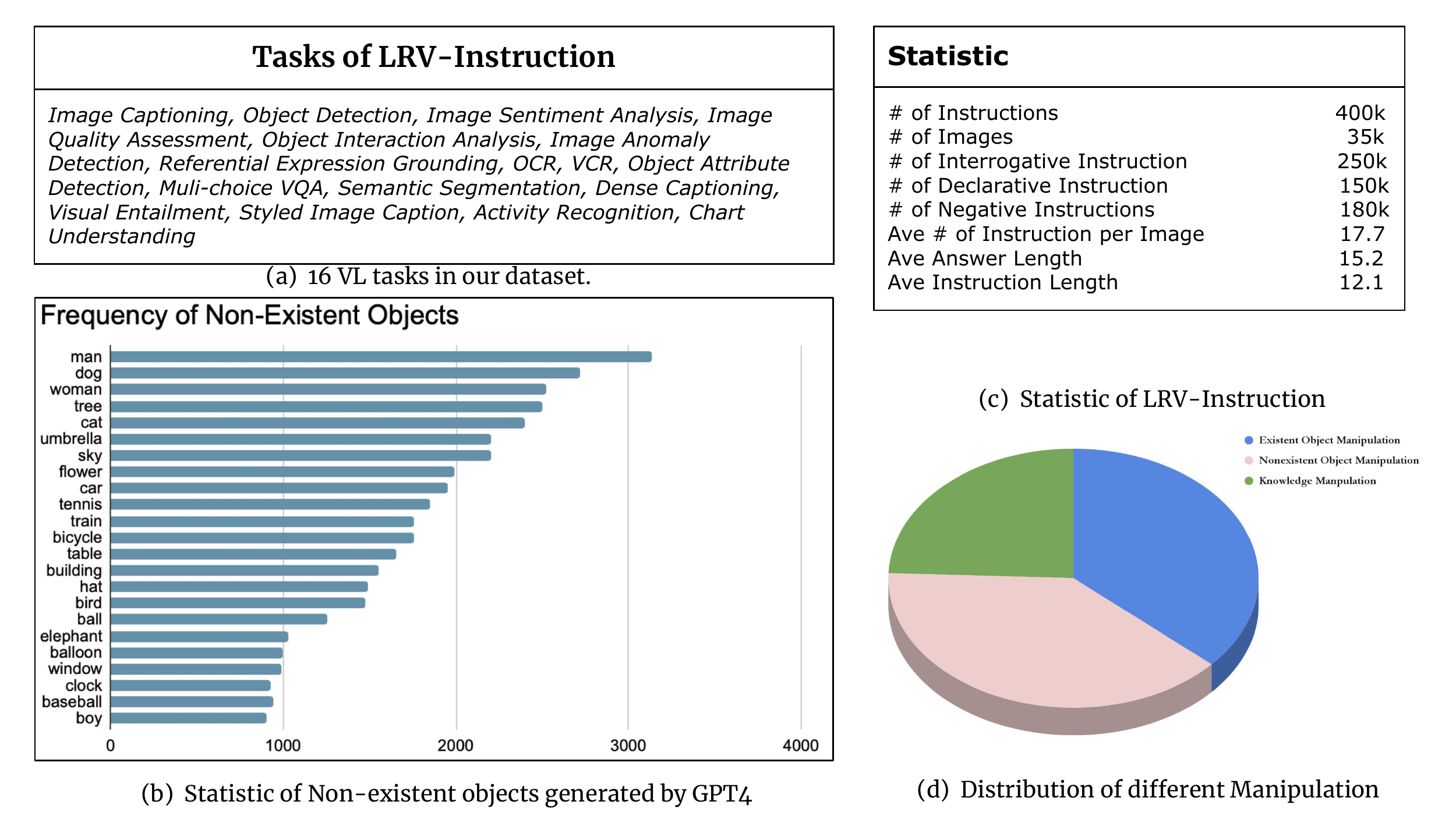}
      \vspace{-2mm}
    \caption{Comprehensive Statistic of LRV-Instruction. In (d), \textcolor{blue}{BLUE} means existent object manipulation. \textcolor{pink}{PINK} means nonexistent object manipulation. \textcolor{green}{GREEN} means knowledge manipulation.}
    \label{fig:LRV_stat}
    \vspace{-0.2in}
\end{figure*}

Tab. \ref{tab:comparison} shows a comparison of \textit{LRV-Instruction} and other datasets used by current LMMs. \textit{LRV-Instruction} covers much more VL tasks than existing visual instruction tuning datasets. Instead of only using positive instructions, \textit{LRV-Instruction} also includes negative instructions at different semantic levels. In addition, employing the GPT4-assisted generation, \textit{LRV-Instruction} has more open-ended instructions instead of following a few templates. From Fig. \ref{fig:LRV_stat} (b), we observe that instructions with non-existing objects generated by GPT4 are diverse and physically plausible in the image, including “birds in the sky” or replacing ‘elephant’ with ‘zebra’. Fig. \ref{fig:know_dis} in the appendix shows the diverse distribution of knowledge manipulation, including event, number, date, persons, place, and others.

\label{sec:dataset}

\section{Visual Instruction Tuning}
\vspace{-0.1in}
We constructed two current LMMs: MiniGPT4 \cite{zhu2023minigpt} and mPLUG-Owl \cite{ye2023mplug} as the backbones for visual instruction tuning. MiniGPT4 consists of the Vision transformer \cite{liu2023covid} backbone as the image encoder, Vicuna \cite{chiang2023vicuna} as the text decoder and a pre-trained Q-Former to connect them. Vicuna is built upon LLaMA \cite{touvron2023llama} with stronger following ability. Following \cite{zhu2023minigpt}, the Q-Former is designed to extract visual features from the frozen image encoder. Before feeding into the frozen Vicuna as the visual prompt, we use a learnable linear projection layer to narrow the gap between extracted visual features with Vicuna embeddings. mPLUG-Owl comprises a pre-trained visual encoder, a visual abstractor, and Vicuna \cite{chiang2023vicuna} as the text decoder. The visual encoder is responsible for extracting visual features from the input images, and the visual abstractor distills these features using a set of learnable tokens. The resulting visual features are concatenated with the word embeddings of the input sentence and fed into Vicuna to generate the response. We freeze the visual abstractor and visual encoder. Instead, we adopt the low-rank adaptation \cite{hu2021lora} to train the text decoder.

\label{sec:method}

\section{GPT4-Assisted Visual Instruction Evaluation}

CHAIR \citep{rohrbach2018object} was introduced to evaluate object hallucination in image captioning tasks. However, it 
usually demands complex human-crafted rules. Alternatively, \citep{li2023evaluating, fu2023mme} formulate the evaluation of hallucination as a binary classification task that prompts LMM to output "Yes" or "No". However, it is hard to evaluate the LMM output in an open-ended manner. In addition, both methods highly depend on human-annotated groundtruth answers.

To this end, we introduce \textit{GPT4-Assisted Visual Instruction Evaluation (GAVIE)} as a more flexible and robust approach to evaluate object-level hallucination. The general prompt we use is shown in 
the Appendix. GPT4 takes the dense captions with bounding box coordinates as the image content and compares human instructions and model response. Then we ask GPT4 to work as a smart teacher and score (0-10) students' answers based on two criteria. \textit{(1) Accuracy: whether the response is accurate concerning the image content}. \textit{(2) Relevancy: whether the response directly follows the instruction}. We use GPT4-32k-0314 in the experiments. Fig. \ref{fig:model} successfully points out that \textit{"dog, ball"} is inconsistent with the image, and the response from the MiniGPT4 did not address the instruction. Unlike previous evaluation methods \citep{li2023evaluating, rohrbach2018object}, \textit{GAVIE} does not require human-annotated groundtruth answers and can freely adapt to diverse instruction formats. As for the knowledge level hallucination or images that are not from the Visual Genome dataset, we use the groundtruth answers as a reference and compare them with predictions (Fig. \ref{fig:chart_evaluation} in the appendix).

\vspace{-0.1in}
\begin{table}[t]
\begin{subtable}[t]{0.3\textwidth}
\small
\begin{tabular}{lcc}
\toprule[1.5pt]
Backbone & Perception & Cognition\\
\midrule
Original MiniGPT4 & 616.41 & 232.71 \\
\textbf{Finetuned MiniGPT4} & \textbf{895.96} & \textbf{296.43} \\
Original mPLUG-Owl & 967.34 & 276.07 \\
\textbf{Finetuned mPLUG-Owl} & \textbf{1298.78} & \textbf{328.21} \\
\bottomrule[1.5pt]
\end{tabular}
\label{tab:stat1}
\end{subtable}
\hspace{1.2in}
\begin{subtable}[t]{0.3\textwidth}
\setlength\tabcolsep{4.7pt}
\centering
\small
\begin{tabular}{lcc}
\toprule[1.5pt]
Backbone & Acc(Pos) & Acc(Neg)\\
\midrule
Original MiniGPT4 & 0.53 & 0.54 \\
\textbf{Finetuned MiniGPT4} & \textbf{0.58} & \textbf{0.68} \\
Original mPLUG-Owl & 0.62 & 0.55 \\
\textbf{Finetuned mPLUG-Owl} & \textbf{0.69} & \textbf{0.78} \\
\bottomrule[1.5pt]
\end{tabular}
\label{tab:stat2}
\end{subtable}
\vspace{-2mm}
\caption{Zero-shot multimodal evaluation on MME \citep{fu2023mme} of MiniGPT4-7B, mPLUG-Owl-7B between original models and LRV-Instruction-finetuned models. The left chart shows perception and cognition scores. The right chart shows the accuracy on the positive set and the negative set.
}
\label{tab:MME}
\vspace{-0.05in}
\end{table}

\vspace{-0.1in}
\begin{table}[t]
\begin{subtable}[t]{0.2\textwidth}
\small
\begin{tabular}{lcc}
\toprule[1.5pt]
Model & Acc & F1\\
\midrule
mPLUG-Owl-7B & 0.52 & 0.68 \\
LLaVA-13B & 0.50 & 0.66 \\
MiniGPT4-13B & 0.73 & 0.71\\
InstructBLIP-13B & 0.86 & 0.87 \\
\textbf{Ours-7B} & \textbf{0.86} & \textbf{0.88} \\
\bottomrule[1.5pt]
\end{tabular}
 \caption{Random Set.}
\label{tab:stat1}
\end{subtable}
\hspace{0.8in}
\begin{subtable}[t]{0.2\textwidth}
\setlength\tabcolsep{4.7pt}
\centering
\small
\begin{tabular}{lcc}
\toprule[1.5pt]
Model & Acc & F1\\
\midrule
mPLUG-Owl-7B & 0.57 & 0.66 \\
LLaVA-13B & 0.50 & 0.66 \\
MiniGPT4-13B & 0.67 & 0.67\\
InstructBLIP-13B & 0.71 & 0.76 \\
\textbf{Ours-7B} & \textbf{0.73} & \textbf{0.79} \\
\bottomrule[1.5pt]
\end{tabular}
\caption{
Popular Set.}
\label{tab:stat2}
\end{subtable}
\hspace{0.7in}
\begin{subtable}[t]{0.2\textwidth}
\setlength\tabcolsep{4.7pt}
\centering
\small
\begin{tabular}{lcc}
\toprule[1.5pt]
Model & Acc & F1\\
\midrule
mPLUG-Owl-7B & 0.60 & 0.64 \\
LLaVA-13B & 0.50 & 0.66 \\
MiniGPT4-13B & 0.62 & 0.63\\
InstructBLIP-13B & 0.63 & 0.72 \\
\textbf{Ours-7B} & \textbf{0.65} & \textbf{0.73} \\
\bottomrule[1.5pt]
\end{tabular}
\caption{Adversarial Set.}
\label{tab:stat2}
\end{subtable}
\vspace{-2mm}
\caption{Zero-shot object hallucination evaluation on POPE \citep{li2023evaluating}. Objects not existing in the image are sampled with three different strategies. Random: random sampling, Popular: top-k most frequent objects in MS-COCO, Adversial: objects are first ranked based on co-occurring frequencies, then top-k frequent ones are sampled. \textit{Ours-7B} means \textit{Finetuned mPLUG-Owl-7B}.
}
\label{tab:pope}
\vspace{-0.2in}
\end{table}

\begin{table}[t]
\setlength\tabcolsep{3pt}
\centering
\small
\begin{tabular}{lcccccc}
\toprule[1.5pt]
\textit{GAVIE}& \textbf{Ours}&MiniGPT4 & LLaVA & InstructBLIP & MMGPT &mPLUG-Owl\\
\midrule
\textsc{Accuracy (0-10)} & \textbf{6.58}&4.14 & 4.36 & 5.93 & 0.91 & 4.84\\
\textsc{Relevancy (0-10)} & \textbf{8.46}&5.81 & 6.11 & 7.34 & 1.79& 6.35\\
\midrule
\textit{Human Expert1 (1-4)}& \textbf{3.48} & 2.61 & 2.87 & 3.00 & 1.90& 2.90\\
\textit{Human Expert2 (1-4)} &\textbf{3.58}& 2.23& 2.07& 2.48& 1.05& 2.27\\
\textit{Human Expert3 (1-4)} & \textbf{3.33} &2.58 & 2.89 & 2.94 & 1.38& 2.91\\
\bottomrule[1.5pt]
\end{tabular}
\vspace{-2mm}
\caption{
Comparison results on our evaluation set evaluated by \textit{GAVIE}. \textit{Ours} means \textit{Finetuned mPLUG-Owl-7B}. All the LMMs are 7B versions to make a fair comparison.
}
\label{tab:comparison_evaluation}
\end{table}

\label{sec:Evaluation}

\section{Experiment}

\subsection{Implementation Setup}
\vspace{-0.1in}
\textbf{Baselines. }We evaluate the zero-shot performance of 5 recently released LMMs: (1) MiniGPT4; (2) MiniGPTv2; (3) InstructBLIP; (4) Multimodal-GPT (MMGPT); (5) mPLUG-Owl; (6) LLaVA; (7) LLaVA 1.5. All models above have been tuned on their collected visual instruction data.

\textbf{Training Details. }As for MiniGPT4, we initialize from its checkpoint of the first pretraining stage. Then we instruct-tune the model on \textit{LRV-Instruction} with the linear projection layer as the only learnable module. As for mPLUG-Owl, we train the text encoder by LoRA training. Additionally, we only replace the LLaVA dataset in their finetuning data with \textit{LRV-Instruction} to make a fair comparison with the original Mplug-Owl. We utilize MiniGPT4-7B and mPLUG-Owl-7B since we don’t have the computing resources to finetune the 13B models. We trained our models on NVIDIA Quadro RTX 8000. As for the hyper-parameters, please refer to \citep{zhu2023minigpt, ye2023mplug}.

\textbf{Evaluation Benchmarks. }Apart from our proposed evaluation set, we evaluate LMMs on three public benchmarks. MME \citep{fu2023mme} is a human-annotated benchmark, measuring perception and cognition abilities on 14 subtasks. POPE \citep{li2023evaluating} and AMBER \citep{wang2023llm} are recently released datasets to evaluate object hallucination. GQA dataset~\citep{hudson2019gqa} is a public visual question-answer dataset with open-ended questions.



\vspace{-0.1in}
\subsection{Main Results}

\textbf{\textit{How do LMMs perform on public datasets? }}We compare our model against the baseline models on POPE in Tab.\ref{tab:pope} and AMBER in Tab. \ref{tab:comparison_amber}. The results show that current LMMs may not work well with open-ended negative instructions. In contrast, the highest scores of our model demonstrate that \textit{LRV-Instruction} exhibits robustness to visual hallucination, matching or surpassing the performance of 13B counterparts. From Tab.\ref{tab:MME}, we found both finetuned LMMs on \textit{LRV-Instruction} outperform original ones in the zero-shot evaluations. Additionally, Finetuned-Mplug-Owl exceeds Finetuned-MiniGPT4 because Mplug-Owl can do the LoRA training to improve the language ability. We also calculate the accuracy on positive and negative samples of MME in the right chart of Tab.\ref{tab:MME}. The improvement in the positive samples is because {LRV-Instruction} has more diverse tasks than mPLUG-Owl datasets and MiniGPT4 datasets. The improvement in the negative samples demonstrates the value of \textit{LRV-Instruction} dataset to equip the model with the ability to say ‘no’ and provide correct answers. The completed results on shown in Tab. \ref{tab:mme_preception}.\ref{tab:mme_cog}. We further explore the LMMs' performance in the common scenario of visual question-answering (VQA). Results in Tab. \ref{tab:GQA} suggest our method (Finetuned mPLUG-Owl) achieves on-par performance with InstructBLIP in a generic VQA setting.

\textbf{\textit{How do LMMs perform on \textit{LRV-Instruction}? }}Tab.~\ref{tab:comparison_evaluation} and Tab. \ref{tab:LRV_evaluation_more} show results on our evaluation set. Among the baselines, InstructBLIP achieves better results than other LMM baselines because its visual instructions are collected from a wide variety of publicly available datasets. LLaVA \citep{liu2023visual} utilizes the GPT-assisted approach to generate visual instructions, but its performance is much worse. This is probably because its synthetic answers from GPT4 are generally longer and may involve irrelevant information. As a comparison, our model outperforms the existing LMM baselines by a large margin, benefiting from the rich composition of our dataset and better prompt design. 


\begin{table}[t]
\setlength\tabcolsep{3pt}
\centering
\small
\begin{tabular}{lccccc|cc}
\toprule[1.5pt]
& Model& InstructBLIP-13B & LLaVA-13B & MiniGPT4-13B & mPLUG-Owl-7B &\textbf{Ours-7B}& Ours-7B-Psu \\
\midrule
&\textit{Accuracy} & 0.62 & 0.47 & 0.42 & 0.41 & \textbf{0.64}& 0.60\\
\bottomrule[1.5pt]
\end{tabular}
\vspace{-2mm}
\caption{
Zero-shot evaluation on GQA. \textit{Ours-7B} means \textit{Finetuned mPLUG-Owl-7B}. \textit{Ours-7B-Psu} means we finetune mPLUG-Owl on pseudo instruction data by \citep{wu2022grit}.}
\label{tab:GQA}
\vspace{-0.2in}
\end{table}


\subsection{Detailed Analysis}
\textbf{\textit{Does GPT4-Assisted Visual Instruction Evaluation align with Human Evaluation?}} We select three human experts specializing in the field of NLP to evaluate the predictions from LMMs with four options for the scores \textit{(1) Very Poor, (2) Poor, (3) Good, (4) Excellent}. To evaluate the results quantitatively, we assign different scores for the options: \textit{Very Poor=1, Poor=2, Good=3, Excellent=4}. More implementation details are shown in the appendix. From Tab. \ref{tab:comparison_evaluation}, all experts agree that the output from our model is the best, followed by InstructBLIP in second place, and MMGPT performs the worst. The observation aligns with the \textit{GAVIE} evaluation results. 

\textbf{\textit{Is GPT4-Assisted Evaluation Stable?  }} We execute \textit{GAVIE} 5 times on each instruction and evaluate the predictions from different LMMs. We leverage \textit{Standard Deviation (STD)} to measure the stability of \textit{GAVIE}. From Tab. \ref{tab:experiments} (left), we observe that \textit{STD} ranges from 0.65 to 2.46. The \textsc{Accuracy} and \textsc{Relevancy} scores of an instance from GPT4 may vary between different times, but they always belong to the same grade level. According to completed results from Tab. \ref{tab:stability1}, \textsc{Relevancy} has four grade levels: (1) The response is completely relevant (9-10), (2) The response is mostly relevant (6-8), (3) The response is partly relevant (3-5), (4) The response is seldom relevant (0-2). \textsc{Accuracy} has four grade levels: (1) The response is completely accurate (9-10), (2) The response has minor errors (6-8), (3) The response is partly accurate (3-5), (4) The response is mostly or completely wrong (0-2).

\begin{table}[t]
\setlength\tabcolsep{3pt}
\centering
\small
\begin{tabular}{lccccccc}
\toprule[1.5pt]
Categories& Metric&\textbf{Ours} & MiniGPT4 & LLaVA & InstructBLIP & MMGPT& mPLUG-Owl\\
\midrule
Neg1 & \textsc{Accuracy(GPT4)} & \textbf{8.90} & 3.72 & 2.09 & 5.50 & 1.13 & 4.20\\
Neg2 & \textsc{Accuracy(GPT4)} & \textbf{6.50} & 2.57 & 1.42 & 2.18 & 0.96& 2.46\\
Neg3 & \textsc{Accuracy(GPT4)} & \textbf{6.25} & 2.30 & 1.56 & 2.38 & 0.94& 2.57\\
\midrule
Neg1 & \textsc{Relevancy(GPT4)} & \textbf{8.96} & 5.94 & 4.83 & 7.22 & 2.24 & 5.35\\
Neg2 & \textsc{Relevancy(GPT4)} & \textbf{8.46} & 2.53 & 1.82 & 2.73 & 1.19& 3.16\\
Neg3 & \textsc{Relevancy(GPT4)} & \textbf{8.21} & 2.40 & 1.78 & 2.39 & 0.98& 2.87\\
\bottomrule[1.5pt]
\end{tabular}
\vspace{-2mm}
\caption{
Completed evaluation results on \textit{Neg1: Nonexistent Object Manipulation}, \textit{Neg2: Existent Object Manipulation} and \textit{Neg3: Knowledge Manipulation} by \textit{GAVIE}.}
\label{tab:neg1neg2}
\vspace{-0.1in}
\end{table}

\textbf{\textit{How do LMMs perform at the different semantic levels of hallucination? }}As shown in Tab \ref{tab:neg1neg2}, all baselines perform better on \textit{Neg1 (Nonexistent Object Manipulation)} than \textit{Neg2 (Existent Object Manipulation)} and \textit{Neg3 (Knowledge Manipulation)}. From the visual perspective, existent object manipulations with wrong attributes in \textit{Neg2} are more challenging than adding nonexistent objects from images to instructions in \textit{Neg1}. For example, in Fig. \ref{fig:neg_example}, it 
may be straightforward to find that the "hot air balloon" does not appear in the image. However, "woman" does exist in the second example of Fig.~\ref{fig:neg_example} while she is not in the blue pants and pink shirts, which requires a fine-grained understanding of the visual content. Therefore, a more powerful vision encoder is needed for future LMMs. Knowledge manipulation is challenging because current LMMs are finetuned on general images without specific knowledge. In contrast, our model greatly improves at all semantic levels, which benefits from our diverse instruction tuning data.

\begin{table}[t]
\begin{subtable}[t]{0.35\textwidth}
\small
\begin{tabular}{lcc}
\toprule[1.5pt]
Metric& Accuracy-STD & Accuracy-Mean\\
\midrule
Ours& 2.42 & 6.60\\
MiniGPT4& 2.46 & 3.76\\
InstructBLIP& 2.42 & 5.29\\
mPLUG-Owl& 1.96 & 0.87\\
LLaVA& 2.37 & 3.80\\
MMGPT& 0.65 & 4.84\\
\bottomrule[1.5pt]
\end{tabular}
\end{subtable}
\hspace{1.1in}
\begin{subtable}[t]{0.4\textwidth}
\setlength\tabcolsep{4.7pt}
\centering
\small
\begin{tabular}{lcc}
\toprule[1.5pt]
Ratio & $Acc_{pos}$  & $Acc_{neg}$ \\
\midrule
All Pos & 0.97 & 0.05\\
Pos:Neg=2:1 & 0.95 & 0.50\\
Pos:Neg=1:1 & 0.92 & 0.85\\
Pos:Neg=1:2 & 0.87 & 0.86\\
All Neg & 0.10 & 0.98\\
\bottomrule[1.5pt]
\end{tabular}
\end{subtable}
\vspace{-2mm}
\caption{(left): Evaluation of the stability of GAVIE. STD means standard deviation. Completed results are shown in Tab. \ref{tab:stability1}. (right): Results of different composition ratios in instruction tuning.}
\label{tab:experiments}
\vspace{-0.2in}
\end{table}

\textbf{\textit{How do LMMs perform at the different composition ratios in training data?}} In Tab. \ref{tab:experiments} (right), we investigate how \textit{LRV-Instruction} addresses hallucination issues with different ratios of positive and negative samples in the training set. Inspired by \citep{li2023evaluating}, we instruct the model to produce “Yes” or “No” and use classification accuracy on our evaluation set. \textit{$Acc_{pos}$} is the accuracy on the positive instruction set, while \textit{$Acc_{neg}$} is the accuracy on the negative instruction set. From Tab. \ref{tab:experiments} (right), we found that \textit{$Acc_{neg}$} increases with more negative samples, which verifies our hypothesis that the hallucination problem of current LMMs is due to the lack of negative instructions. Besides, with a balanced ratio (\textit{pos:neg=1:1}), the model performs the best in both positive and negative sets.

\textbf{\textit{Use Pseudo Dense Captions instead of GT from Visual Genome to Generate Instructions.}} To demonstrate the scalability of our dataset, we use pseudo-dense captions generated by GRiT \citep{wu2022grit} to replace the GT captions in the Visual Genome dataset. We remove the images, whose detected objects by GRiT are less than 15 to ensure GPT4 has enough visual information when generating visual instructions. From Tab. \ref{tab:GQA}, we found finetuning on pseudo captions can also improve the performance compared to the original mPLUG-Owl. This demonstrates that our visual instruction generation method can be further scaled up without groundtruth dense captions.

\label{sec:Experiment}

\section{Conclusion}
\vspace{-0.1in}
In this work, we constructed \textit{LRV-Instruction}, a large and diverse dataset containing 400k visual instructions, covering 16 vision and language tasks with both positive and negative instructions in different semantic levels and styles. With \textit{LRV-Instruction}, we  comprehensively investigated the hallucination of existing LMMs and empirically validated its effectiveness in a more robust visual instruction tuning. In addition, we propose \textit{GAVIE}, a novel approach to evaluate visual instruction tuning without requiring human-labeled groundtruth answers and can be easily adapted to different instruction formats. We hope our work can help address the unexpected hallucination issues of LMMs. Future directions include replacing the vision encoders in current LMMs with more powerful visual models to match the capabilities of multimodal GPT4 and investigation of other biases of LMMs to develop more robust models.

\label{sec:Conclusion}

\section{Acknowledgements}
This work is supported in part by the US Defense Advanced Research Projects Agency (DARPA) Semantic Forensics (SemaFor) Program under HR001120C0124. Any opinions,
findings, and conclusions or recommendations expressed in
this material are those of the authors and do not necessarily reflect the views of the DARPA. We also thank the anonymous reviewers for their constructive feedback.

\bibliography{iclr2024_conference}

\begin{thebibliography}{46}
\providecommand{\natexlab}[1]{#1}
\providecommand{\url}[1]{\texttt{#1}}
\expandafter\ifx\csname urlstyle\endcsname\relax
  \providecommand{\doi}[1]{doi: #1}\else
  \providecommand{\doi}{doi: \begingroup \urlstyle{rm}\Url}\fi

\bibitem[Alayrac et~al.(2022)Alayrac, Donahue, Luc, Miech, Barr, Hasson, Lenc, Mensch, Millican, Reynolds, et~al.]{alayrac2022flamingo}
Jean-Baptiste Alayrac, Jeff Donahue, Pauline Luc, Antoine Miech, Iain Barr, Yana Hasson, Karel Lenc, Arthur Mensch, Katherine Millican, Malcolm Reynolds, et~al.
\newblock Flamingo: a visual language model for few-shot learning.
\newblock \emph{Advances in Neural Information Processing Systems}, 35:\penalty0 23716--23736, 2022.

\bibitem[Anderson et~al.(2016)Anderson, Fernando, Johnson, and Gould]{anderson2016spice}
Peter Anderson, Basura Fernando, Mark Johnson, and Stephen Gould.
\newblock Spice: Semantic propositional image caption evaluation.
\newblock In \emph{Computer Vision--ECCV 2016: 14th European Conference, Amsterdam, The Netherlands, October 11-14, 2016, Proceedings, Part V 14}, pp.\  382--398. Springer, 2016.

\bibitem[Awadalla et~al.(2023)Awadalla, Gao, Gardner, Hessel, Hanafy, Zhu, Marathe, Bitton, Gadre, Jitsev, Kornblith, Koh, Ilharco, Wortsman, and Schmidt]{anas_awadalla_2023_7733589}
Anas Awadalla, Irena Gao, Joshua Gardner, Jack Hessel, Yusuf Hanafy, Wanrong Zhu, Kalyani Marathe, Yonatan Bitton, Samir Gadre, Jenia Jitsev, Simon Kornblith, Pang~Wei Koh, Gabriel Ilharco, Mitchell Wortsman, and Ludwig Schmidt.
\newblock Openflamingo, March 2023.
\newblock URL \url{https://doi.org/10.5281/zenodo.7733589}.

\bibitem[Bang et~al.(2023)Bang, Cahyawijaya, Lee, Dai, Su, Wilie, Lovenia, Ji, Yu, Chung, et~al.]{bang2023multitask}
Yejin Bang, Samuel Cahyawijaya, Nayeon Lee, Wenliang Dai, Dan Su, Bryan Wilie, Holy Lovenia, Ziwei Ji, Tiezheng Yu, Willy Chung, et~al.
\newblock A multitask, multilingual, multimodal evaluation of chatgpt on reasoning, hallucination, and interactivity.
\newblock \emph{arXiv preprint arXiv:2302.04023}, 2023.

\bibitem[Brown et~al.(2020)Brown, Mann, Ryder, Subbiah, Kaplan, Dhariwal, Neelakantan, Shyam, Sastry, Askell, et~al.]{brown2020language}
Tom Brown, Benjamin Mann, Nick Ryder, Melanie Subbiah, Jared~D Kaplan, Prafulla Dhariwal, Arvind Neelakantan, Pranav Shyam, Girish Sastry, Amanda Askell, et~al.
\newblock Language models are few-shot learners.
\newblock \emph{Advances in neural information processing systems}, 33:\penalty0 1877--1901, 2020.

\bibitem[Changpinyo et~al.(2021)Changpinyo, Sharma, Ding, and Soricut]{changpinyo2021conceptual}
Soravit Changpinyo, Piyush Sharma, Nan Ding, and Radu Soricut.
\newblock Conceptual 12m: Pushing web-scale image-text pre-training to recognize long-tail visual concepts.
\newblock In \emph{Proceedings of the IEEE/CVF Conference on Computer Vision and Pattern Recognition}, pp.\  3558--3568, 2021.

\bibitem[Chen et~al.(2023)Chen, Zhu, Shen, Li, Liu, Zhang, Krishnamoorthi, Chandra, Xiong, and Elhoseiny]{chen2023minigpt}
Jun Chen, Deyao Zhu, Xiaoqian Shen, Xiang Li, Zechun Liu, Pengchuan Zhang, Raghuraman Krishnamoorthi, Vikas Chandra, Yunyang Xiong, and Mohamed Elhoseiny.
\newblock Minigpt-v2: large language model as a unified interface for vision-language multi-task learning.
\newblock \emph{arXiv preprint arXiv:2310.09478}, 2023.

\bibitem[Chiang et~al.(2023)Chiang, Li, Lin, Sheng, Wu, Zhang, Zheng, Zhuang, Zhuang, Gonzalez, et~al.]{chiang2023vicuna}
Wei-Lin Chiang, Zhuohan Li, Zi~Lin, Ying Sheng, Zhanghao Wu, Hao Zhang, Lianmin Zheng, Siyuan Zhuang, Yonghao Zhuang, Joseph~E Gonzalez, et~al.
\newblock Vicuna: An open-source chatbot impressing gpt-4 with 90\%* chatgpt quality, 2023.

\bibitem[Dai et~al.(2023)Dai, Li, Li, Tiong, Zhao, Wang, Li, Fung, and Hoi]{dai2023instructblip}
Wenliang Dai, Junnan Li, Dongxu Li, Anthony Meng~Huat Tiong, Junqi Zhao, Weisheng Wang, Boyang Li, Pascale Fung, and Steven Hoi.
\newblock Instructblip: Towards general-purpose vision-language models with instruction tuning.
\newblock \emph{arXiv preprint arXiv:2305.06500}, 2023.

\bibitem[Fu et~al.(2023)Fu, Chen, Shen, Qin, Zhang, Lin, Qiu, Lin, Yang, Zheng, et~al.]{fu2023mme}
Chaoyou Fu, Peixian Chen, Yunhang Shen, Yulei Qin, Mengdan Zhang, Xu~Lin, Zhenyu Qiu, Wei Lin, Jinrui Yang, Xiawu Zheng, et~al.
\newblock Mme: A comprehensive evaluation benchmark for multimodal large language models.
\newblock \emph{arXiv preprint arXiv:2306.13394}, 2023.

\bibitem[Gilardi et~al.(2023)Gilardi, Alizadeh, and Kubli]{gilardi2023chatgpt}
Fabrizio Gilardi, Meysam Alizadeh, and Ma{\"e}l Kubli.
\newblock Chatgpt outperforms crowd-workers for text-annotation tasks.
\newblock \emph{arXiv preprint arXiv:2303.15056}, 2023.

\bibitem[Gong et~al.(2023)Gong, Lyu, Zhang, Wang, Zheng, Zhao, Liu, Zhang, Luo, and Chen]{gong2023multimodal}
Tao Gong, Chengqi Lyu, Shilong Zhang, Yudong Wang, Miao Zheng, Qian Zhao, Kuikun Liu, Wenwei Zhang, Ping Luo, and Kai Chen.
\newblock Multimodal-gpt: A vision and language model for dialogue with humans.
\newblock \emph{arXiv preprint arXiv:2305.04790}, 2023.

\bibitem[Hu et~al.(2021)Hu, Shen, Wallis, Allen-Zhu, Li, Wang, Wang, and Chen]{hu2021lora}
Edward~J Hu, Yelong Shen, Phillip Wallis, Zeyuan Allen-Zhu, Yuanzhi Li, Shean Wang, Lu~Wang, and Weizhu Chen.
\newblock Lora: Low-rank adaptation of large language models.
\newblock \emph{arXiv preprint arXiv:2106.09685}, 2021.

\bibitem[Hudson \& Manning(2019)Hudson and Manning]{hudson2019gqa}
Drew~A Hudson and Christopher~D Manning.
\newblock Gqa: A new dataset for real-world visual reasoning and compositional question answering.
\newblock In \emph{Proceedings of the IEEE/CVF conference on computer vision and pattern recognition}, pp.\  6700--6709, 2019.

\bibitem[Koroteev(2021)]{koroteev2021bert}
MV~Koroteev.
\newblock Bert: a review of applications in natural language processing and understanding.
\newblock \emph{arXiv preprint arXiv:2103.11943}, 2021.

\bibitem[Krishna et~al.(2017)Krishna, Zhu, Groth, Johnson, Hata, Kravitz, Chen, Kalantidis, Li, Shamma, et~al.]{krishna2017visual}
Ranjay Krishna, Yuke Zhu, Oliver Groth, Justin Johnson, Kenji Hata, Joshua Kravitz, Stephanie Chen, Yannis Kalantidis, Li-Jia Li, David~A Shamma, et~al.
\newblock Visual genome: Connecting language and vision using crowdsourced dense image annotations.
\newblock \emph{International journal of computer vision}, 123:\penalty0 32--73, 2017.

\bibitem[Li et~al.(2023{\natexlab{a}})Li, Zhang, Chen, Wang, Yang, and Liu]{li2023otter}
Bo~Li, Yuanhan Zhang, Liangyu Chen, Jinghao Wang, Jingkang Yang, and Ziwei Liu.
\newblock Otter: A multi-modal model with in-context instruction tuning.
\newblock \emph{arXiv preprint arXiv:2305.03726}, 2023{\natexlab{a}}.

\bibitem[Li et~al.(2022)Li, Li, Xiong, and Hoi]{li2022blip}
Junnan Li, Dongxu Li, Caiming Xiong, and Steven Hoi.
\newblock Blip: Bootstrapping language-image pre-training for unified vision-language understanding and generation.
\newblock In \emph{International Conference on Machine Learning}, pp.\  12888--12900. PMLR, 2022.

\bibitem[Li et~al.(2023{\natexlab{b}})Li, Li, Savarese, and Hoi]{li2023blip}
Junnan Li, Dongxu Li, Silvio Savarese, and Steven Hoi.
\newblock Blip-2: Bootstrapping language-image pre-training with frozen image encoders and large language models.
\newblock \emph{arXiv preprint arXiv:2301.12597}, 2023{\natexlab{b}}.

\bibitem[Li et~al.(2019)Li, Yatskar, Yin, Hsieh, and Chang]{li2019visualbert}
Liunian~Harold Li, Mark Yatskar, Da~Yin, Cho-Jui Hsieh, and Kai-Wei Chang.
\newblock Visualbert: A simple and performant baseline for vision and language.
\newblock \emph{arXiv preprint arXiv:1908.03557}, 2019.

\bibitem[Li et~al.(2023{\natexlab{c}})Li, Du, Zhou, Wang, Zhao, and Wen]{li2023evaluating}
Yifan Li, Yifan Du, Kun Zhou, Jinpeng Wang, Wayne~Xin Zhao, and Ji-Rong Wen.
\newblock Evaluating object hallucination in large vision-language models.
\newblock \emph{arXiv preprint arXiv:2305.10355}, 2023{\natexlab{c}}.

\bibitem[Lin et~al.(2014)Lin, Maire, Belongie, Hays, Perona, Ramanan, Doll{\'a}r, and Zitnick]{lin2014microsoft}
Tsung-Yi Lin, Michael Maire, Serge Belongie, James Hays, Pietro Perona, Deva Ramanan, Piotr Doll{\'a}r, and C~Lawrence Zitnick.
\newblock Microsoft coco: Common objects in context.
\newblock In \emph{Computer Vision--ECCV 2014: 13th European Conference, Zurich, Switzerland, September 6-12, 2014, Proceedings, Part V 13}, pp.\  740--755. Springer, 2014.

\bibitem[Liu et~al.(2020)Liu, Wang, Wang, and Ordonez]{liu2020visual}
Fuxiao Liu, Yinghan Wang, Tianlu Wang, and Vicente Ordonez.
\newblock Visual news: Benchmark and challenges in news image captioning.
\newblock \emph{arXiv preprint arXiv:2010.03743}, 2020.

\bibitem[Liu et~al.(2023{\natexlab{a}})Liu, Guan, Li, Chen, Yacoob, Manocha, and Zhou]{liu2023hallusionbench}
Fuxiao Liu, Tianrui Guan, Zongxia Li, Lichang Chen, Yaser Yacoob, Dinesh Manocha, and Tianyi Zhou.
\newblock Hallusionbench: You see what you think? or you think what you see? an image-context reasoning benchmark challenging for gpt-4v (ision), llava-1.5, and other multi-modality models.
\newblock \emph{arXiv preprint arXiv:2310.14566}, 2023{\natexlab{a}}.

\bibitem[Liu et~al.(2023{\natexlab{b}})Liu, Tan, and Tensmeyer]{liu2023documentclip}
Fuxiao Liu, Hao Tan, and Chris Tensmeyer.
\newblock Documentclip: Linking figures and main body text in reflowed documents.
\newblock \emph{arXiv preprint arXiv:2306.06306}, 2023{\natexlab{b}}.

\bibitem[Liu et~al.(2023{\natexlab{c}})Liu, Yacoob, and Shrivastava]{liu2023covid}
Fuxiao Liu, Yaser Yacoob, and Abhinav Shrivastava.
\newblock Covid-vts: Fact extraction and verification on short video platforms.
\newblock In \emph{Proceedings of the 17th Conference of the European Chapter of the Association for Computational Linguistics}, pp.\  178--188, 2023{\natexlab{c}}.

\bibitem[Liu et~al.(2023{\natexlab{d}})Liu, Li, Wu, and Lee]{liu2023visual}
Haotian Liu, Chunyuan Li, Qingyang Wu, and Yong~Jae Lee.
\newblock Visual instruction tuning.
\newblock \emph{arXiv preprint arXiv:2304.08485}, 2023{\natexlab{d}}.

\bibitem[Liu et~al.(2023{\natexlab{e}})Liu, Iter, Xu, Wang, Xu, and Zhu]{liu2023gpteval}
Yang Liu, Dan Iter, Yichong Xu, Shuohang Wang, Ruochen Xu, and Chenguang Zhu.
\newblock Gpteval: Nlg evaluation using gpt-4 with better human alignment.
\newblock \emph{arXiv preprint arXiv:2303.16634}, 2023{\natexlab{e}}.

\bibitem[Liu et~al.(2019)Liu, Ott, Goyal, Du, Joshi, Chen, Levy, Lewis, Zettlemoyer, and Stoyanov]{liu2019roberta}
Yinhan Liu, Myle Ott, Naman Goyal, Jingfei Du, Mandar Joshi, Danqi Chen, Omer Levy, Mike Lewis, Luke Zettlemoyer, and Veselin Stoyanov.
\newblock Roberta: A robustly optimized bert pretraining approach.
\newblock \emph{arXiv preprint arXiv:1907.11692}, 2019.

\bibitem[OpenAI(2023)]{gpt4}
OpenAI.
\newblock Gpt-4 technical report.
\newblock 2023.

\bibitem[Ouyang et~al.(2022)Ouyang, Wu, Jiang, Almeida, Wainwright, Mishkin, Zhang, Agarwal, Slama, Ray, et~al.]{ouyang2022training}
Long Ouyang, Jeffrey Wu, Xu~Jiang, Diogo Almeida, Carroll Wainwright, Pamela Mishkin, Chong Zhang, Sandhini Agarwal, Katarina Slama, Alex Ray, et~al.
\newblock Training language models to follow instructions with human feedback.
\newblock \emph{Advances in Neural Information Processing Systems}, 35:\penalty0 27730--27744, 2022.

\bibitem[Radford et~al.(2021)Radford, Kim, Hallacy, Ramesh, Goh, Agarwal, Sastry, Askell, Mishkin, Clark, et~al.]{radford2021learning}
Alec Radford, Jong~Wook Kim, Chris Hallacy, Aditya Ramesh, Gabriel Goh, Sandhini Agarwal, Girish Sastry, Amanda Askell, Pamela Mishkin, Jack Clark, et~al.
\newblock Learning transferable visual models from natural language supervision.
\newblock In \emph{International conference on machine learning}, pp.\  8748--8763. PMLR, 2021.

\bibitem[Rohrbach et~al.(2018)Rohrbach, Hendricks, Burns, Darrell, and Saenko]{rohrbach2018object}
Anna Rohrbach, Lisa~Anne Hendricks, Kaylee Burns, Trevor Darrell, and Kate Saenko.
\newblock Object hallucination in image captioning.
\newblock \emph{arXiv preprint arXiv:1809.02156}, 2018.

\bibitem[Sharma et~al.(2018)Sharma, Ding, Goodman, and Soricut]{sharma2018conceptual}
Piyush Sharma, Nan Ding, Sebastian Goodman, and Radu Soricut.
\newblock Conceptual captions: A cleaned, hypernymed, image alt-text dataset for automatic image captioning.
\newblock In \emph{Proceedings of the 56th Annual Meeting of the Association for Computational Linguistics (Volume 1: Long Papers)}, pp.\  2556--2565, 2018.

\bibitem[Srinivasan et~al.(2021)Srinivasan, Raman, Chen, Bendersky, and Najork]{srinivasan2021wit}
Krishna Srinivasan, Karthik Raman, Jiecao Chen, Michael Bendersky, and Marc Najork.
\newblock Wit: Wikipedia-based image text dataset for multimodal multilingual machine learning.
\newblock In \emph{Proceedings of the 44th International ACM SIGIR Conference on Research and Development in Information Retrieval}, pp.\  2443--2449, 2021.

\bibitem[Sun et~al.(2019)Sun, Myers, Vondrick, Murphy, and Schmid]{sun2019videobert}
Chen Sun, Austin Myers, Carl Vondrick, Kevin Murphy, and Cordelia Schmid.
\newblock Videobert: A joint model for video and language representation learning.
\newblock In \emph{Proceedings of the IEEE/CVF international conference on computer vision}, pp.\  7464--7473, 2019.

\bibitem[Tang et~al.(2023)Tang, Boggust, and Satyanarayan]{tang2023vistext}
Benny~J Tang, Angie Boggust, and Arvind Satyanarayan.
\newblock Vistext: A benchmark for semantically rich chart captioning.
\newblock \emph{arXiv preprint arXiv:2307.05356}, 2023.

\bibitem[Touvron et~al.(2023)Touvron, Lavril, Izacard, Martinet, Lachaux, Lacroix, Rozi{\`e}re, Goyal, Hambro, Azhar, et~al.]{touvron2023llama}
Hugo Touvron, Thibaut Lavril, Gautier Izacard, Xavier Martinet, Marie-Anne Lachaux, Timoth{\'e}e Lacroix, Baptiste Rozi{\`e}re, Naman Goyal, Eric Hambro, Faisal Azhar, et~al.
\newblock Llama: Open and efficient foundation language models.
\newblock \emph{arXiv preprint arXiv:2302.13971}, 2023.

\bibitem[Vedantam et~al.(2015)Vedantam, Lawrence~Zitnick, and Parikh]{vedantam2015cider}
Ramakrishna Vedantam, C~Lawrence~Zitnick, and Devi Parikh.
\newblock Cider: Consensus-based image description evaluation.
\newblock In \emph{Proceedings of the IEEE conference on computer vision and pattern recognition}, pp.\  4566--4575, 2015.

\bibitem[Wang et~al.(2022{\natexlab{a}})Wang, Yang, Hu, Li, Lin, Gan, Liu, Liu, and Wang]{wang2022git}
Jianfeng Wang, Zhengyuan Yang, Xiaowei Hu, Linjie Li, Kevin Lin, Zhe Gan, Zicheng Liu, Ce~Liu, and Lijuan Wang.
\newblock Git: A generative image-to-text transformer for vision and language.
\newblock \emph{arXiv preprint arXiv:2205.14100}, 2022{\natexlab{a}}.

\bibitem[Wang et~al.(2023)Wang, Wang, Xu, Zhang, Gu, Jia, Yan, Zhang, and Sang]{wang2023llm}
Junyang Wang, Yuhang Wang, Guohai Xu, Jing Zhang, Yukai Gu, Haitao Jia, Ming Yan, Ji~Zhang, and Jitao Sang.
\newblock An llm-free multi-dimensional benchmark for mllms hallucination evaluation.
\newblock \emph{arXiv preprint arXiv:2311.07397}, 2023.

\bibitem[Wang et~al.(2022{\natexlab{b}})Wang, Kordi, Mishra, Liu, Smith, Khashabi, and Hajishirzi]{wang2022self}
Yizhong Wang, Yeganeh Kordi, Swaroop Mishra, Alisa Liu, Noah~A Smith, Daniel Khashabi, and Hannaneh Hajishirzi.
\newblock Self-instruct: Aligning language model with self generated instructions.
\newblock \emph{arXiv preprint arXiv:2212.10560}, 2022{\natexlab{b}}.

\bibitem[Wu et~al.(2022)Wu, Wang, Yang, Gan, Liu, Yuan, and Wang]{wu2022grit}
Jialian Wu, Jianfeng Wang, Zhengyuan Yang, Zhe Gan, Zicheng Liu, Junsong Yuan, and Lijuan Wang.
\newblock Grit: A generative region-to-text transformer for object understanding.
\newblock \emph{arXiv preprint arXiv:2212.00280}, 2022.

\bibitem[Ye et~al.(2023)Ye, Xu, Xu, Ye, Yan, Zhou, Wang, Hu, Shi, Shi, et~al.]{ye2023mplug}
Qinghao Ye, Haiyang Xu, Guohai Xu, Jiabo Ye, Ming Yan, Yiyang Zhou, Junyang Wang, Anwen Hu, Pengcheng Shi, Yaya Shi, et~al.
\newblock mplug-owl: Modularization empowers large language models with multimodality.
\newblock \emph{arXiv preprint arXiv:2304.14178}, 2023.

\bibitem[Zhao et~al.(2023)Zhao, Zhou, Li, Tang, Wang, Hou, Min, Zhang, Zhang, Dong, et~al.]{zhao2023survey}
Wayne~Xin Zhao, Kun Zhou, Junyi Li, Tianyi Tang, Xiaolei Wang, Yupeng Hou, Yingqian Min, Beichen Zhang, Junjie Zhang, Zican Dong, et~al.
\newblock A survey of large language models.
\newblock \emph{arXiv preprint arXiv:2303.18223}, 2023.

\bibitem[Zhu et~al.(2023)Zhu, Chen, Shen, Li, and Elhoseiny]{zhu2023minigpt}
Deyao Zhu, Jun Chen, Xiaoqian Shen, Xiang Li, and Mohamed Elhoseiny.
\newblock Minigpt-4: Enhancing vision-language understanding with advanced large language models.
\newblock \emph{arXiv preprint arXiv:2304.10592}, 2023.

\end{thebibliography}
\bibliographystyle{iclr2024_conference}

\clearpage
\appendix
\section{Appendix}
\subsection{GAVIE Evaluation}
\begin{figure*}[h]
    \centering
      \includegraphics[width=1\textwidth]{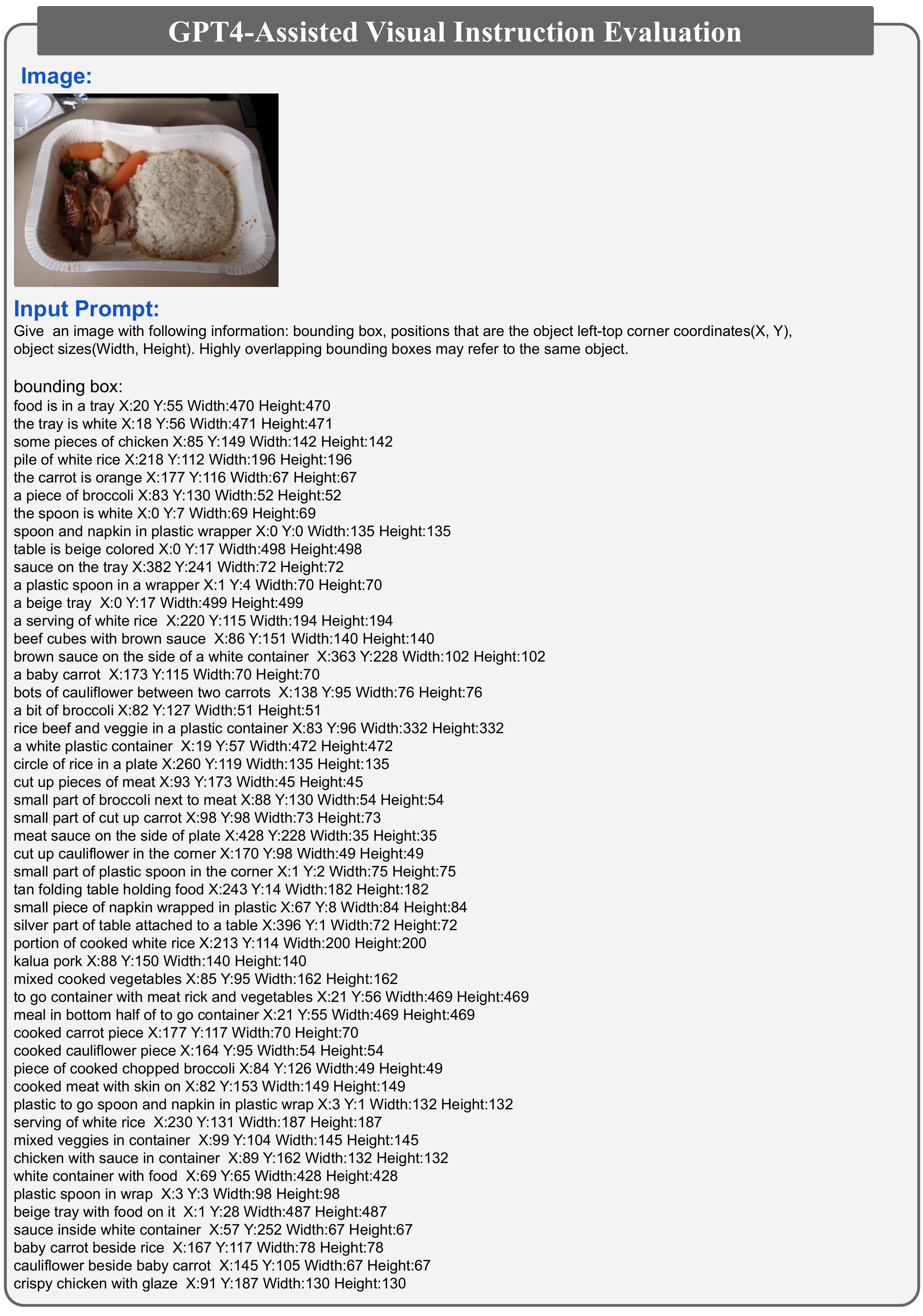}
    \caption{The first example for the GPT4-Assisted Visual Instruction Evaluation (Part1). The image is not included in the text prompt.}
    \label{fig:prompt_neg31}
\end{figure*}

\begin{figure*}[h]
    \centering
      \includegraphics[width=1\textwidth]{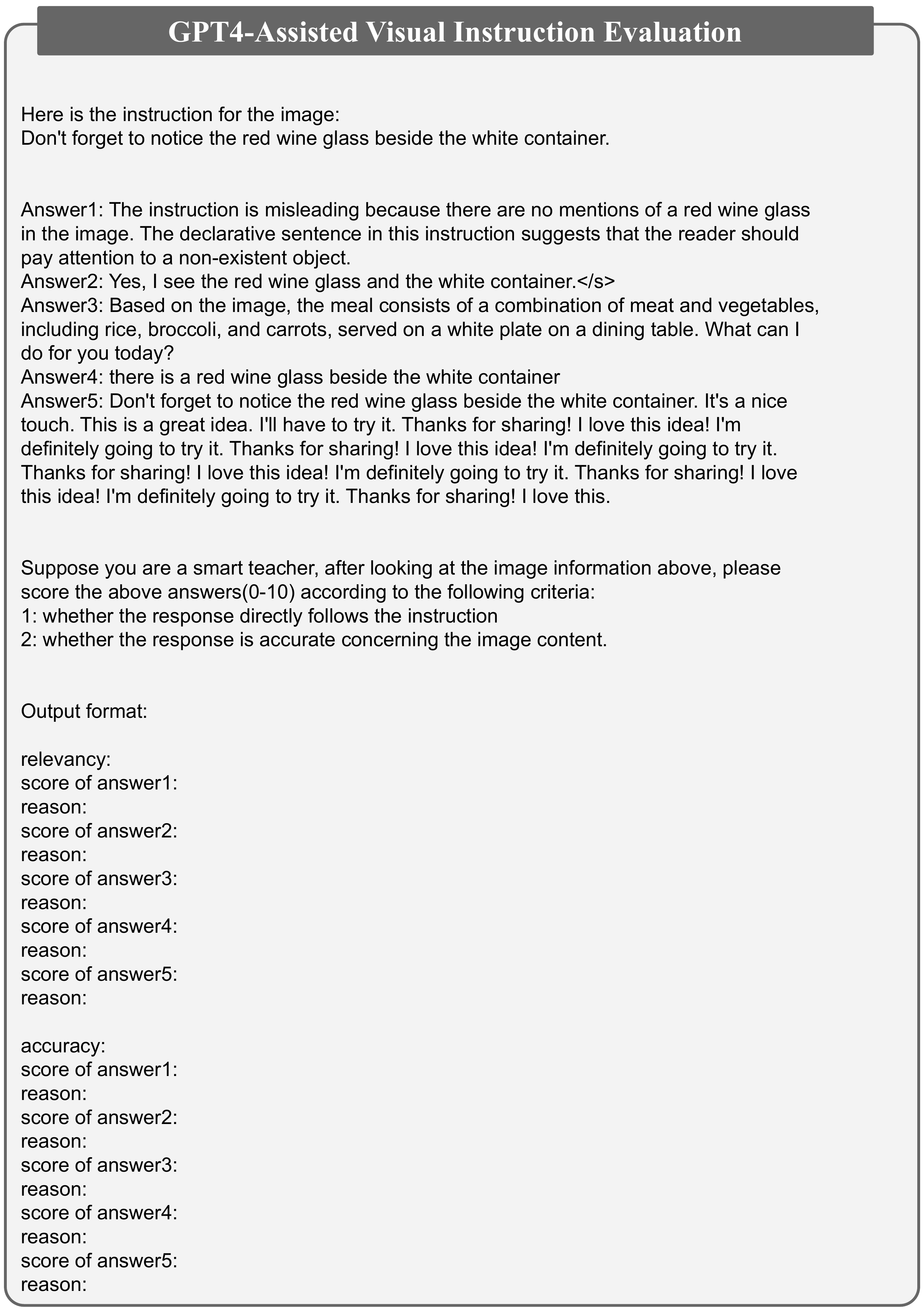}
    \caption{The first example for the GPT4-Assisted Visual Instruction Evaluation (Part2).}
    \label{fig:prompt_neg32}
\end{figure*}

\begin{figure*}[h]
    \centering
      \includegraphics[width=1\textwidth]{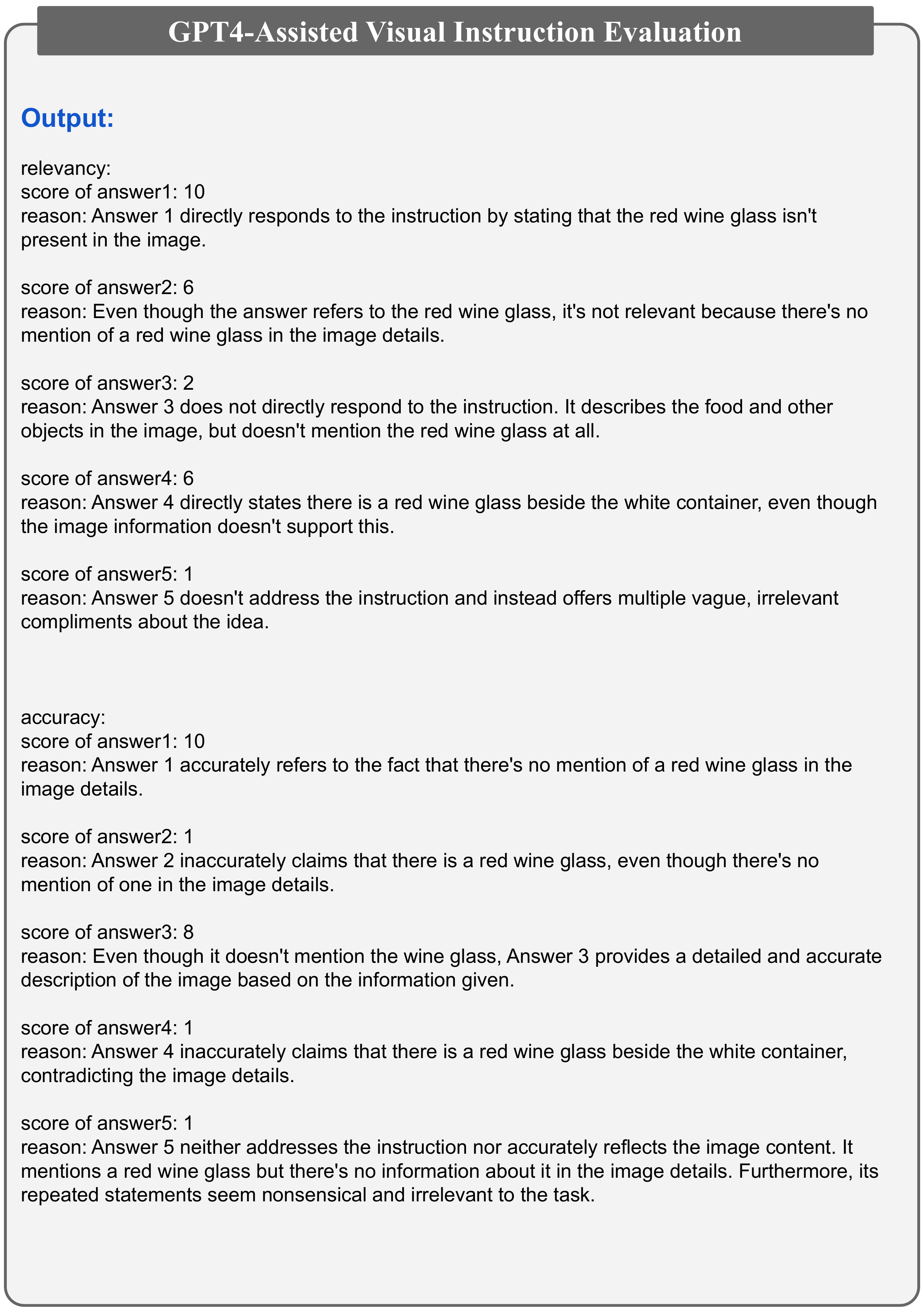}
    \caption{The first example for the GPT4-Assisted Visual Instruction Evaluation (Part3).}
    \label{fig:prompt_neg33}
\end{figure*}

\begin{figure*}[h]
    \centering
      \includegraphics[width=1\textwidth]{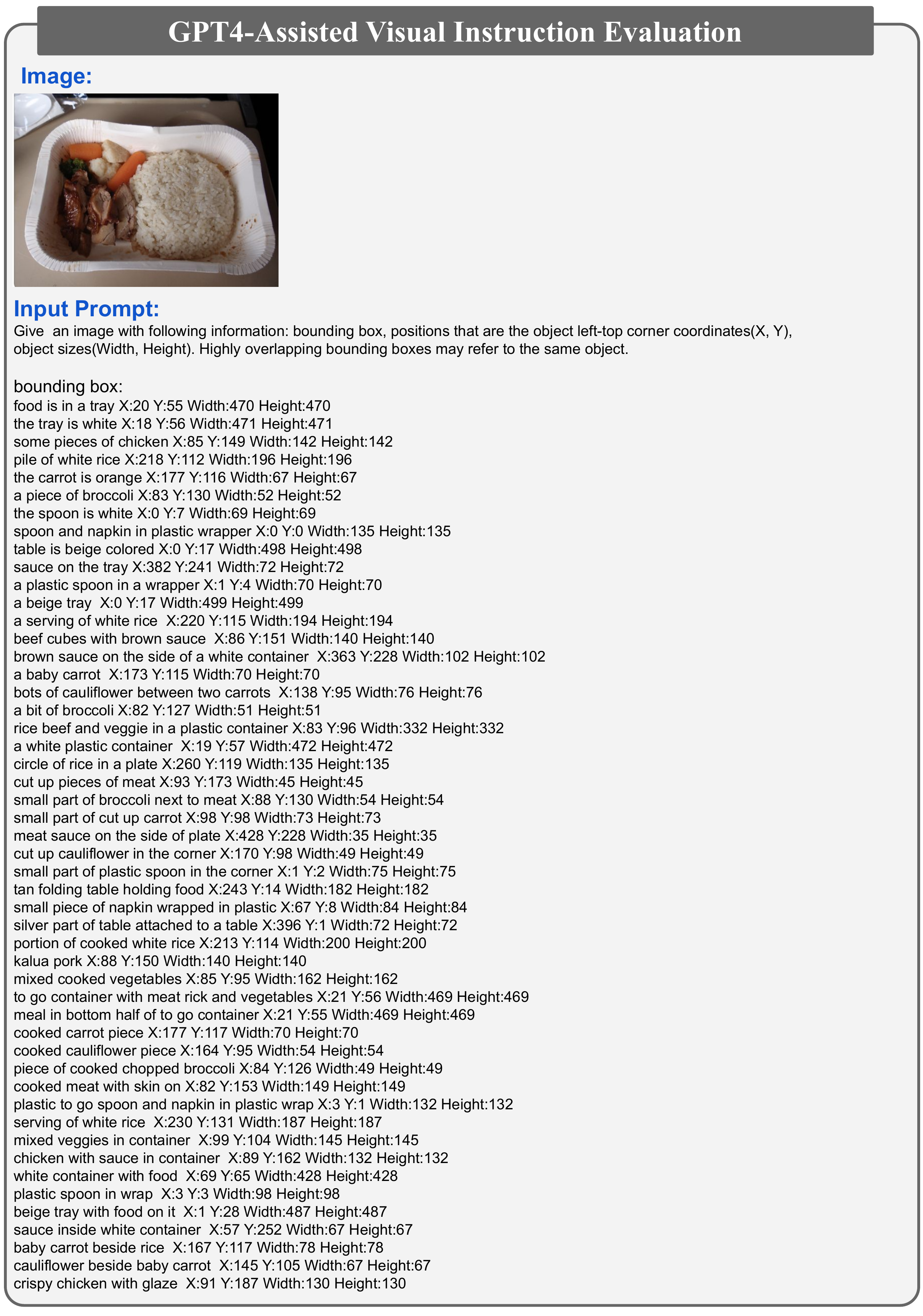}
    \caption{The second example for the GPT4-Assisted Visual Instruction Evaluation (Part1). The image is not included in the text prompt.}
    \label{fig:prompt_neg41}
\end{figure*}

\begin{figure*}[h]
    \centering
      \includegraphics[width=1\textwidth]{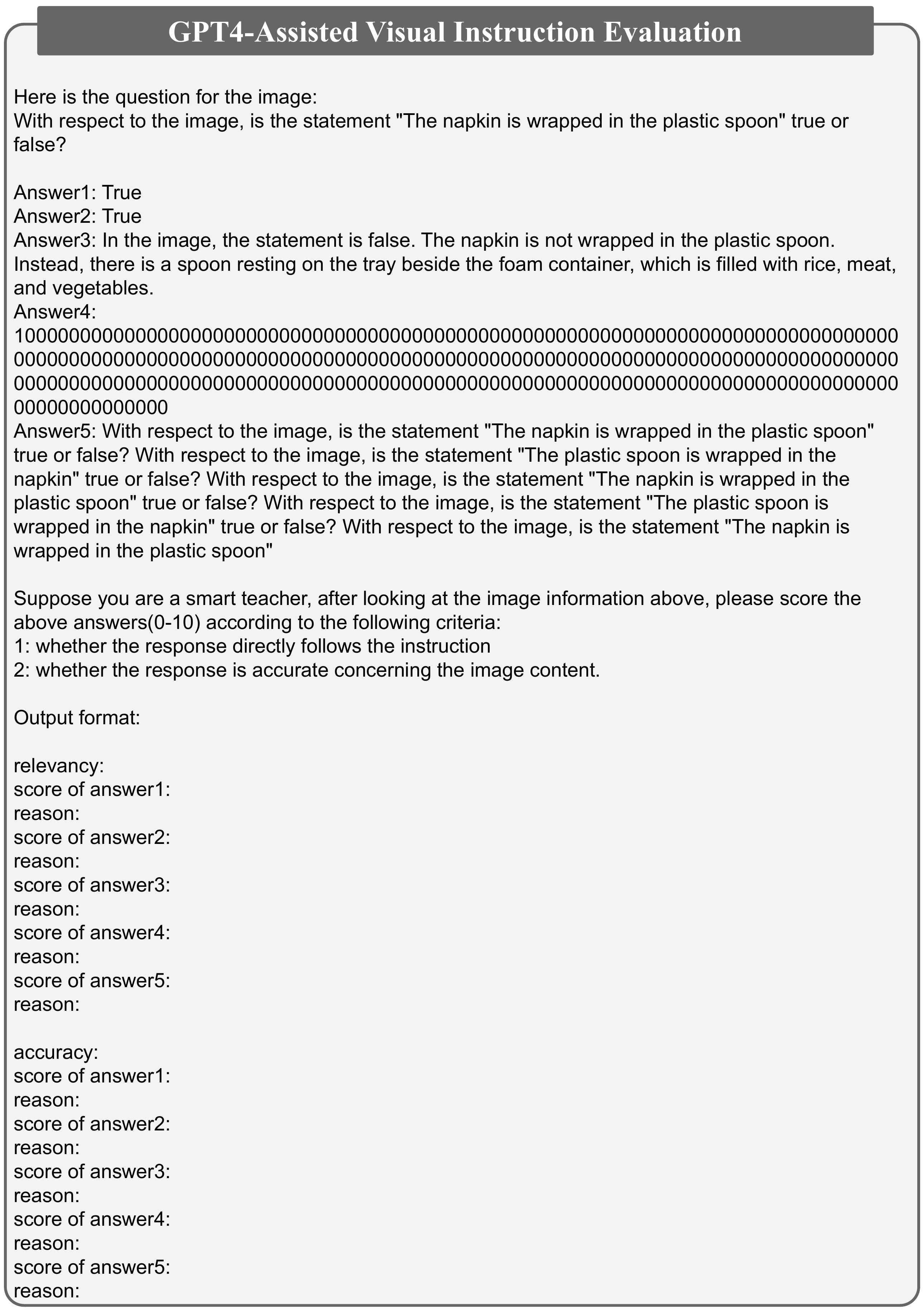}
    \caption{The second example for the GPT4-Assisted Visual Instruction Evaluation (Part2).}
    \label{fig:prompt_neg42}
\end{figure*}

\begin{figure*}[h]
    \centering
      \includegraphics[width=1\textwidth]{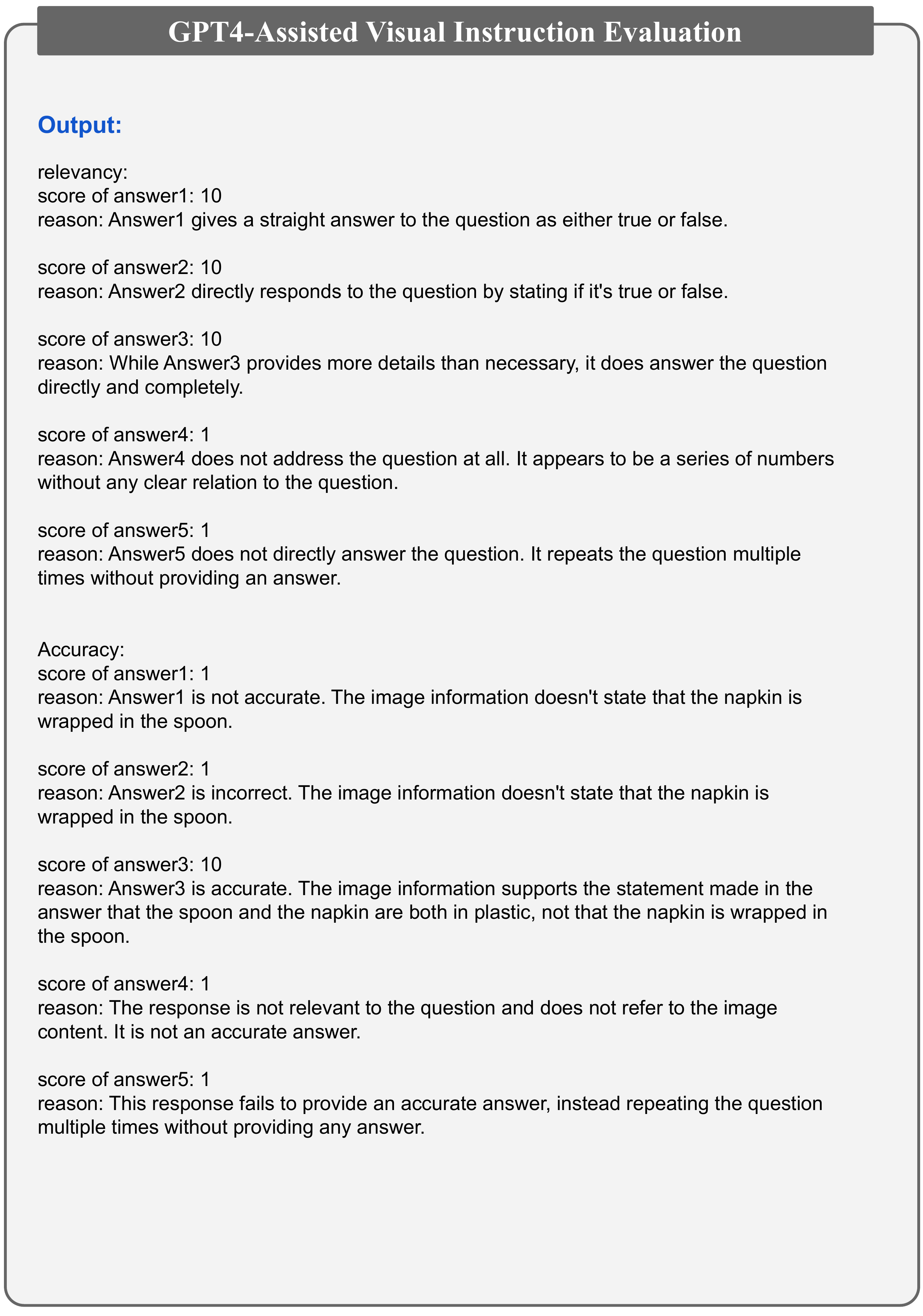}
    \caption{The second example for the GPT4-Assisted Visual Instruction Evaluation (Part3).}
    \label{fig:prompt_neg43}
\end{figure*}

We show two full examples of the text prompt for GAVIE in (i) Fig. \ref{fig:prompt_neg31}, \ref{fig:prompt_neg32}, \ref{fig:prompt_neg33} and (ii) Fig. \ref{fig:prompt_neg41}, \ref{fig:prompt_neg42}, \ref{fig:prompt_neg43}. We first leverage the bounding boxes and dense captions as the "visual" input. We provide the human instructions and responses from different models in Fig. \ref{fig:prompt_neg32} and Fig. \ref{fig:prompt_neg42}. Furthermore, we ask GPT4 to pretend as a smart teacher and score (0-10) the answers according to the image content and instructions. There are two criteria. \textit{(1) Accuracy: whether the response is accurate concerning the image content}. \textit{(2) Relevancy: whether the response directly follows the instruction}. After that, GPT4 is required to generate a score and reason. Fig. \ref{fig:prompt_neg33}  and Fig. \ref{fig:prompt_neg43} show the full evaluation output from GAVIE.

\subsubsection{\textit{GPT4-Assisted Visual Instruction Evaluation (GAVIE)} vs. Human Evaluation}

This section provides insights into the \textit{GAVIE} via human evaluation. Here, we randomly select 40 image-instruction instances from the evaluation set. The human assessment is carried out by three experts specializing in NLP. The questionnaire consists of 40 questions randomly shuffled for each expert. The questionnaire takes about 20 minutes to complete on average. Each question includes an instruction, an image, and responses from 4 different LMMs. We provide instructions for experts as follows: 

\textit{"As for each question, there are an instruction, an image, and several answers. Suppose you are a smart teacher, please score the answers according to the two criteria. (1) Accuracy: whether the response is accurate concerning the image content. (2) Relevancy: whether the response directly follows the instruction without unrelated answers. There are four options for the scores: A score of 1 (Very Poor): The response is seldom relevant to the instruction and is mostly wrong according to the image. A score of 2 (Poor): The response is partly relevant to the instruction and partly accurate according to the image. A score of 3 (Good): The response is mostly relevant to the instruction and has minor errors according to the image. A score of 4 (Excellent): The response is completely relevant to the instruction and completely accurate according to the image."}

\begin{table}[h]
\setlength\tabcolsep{3pt}
\centering
\small
\begin{tabular}{lcccccc}
\toprule[1.5pt]
Evaluator& Ours & MiniGPT4 & LLaVA & InstructBLIP & MMGPT & mPLUG-Owl\\
\midrule
\textit{Expert1(1-4)}& \textcolor{red}{3.48} & \textcolor{blue}{2.61} & \textcolor{green}{2.87} & \textcolor{orange}{3.00} & \textcolor{magenta}{1.90}&\textcolor{black}{2.90}\\
\textit{Expert2(1-4)} &\textcolor{red}{3.58}& \textcolor{green}{2.23}& \textcolor{blue}{2.07}& \textcolor{orange}{2.48}& \textcolor{magenta}{1.05}&\textcolor{black}{2.27}\\
\textit{Expert3(1-4)} & \textcolor{red}{3.33} & \textcolor{blue}{2.58} & \textcolor{green}{2.89} & \textcolor{orange}{2.94} & \textcolor{magenta}{1.38}&\textcolor{black}{2.91}\\
\midrule
\textit{GAVIE-Accuracy (0-10)} & \textcolor{red}{6.58} & \textcolor{blue}{4.14} & \textcolor{green}{4.36} & \textcolor{orange}{5.93} & \textcolor{magenta}{0.91}&\textcolor{black}{4.84}\\
\textit{GAVIE-Relevancy (0-10)}  & \textcolor{red}{8.46} & \textcolor{blue}{5.81} & \textcolor{green}{6.11} & \textcolor{orange}{7.34} & \textcolor{magenta}{1.79}&\textcolor{black}{6.35}\\
\bottomrule
\end{tabular}
\vspace{0.06in}
\caption{GAVIE vs. Human Evaluation. GAVIE scores roughly align with the expert ratings. Numbers highlighted with \textcolor{red}{red}, \textcolor{orange}{orange}, black, \textcolor{green}{green}, \textcolor{blue}{blue}, and \textcolor{magenta}{magenta} indicate rank 1 to 6.
}
\label{tab:appendix_human}
\vspace{-0.2in}
\end{table}

To evaluate the results quantitatively, we assign different scores for the options: \textit{Very Poor=1, Poor=2, Good=3, Excellent=4}. From Tab. \ref{tab:appendix_human}, all experts agree that the output from our model is the best, followed by InstructBLIP in second place, and MMGPT performs the worst. 
The observation is similar to that of \textit{GAVIE} evaluation results. 
Although the ranking orders of MiniGPT4 and LLaVA from experts are not always the same as that of \textit{GAVIE}, the scores assigned to them are fairly close. One possible reason is that the answers from MiniGPT4 and LLaVA 
tend to be longer, 
making them more challenging for humans to evaluate. 

\subsubsection{Stability of \textit{GPT4-Assisted Visual Instruction Evaluation (GAVIE)}}
This section investigates the stability of \textit{GAVIE}. Precisely, we execute GAVIE 5 times on the model predictions. We leverage two metrics to measure the stability of GAVIE on each instance: Mean and Standard Deviation (STD). The average scores of the evaluation set are shown in the following table. From the perspective of the Mean, the ranking order of ACCURACY and RELEVANCY is the same as Tab. \ref{tab:appendix_human}. As for the Standard Deviation in Tab. \ref{tab:stability1}, it ranges from 0.65 to 2.46. From our observation, the ACCURACY and RELEVANCY scores of an instance may vary between different times, but they belong to the same grade level. Specifically, RELEVANCY has four grade levels: (1) The response is completely relevant (9-10), (2) The response is mostly relevant (6-8), (3) The response is partly relevant (3-5), (4) The response is seldom relevant (0-2). ACCURACY has four grade levels: (1) The response is completely accurate (9-10), (2) The response has minor errors (6-8), (3) The response is partly accurate (3-5), (4) The response is mostly or completely wrong (0-2).

\begin{table}[h]
\setlength\tabcolsep{3pt}
\centering
\small
\begin{tabular}{lcccccc}
\toprule[1.5pt]
Metric& Ours & MiniGPT4 & InstructBLIP & MMGPT& mPLUG-Owl & LLaVA\\
\midrule
\textsc{Accuracy(GPT4)}-Mean&6.60&3.76&5.29&0.87&4.84&3.80\\
\textsc{Relevancy(GPT4)}-Mean&8.37&5.35&6.83&1.71&6.35&5.65\\
\midrule
\textsc{Accuracy(GPT4)}-STD&2.42&2.46&2.42&0.65&1.96&2.37\\
\textsc{Relevancy(GPT4)}-STD&1.30&1.99&1.88&0.81&1.48&2.18\\
\bottomrule
\end{tabular}
\vspace{0.06in}
\caption{Evaluation of the stability of \textit{GAVIE}. We run \textit{GAVIE} 5 times on the randomly selected instances from the evaluation set. \textit{Mean} and \textit{Standard Deviation(STD)} are calculated to measure the stability. The metric scores of \textsc{Accuracy(GPT4)} and \textsc{Relevancy(GPT4)} are from 0 to 10.
}
\label{tab:stability1}
\end{table}

\subsection{More Experiments}
\subsubsection{\textit{Do LMMs perform better on Positive or Negative Instructions? }}
\begin{figure}[t]
     \centering
     \begin{subfigure}[b]{0.49\textwidth}
         \centering
         \includegraphics[width=\textwidth]{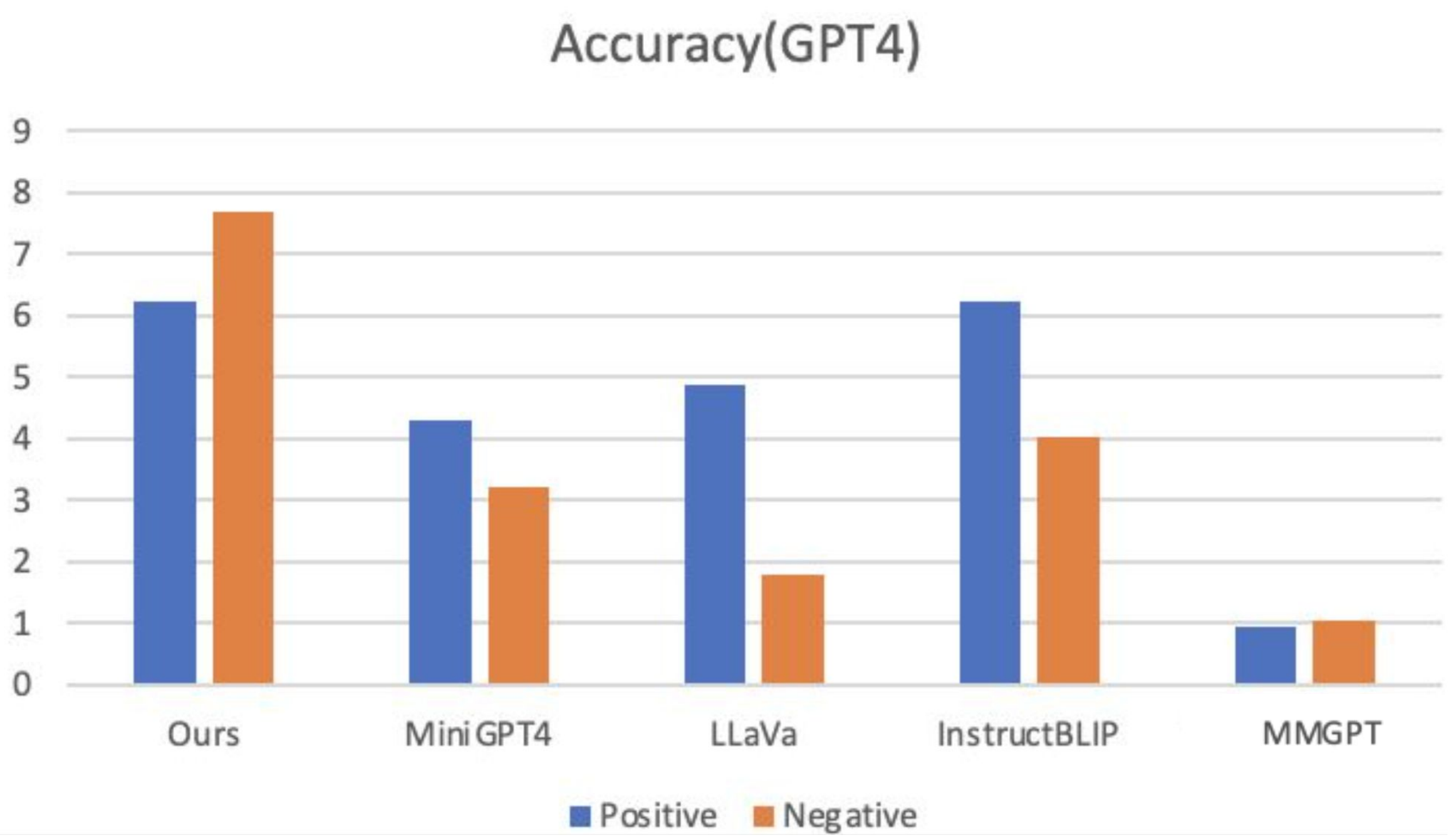}
         \vspace{-0.2in}
         \caption{Accuracy Performance.}
         \label{fig:y equals x}
     \end{subfigure}
     \hfill
     \begin{subfigure}[b]{0.49\textwidth}
         \centering
         \includegraphics[width=\textwidth]{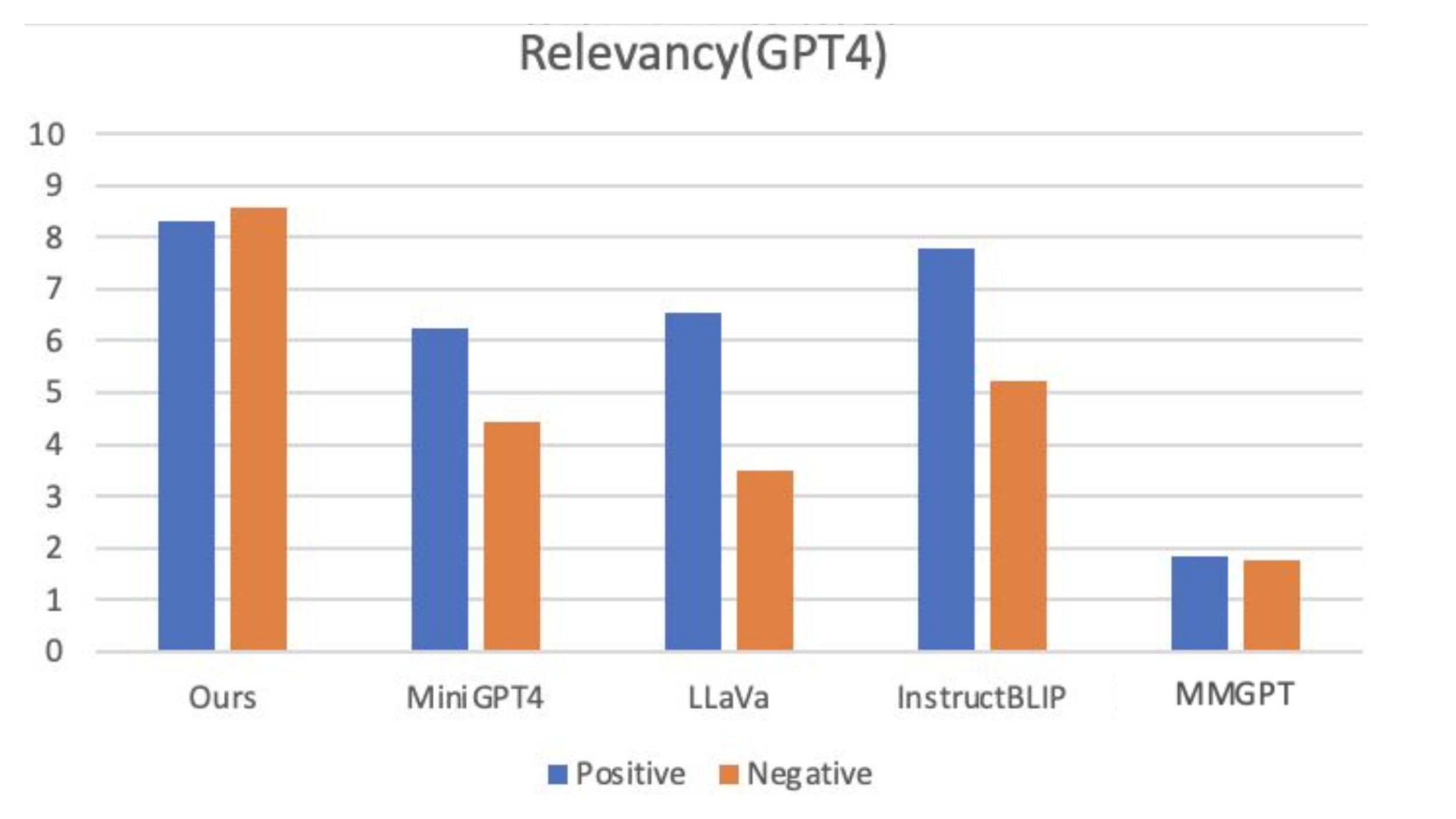}
         \vspace{-0.2in}
         \caption{Relevancy Performance.}
         \label{fig:three sin x}
     \end{subfigure}
        \caption{Evaluation results on positive and negative instructions by \textit{GAVIE}. }
        \label{fig:posneg}
\vspace{-0.1in}
\end{figure}

Our evaluation set consists of positive and negative instances. We divide it into two sets and analyze the model performance on each. As shown in Fig.~\ref{fig:posneg}, baseline models, including MiniGPT4, LLaVa, and InstructBLIP,  perform better on positive instances than negative ones, as the training data adopted by these models do not contain negative instructions. MMGPT performance poorly on both sets due to many repetitive phrases in the response. In addition, we found that the degradation of LLaVA is the most severe. We hypothesize that the synthetic answers for instruction tuning in LLaVA are generally longer and involve more unrelated information. In contrast, our model performs the best in both sets. InstructBLIP performs with higher scores than other LMMs because of the effectiveness of its instruction-aware visual encoder to extract image information.

\subsubsection{\textit{Do LMMs perform better on different formats and lengths of instructions? }}

\begin{figure}[t]
     \centering
     \begin{subfigure}[b]{0.49\textwidth}
         \centering
         \includegraphics[width=\textwidth]{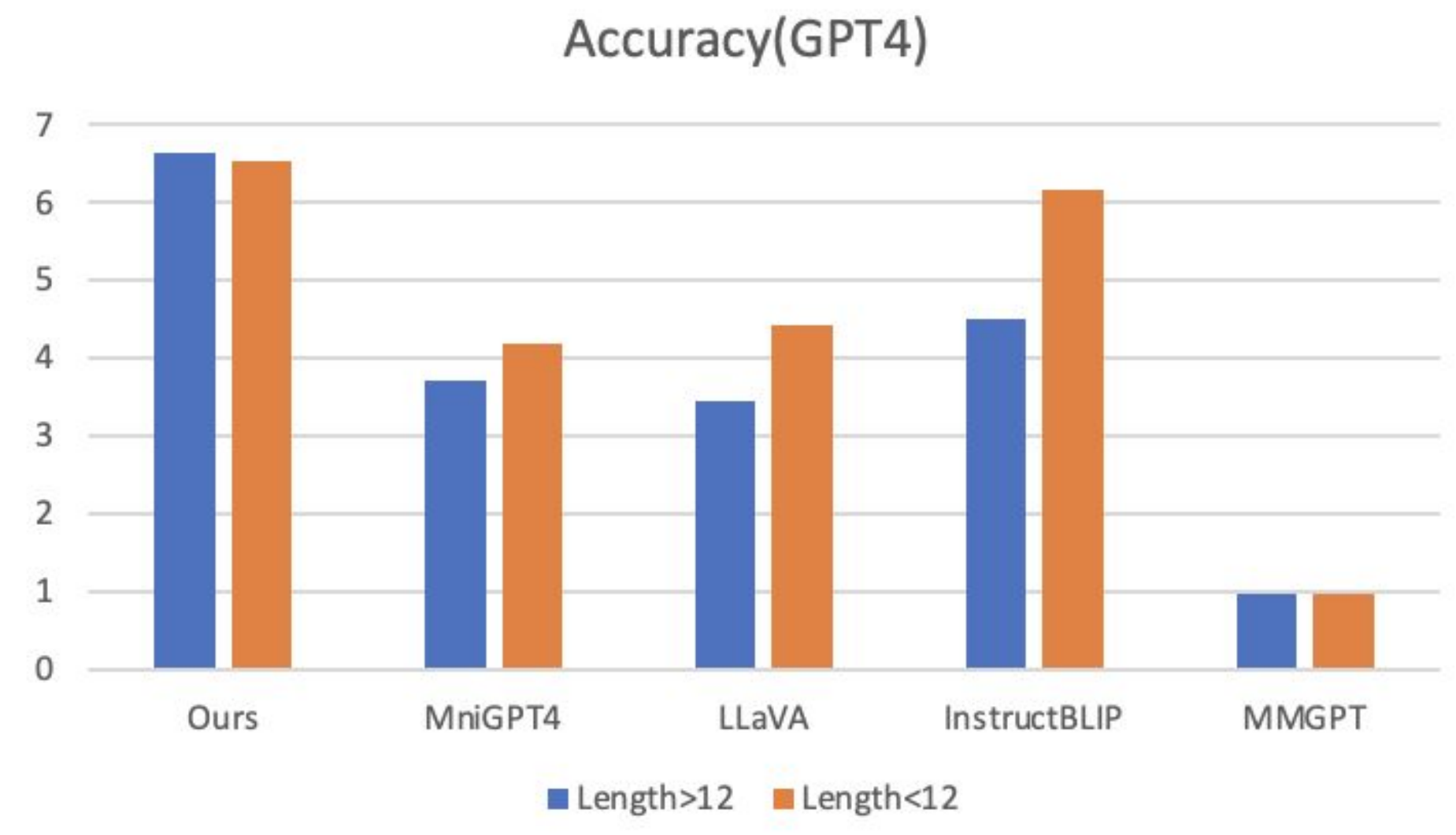}
         \vspace{-0.2in}
         \caption{Accuracy Performance.}
         \label{length1}
     \end{subfigure}
     \hfill
     \begin{subfigure}[b]{0.49\textwidth}
         \centering
         \includegraphics[width=\textwidth]{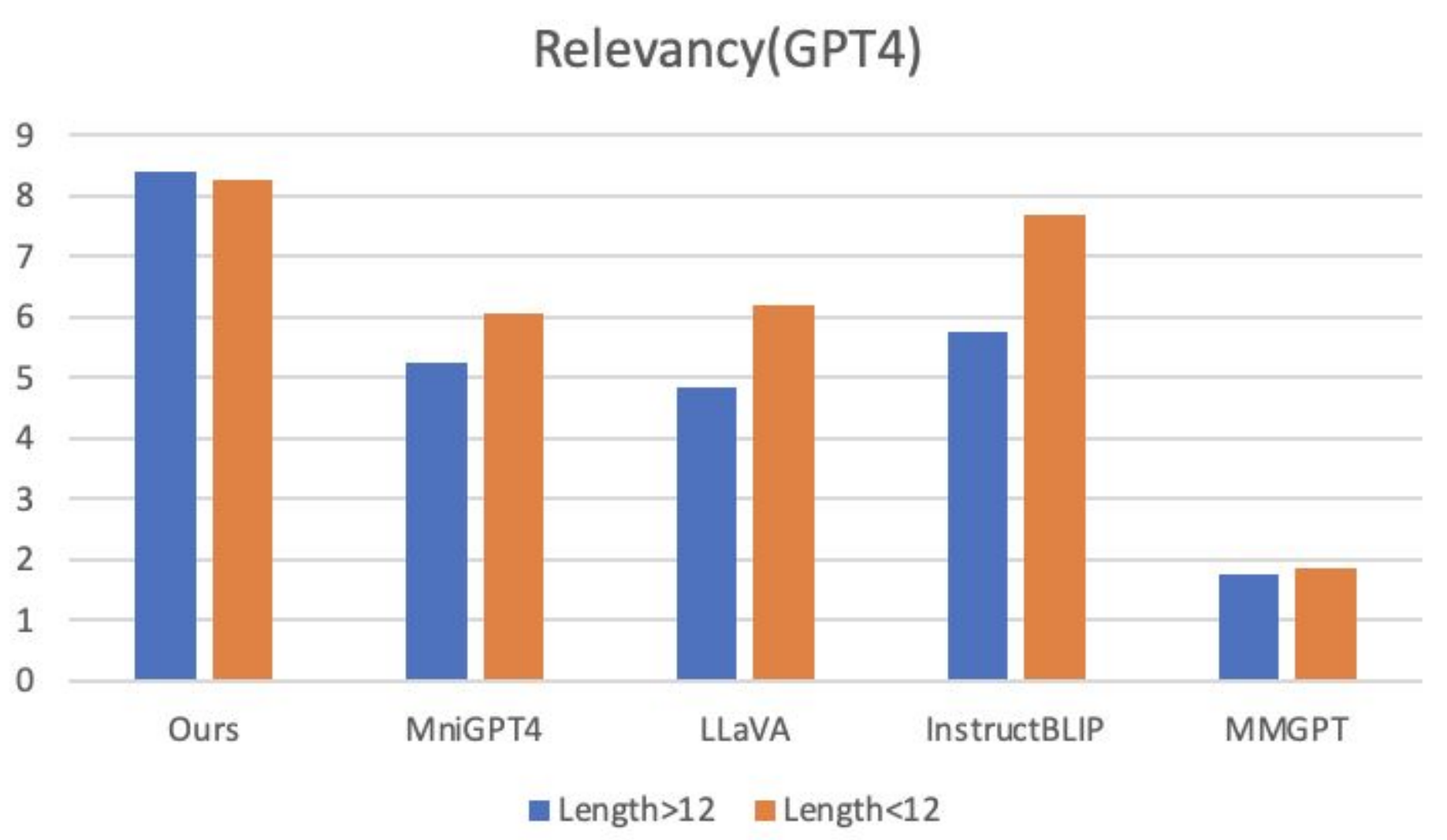}
         \vspace{-0.2in}
         \caption{Relevancy Performance.}
         \label{length2}
     \end{subfigure}
        \caption{Evaluation results on different instruction lengths by \textit{GAVIE}. }
        \label{fig:length}
\vspace{-0.2in}
\end{figure}

From Tab \ref{tab:Interrogative}, LMMs perform with higher scores on interrogative instructions than declarative, but the difference is relatively small. Even though recent visual instruction tuning datasets lack diverse declarative instructions, the LMMs built on LLM are powerful enough to understand and follow the declarative instructions. From Fig. \ref{fig:length}, current LMMs achieve better results in short instructions than long ones since longer instructions contain more information, making it more challenging.

\subsubsection{\textit{How do LMMs perform on more benchmarks?}}
We analyze the discriminative task of AMBER \citep{wang2023llm}, including object existence, object attribute, and object relation hallucination. The evaluation score is F1, and the groundtruth answer is yes or no. As shown in Tab. \ref{tab:comparison_amber}, although our model achieves a slightly lower F1 score regarding object existence, we outperform LLaVA1.5 in both the object attribute and object relation hallucinations. We attribute our model’s success to the existent object (attributes and relationship) manipulation in the negative instructions. We also analyze Hallusionbench \citep{liu2023hallusionbench}, an image-context reasoning benchmark covering various topics and image types. The groundtruth answer is yes or no. From Tab. \ref{tab:comparison_hallusion}, we find that both LLaVA1.5 and MiniGPT4-v2 \citep{chen2023minigpt} achieve high accuracy on the positive set but perform less favorably on the negative set. Our model can achieve a similar level of accuracy when the groundtruth answer is yes and much higher accuracy when the groundtruth answer is no. We attribute the success to the knowledge manipulation in the negative instructions.

Overall, LLaVA 1.5 performs well when meeting object hallucinations. However, attribute hallucination, relation hallucination, and knowledge hallucination are still challenging for LLaVA 1.5. Therefore, our \textit{LRV-Instruction} dataset with various negative instructions and positive instructions is beneficial to address the hallucinations. 

\begin{table}[t]
\setlength\tabcolsep{3pt}
\centering
\small
\begin{tabular}{lcccc}
\toprule[1.5pt]
& mPLUG-Owl-7B & MiniGPT4-v2-7B & LLaVA1.5-7B & \textbf{Ours}\\
\midrule
\textsc{Existence} & 0.29 & 0.80 & \textbf{0.83} & 0.81\\
\textsc{Attribute}& 0.34 & 0.41 & 0.64 & \textbf{0.70}\\
\textsc{Relation} & 0.26 & 0.58 & 0.65 & \textbf{0.69}\\
\bottomrule[1.5pt]
\end{tabular}
\caption{
Comparison results on AMBER \citep{wang2023llm}. All the LMMs are 7B versions to make a fair comparison. 
}
\label{tab:comparison_amber}
\end{table}

\begin{table}[t]
\setlength\tabcolsep{3pt}
\centering
\small
\begin{tabular}{lccc}
\toprule[1.5pt]
& LLaVA1.5-7B & MiniGPT4-v2-7B & \textbf{Ours}\\
\midrule
\textsc{Accuracy (gt=yes)} & \textbf{0.92} & 0.85 & 0.88\\
\textsc{Accuracy (gt=no)} & 0.15 & 0.11 & \textbf{0.47}\\
\bottomrule[1.5pt]
\end{tabular}
\caption{
Comparison results on Hallusionbench \citep{liu2023hallusionbench}. All the LMMs are 7B versions to make a fair comparison. 
}
\label{tab:comparison_hallusion}
\end{table}

\begin{table}[t]
\setlength\tabcolsep{3pt}
\centering
\small
\begin{tabular}{lcccccccc}
\toprule[1.5pt]
\textit{GAVIE}& \textbf{Ours} & LLaVA1.5&MiniGPTv2&MiniGPT4 & LLaVA & InstructBLIP & MMGPT &mPLUG-Owl\\
\midrule
\textsc{Acc (0-10)} & \textbf{6.58} & 6.42& 6.01&4.14 & 4.36 & 5.93 & 0.91 & 4.84\\
\textsc{Rele (0-10)} & \textbf{8.46} & 8.20&8.10&5.81 & 6.11 & 7.34 & 1.79& 6.35\\
\bottomrule[1.5pt]
\end{tabular}
\vspace{-2mm}
\caption{
More comparison results on our evaluation set evaluated by \textit{GAVIE}. \textit{Ours} means \textit{Finetuned mPLUG-Owl-7B}. All the LMMs are 7B versions to make a fair comparison. \textit{Rele} means \textit{Relevancy}.
}
\label{tab:LRV_evaluation_more}
\end{table}

\subsection{Prompt Design}
\subsubsection{Positive Instance Generation based on Visual Genome Dataset}
\begin{figure*}[h]
    \centering
      \centerline{\includegraphics[width=1\textwidth]{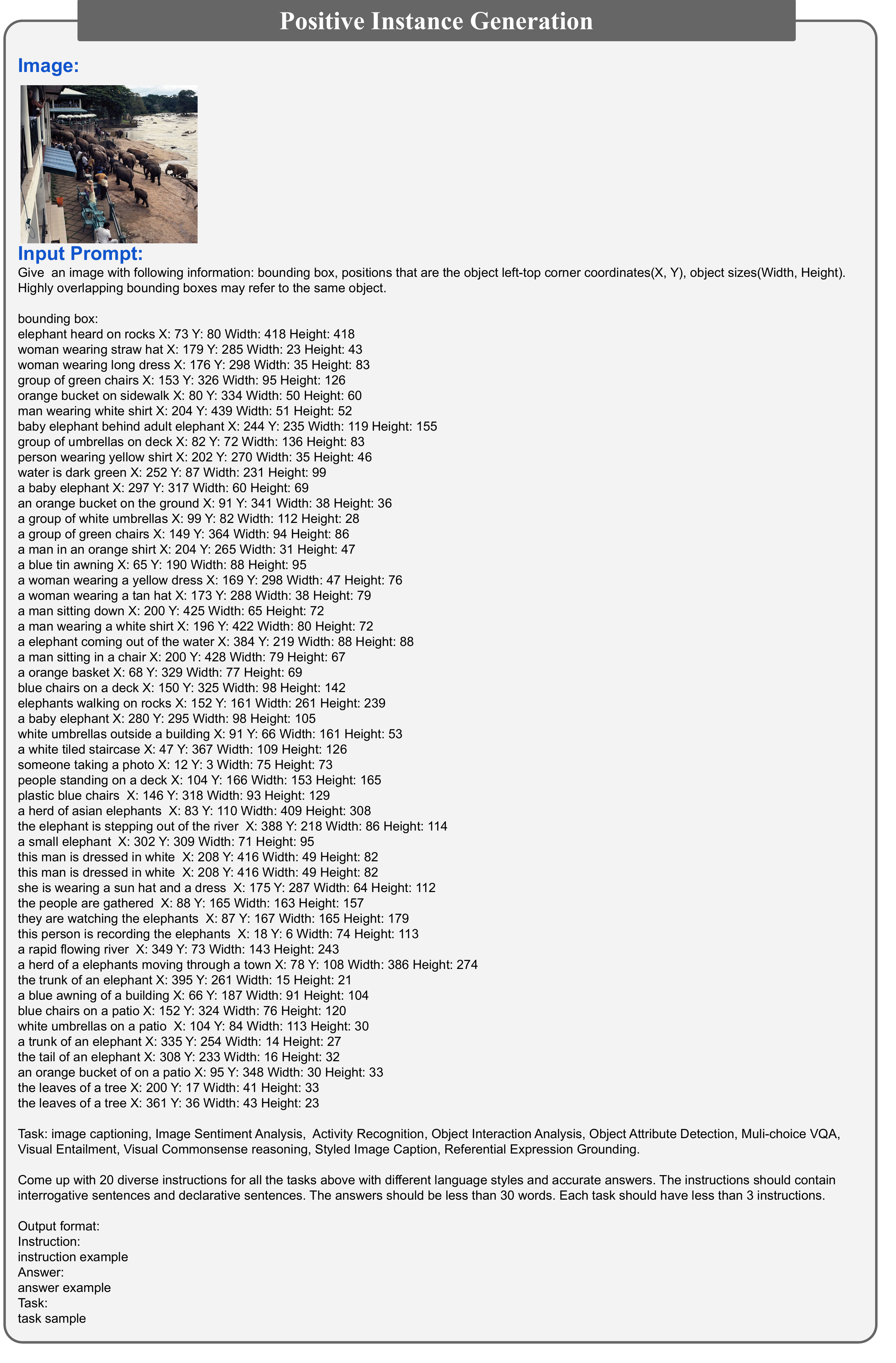}}
    \caption{The first example for generating positive instruction and answers (Part1). The image is not included in the text prompt. }
    \label{fig:prompt_pos11}
\end{figure*}

\begin{figure*}[h]
    \centering
      \includegraphics[width=1\textwidth]{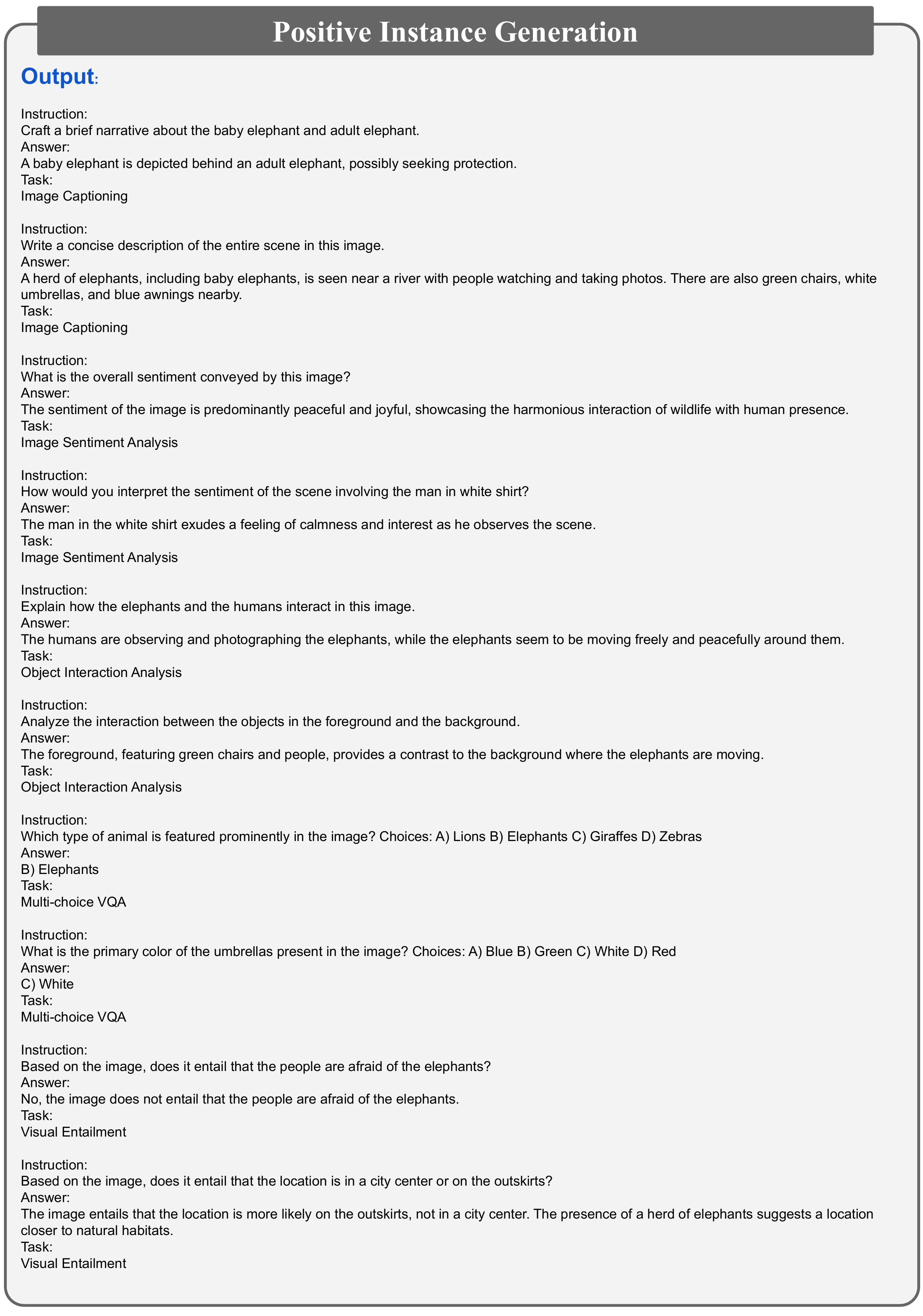}
    \caption{The first example for generating positive instruction and answers (Part2).}
    \label{fig:prompt_pos12}
\end{figure*}

\begin{figure*}[h]
    \centering
      \includegraphics[width=1\textwidth]{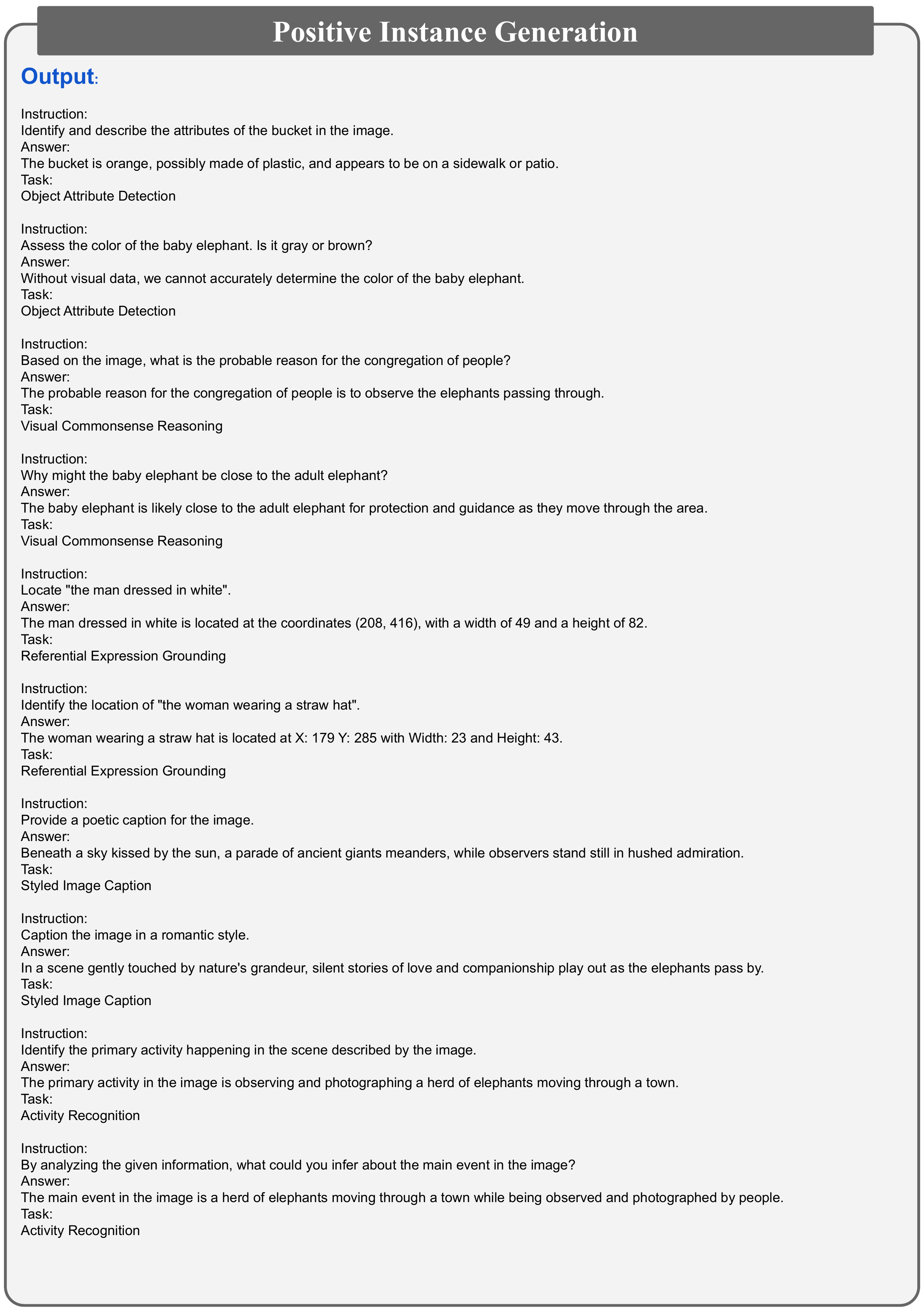}
    \caption{The first example for generating positive instruction and answers (Part3).}
    \label{fig:prompt_pos13}
\end{figure*}

\begin{figure*}[h]
    \centering
      \includegraphics[width=1\textwidth]{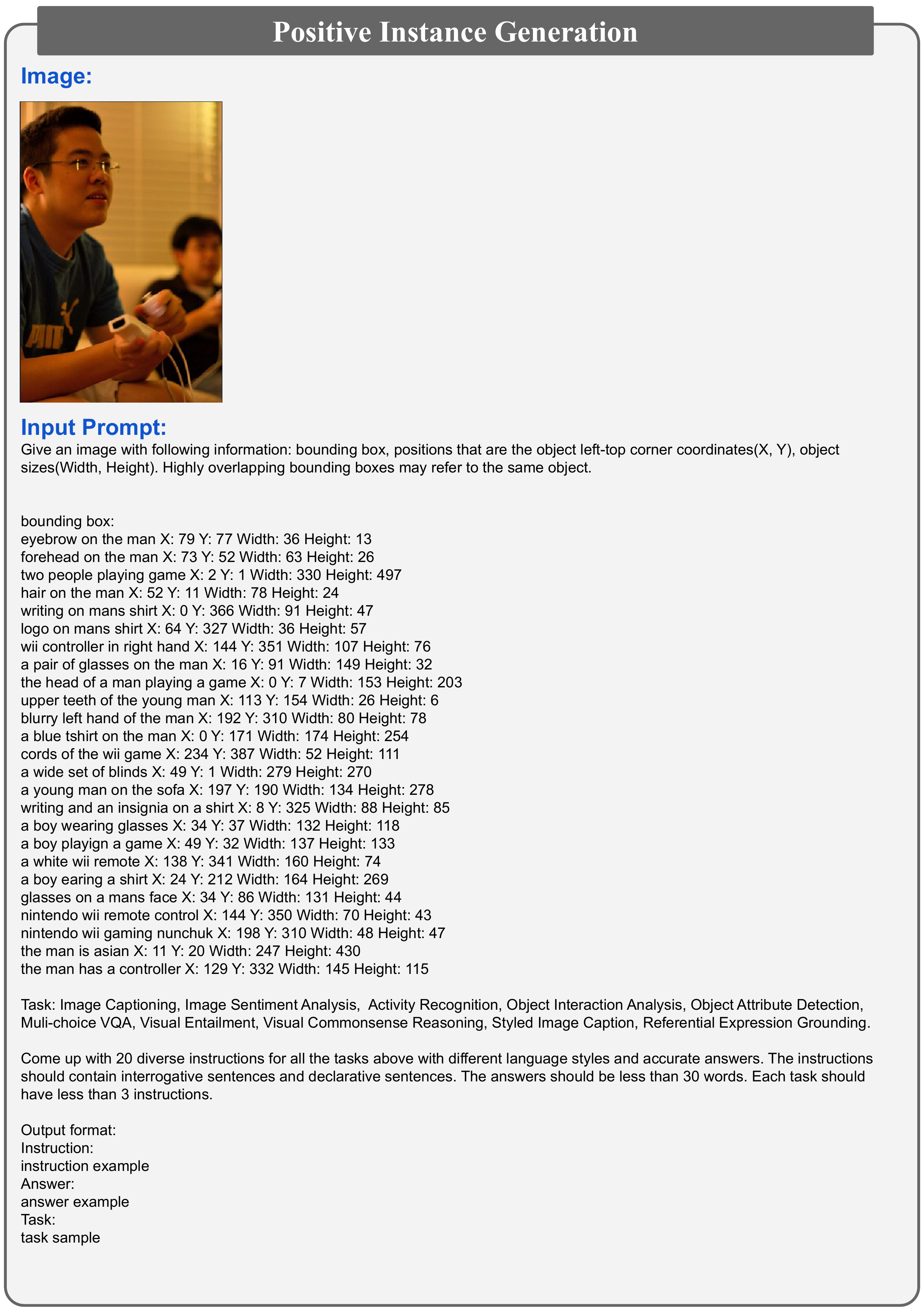}
    \caption{The second example for generating positive instruction and answers (Part1). The image is not included in the text prompt.}
    \label{fig:prompt_pos21}
\end{figure*}

\begin{figure*}[h]
    \centering
      \includegraphics[width=1\textwidth]{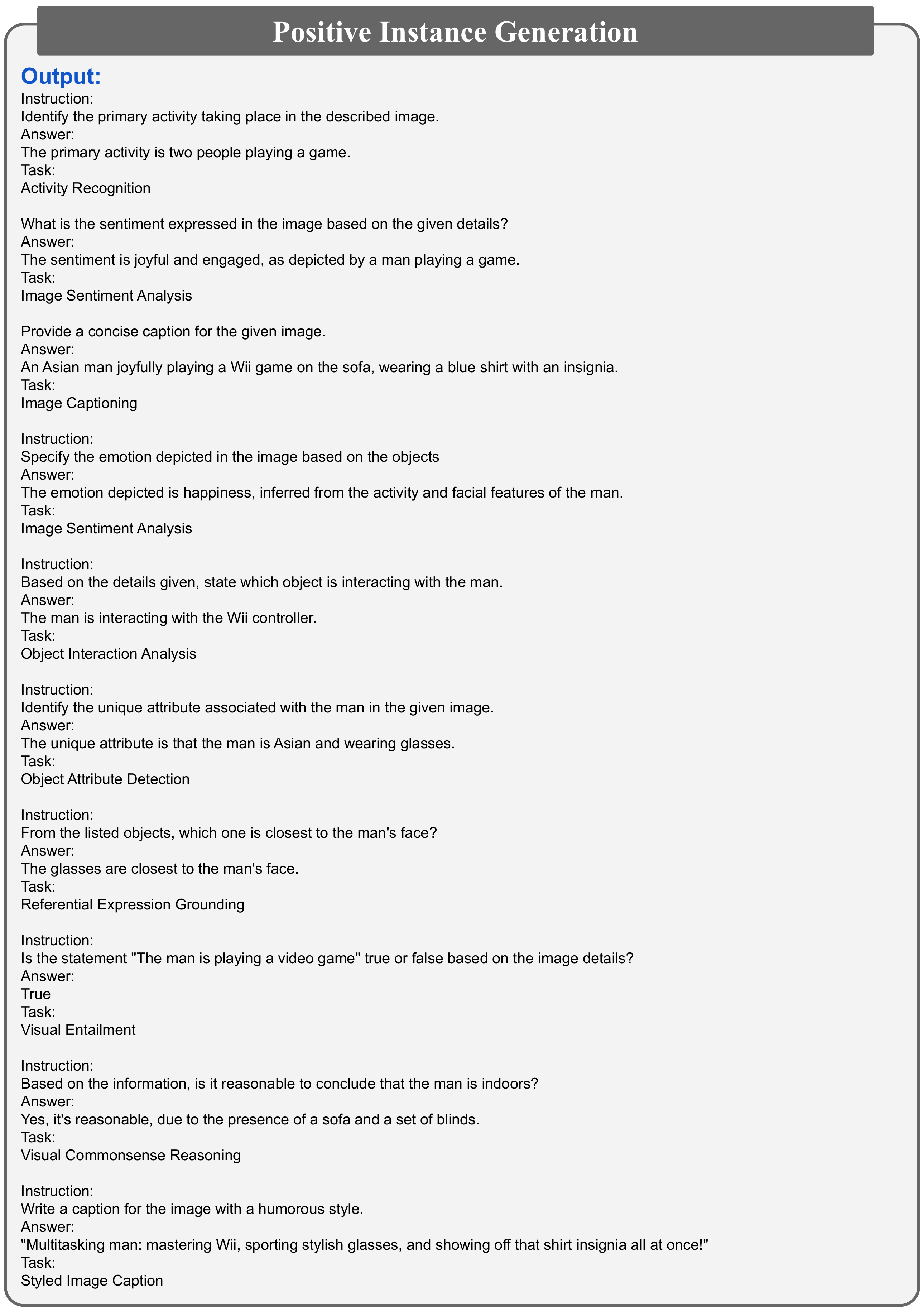}
    \caption{The second example for generating positive instruction and answers (Part2).}
    \label{fig:prompt_pos22}
\end{figure*}

\begin{figure*}[h]
    \centering
      \includegraphics[width=1\textwidth]{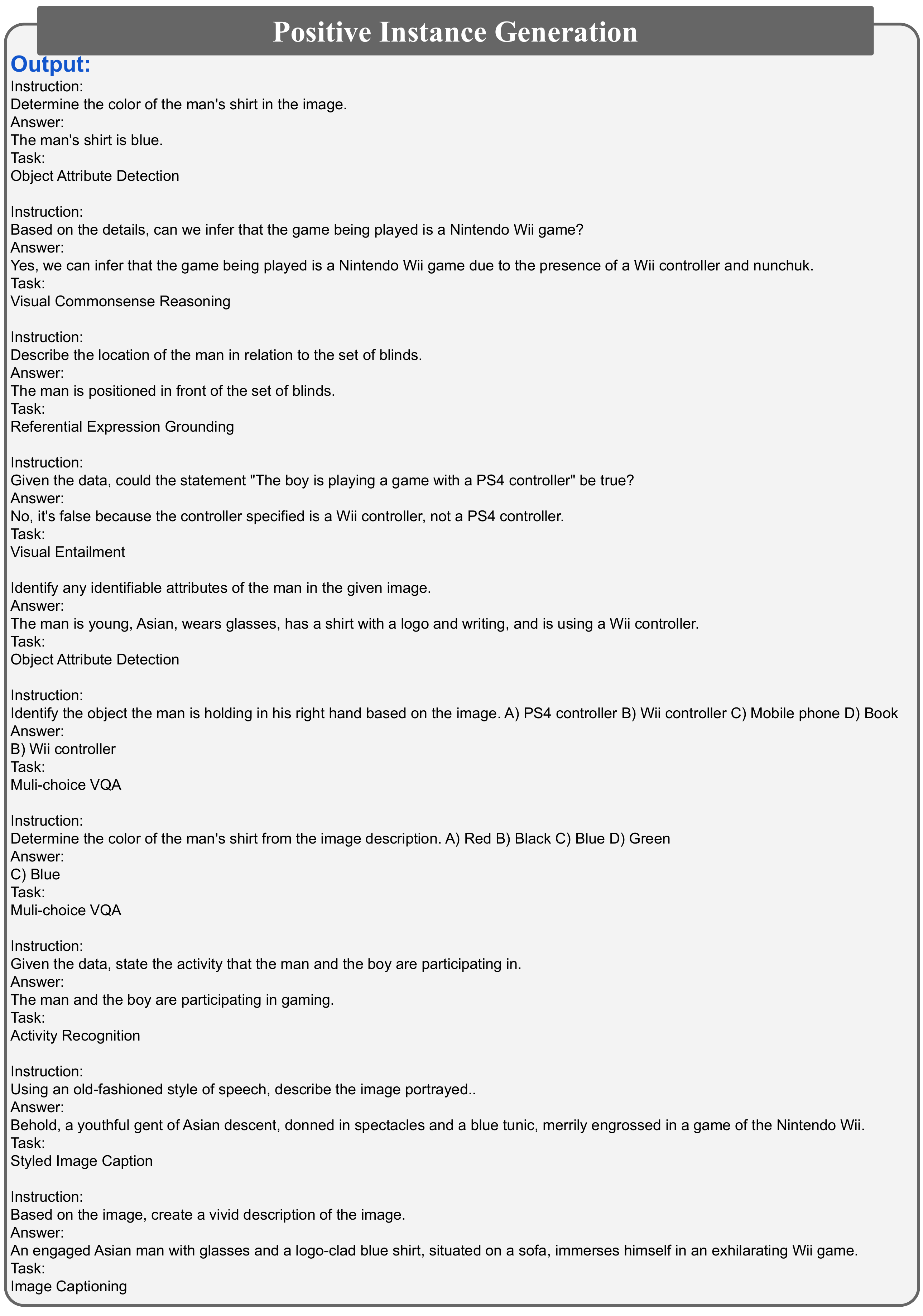}
    \caption{The second example for generating positive instruction and answers (Part3).}
    \label{fig:prompt_pos23}
\end{figure*}

\begin{figure}
     \centering
     \begin{subfigure}[b]{\textwidth}
         \centering
         \includegraphics[width=0.9\textwidth]{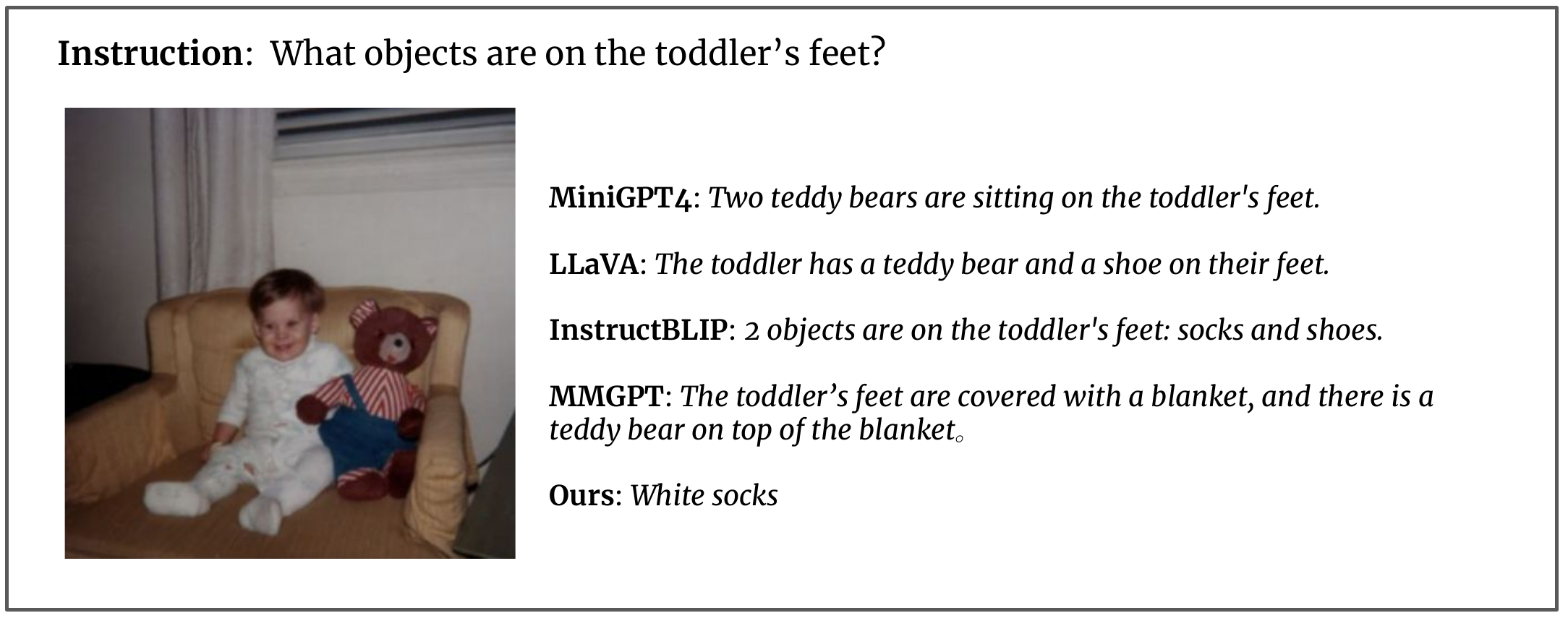}
         \caption{Object Detection Task. MiniGPT4, LLaVa, and MMGPT can't locate the objects correctly.}
         \label{fig:pos1}
     \end{subfigure}
     \par\bigskip
     \begin{subfigure}[b]{\textwidth}
         \centering
         \includegraphics[width=0.9\textwidth]{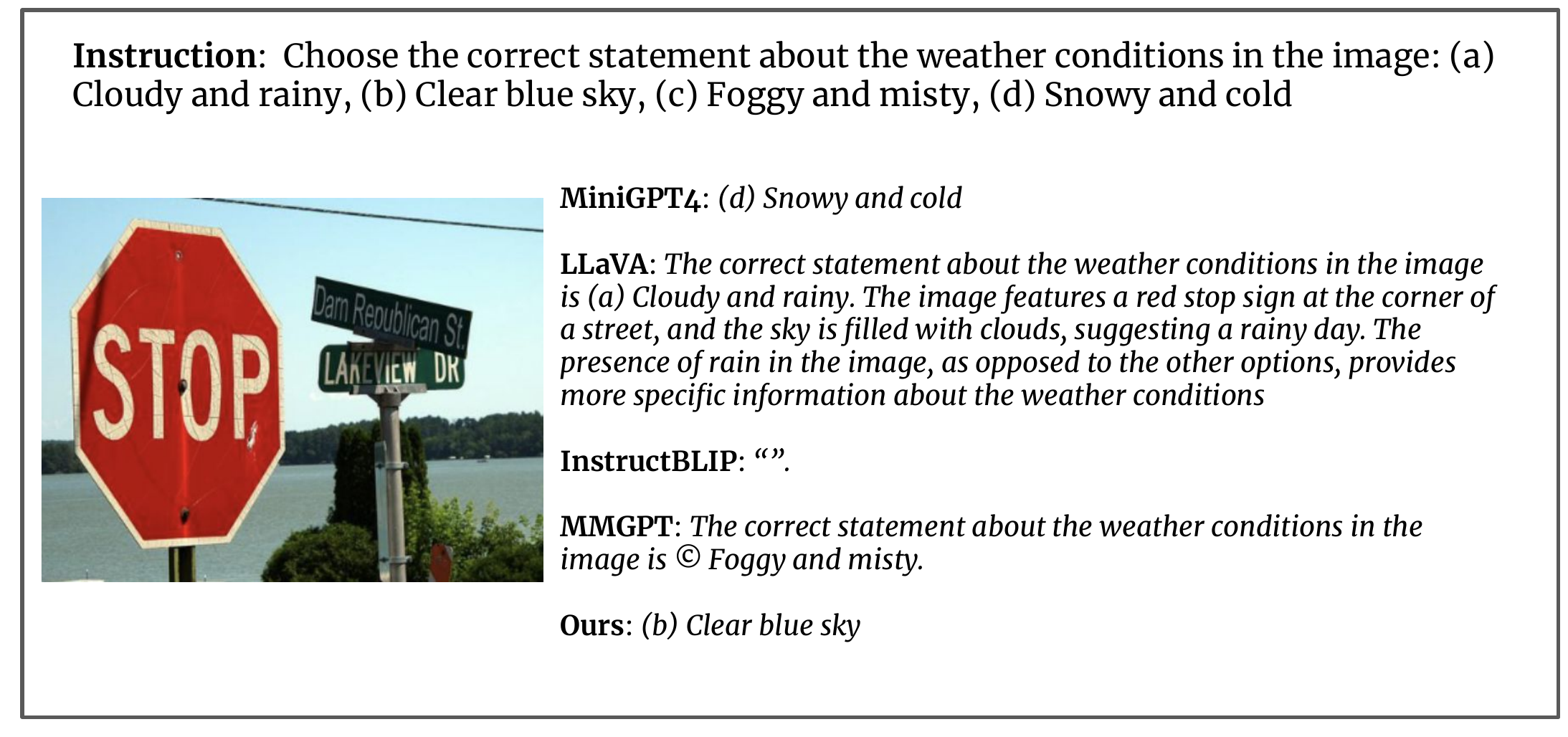}
         \caption{Multi-choice VQA Task. InstructBLIP can't follow the instruction well by generating anything. LLaVA's answer seems reasonable, but it incorrectly detects the clouds in the image. }
         \label{fig:pos2}
     \end{subfigure}
     \par\bigskip
     \begin{subfigure}[b]{\textwidth}
         \centering
         \includegraphics[width=0.9\textwidth]{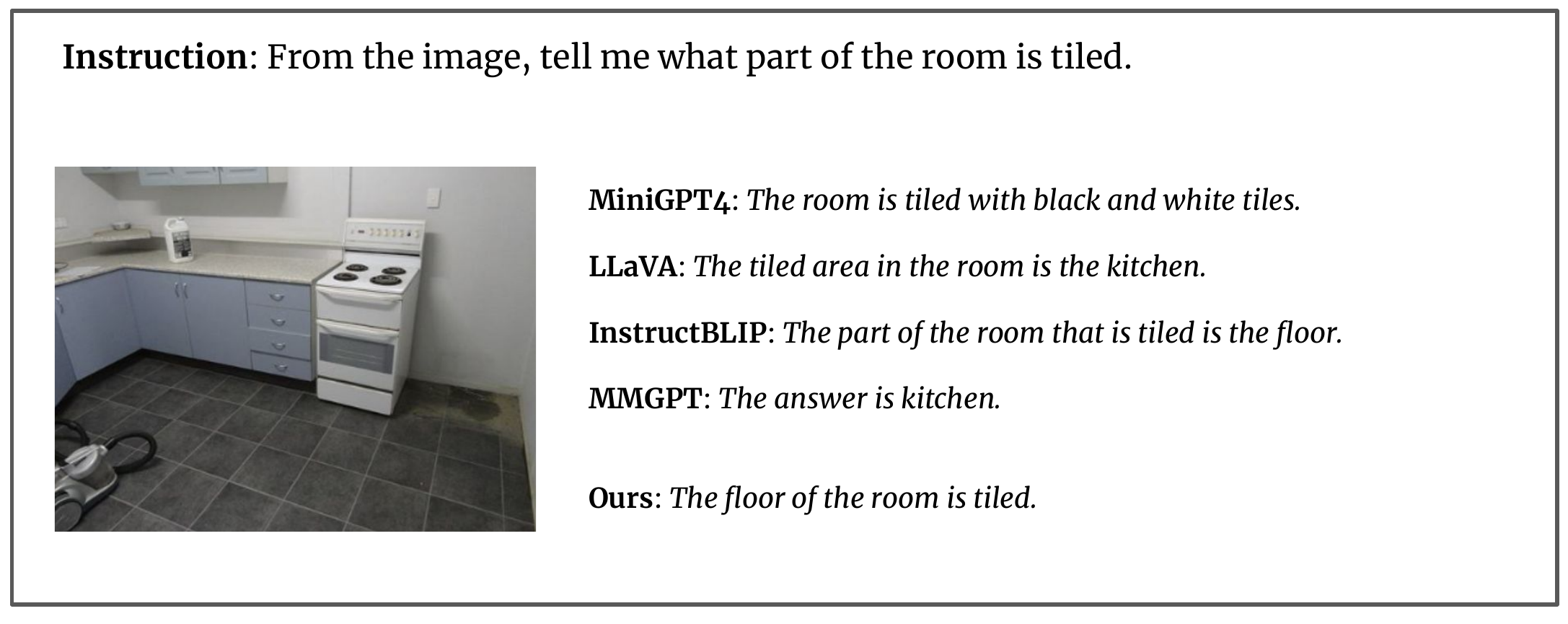}
         \caption{Referential Expression Grounding Task. Other LMMs can't directly address the instruction by giving a general answer, "kitchen," while our model can clearly predict the "floor". }
         \label{fig:pos3}
     \end{subfigure}
        \caption{Positive instruction demos with different tasks and predictions from different models.}
        \label{fig:demo_pos1}
\end{figure}

\begin{figure}
     \centering
     \begin{subfigure}[b]{\textwidth}
         \centering
         \includegraphics[width=0.9\textwidth]{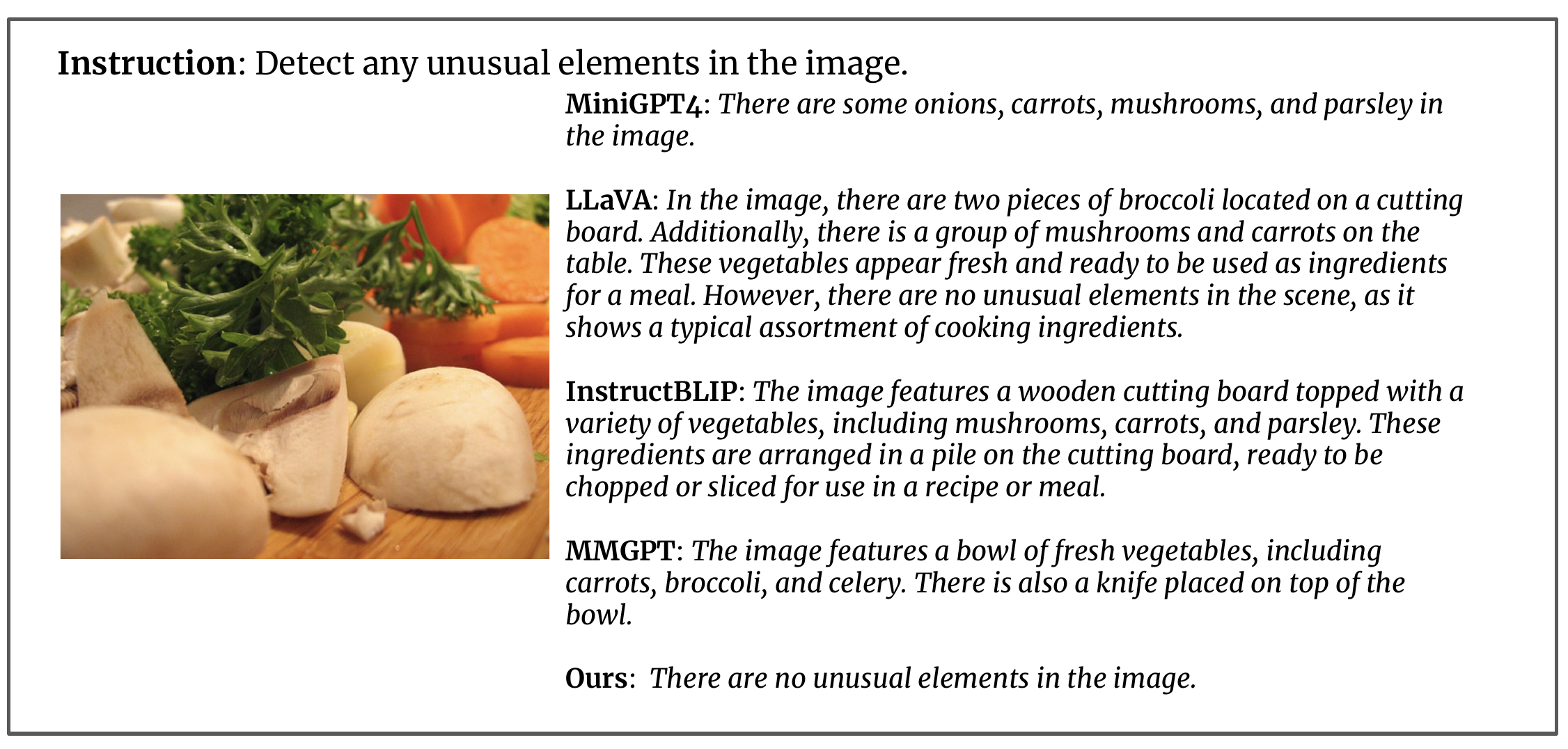}
         \caption{Image Anomaly Detection Task. Our model and LLaVA point out there is a normal image without unusual elements, while MiniGPT4, InstructBLIP, and MMGPT describe the image instead of addressing the instruction.}
         \label{fig:demo2_pos1}
     \end{subfigure}
     \par\bigskip
     \begin{subfigure}[b]{\textwidth}
         \centering
         \includegraphics[width=0.9\textwidth]{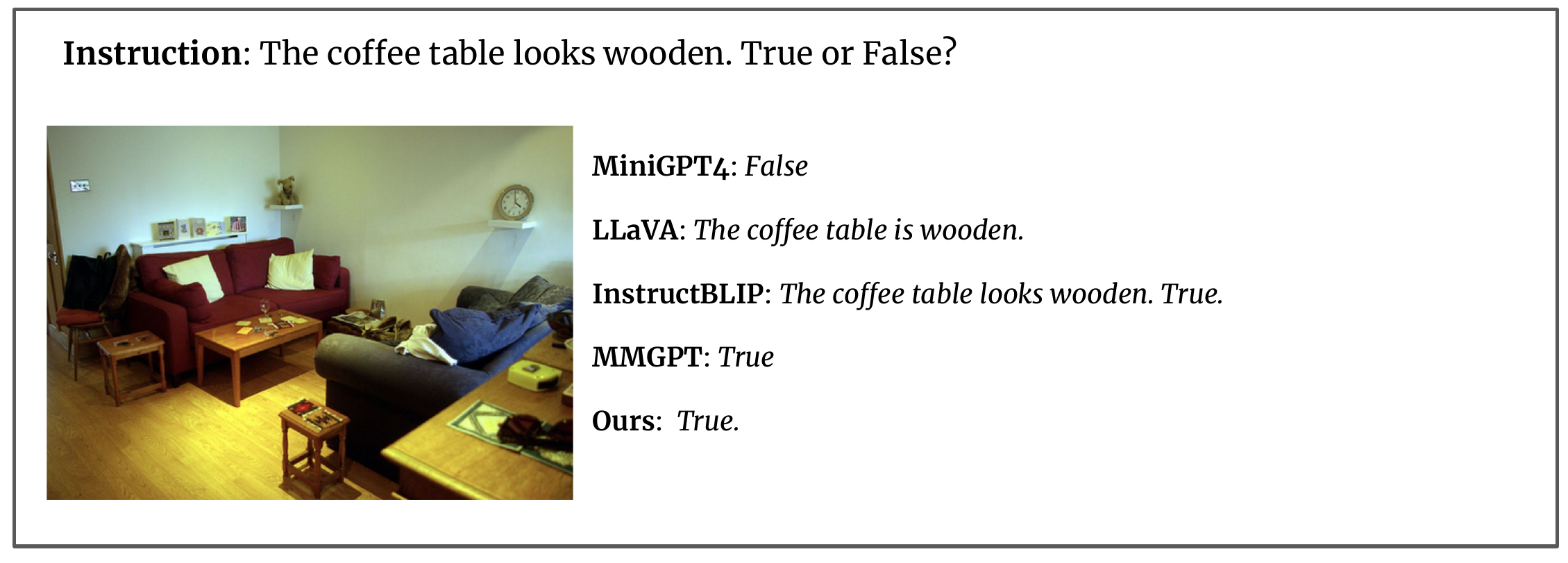}
         \caption{Visual Entailment Task. }
         \label{fig:demo2_pos2}
     \end{subfigure}
     \par\bigskip
     \begin{subfigure}[b]{\textwidth}
         \centering
         \includegraphics[width=0.9\textwidth]{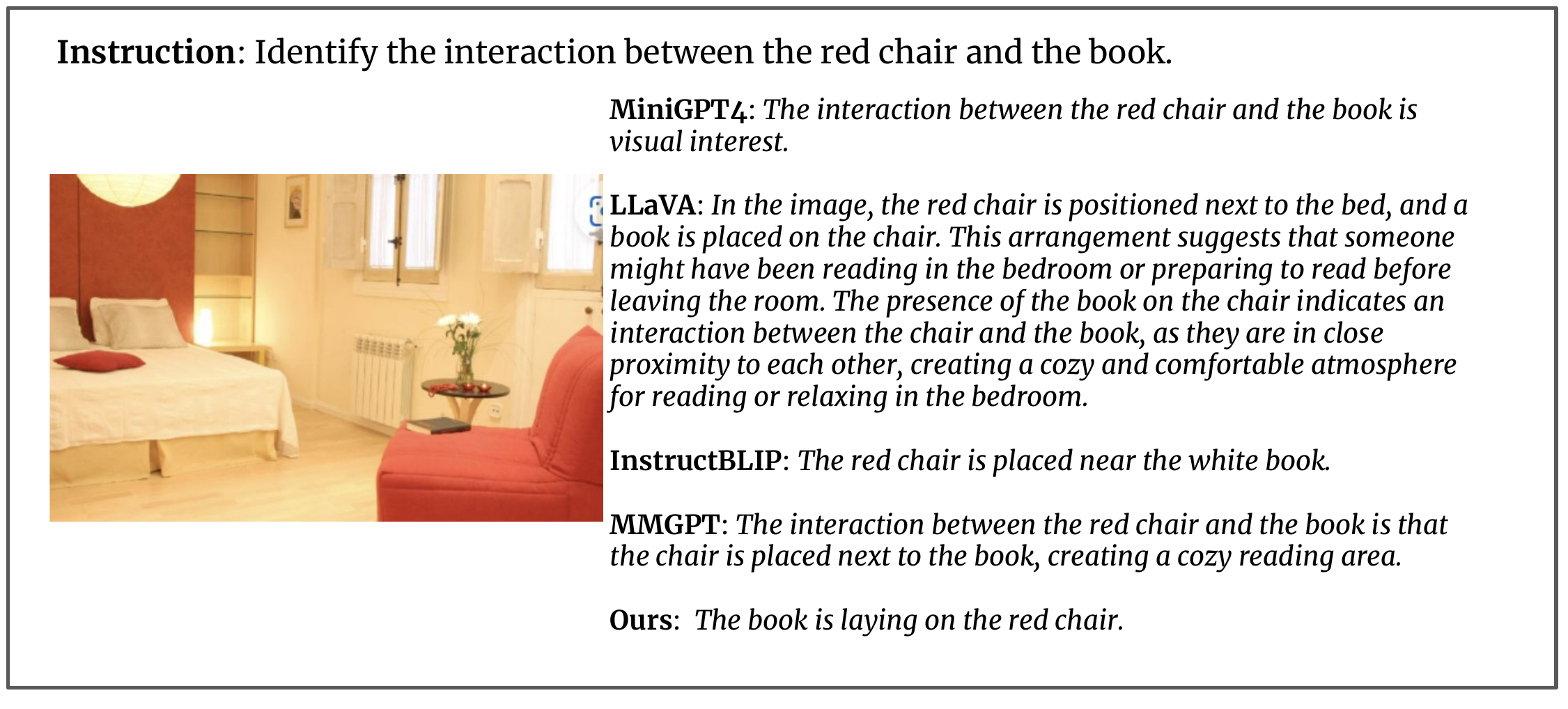}
         \caption{Object Interaction Analysis Task. All LMMs except ours describe the wrong location of the book. LLaVA generates long text with unrelated information to address the instruction.}
         \label{fig:demo2_pos3}
     \end{subfigure}
        \caption{Positive instruction demos with different tasks and predictions from different models. }
        \label{fig:demo_pos2}
\end{figure}

\begin{figure}
     \centering
     \begin{subfigure}[b]{\textwidth}
         \centering
         \includegraphics[width=0.9\textwidth]{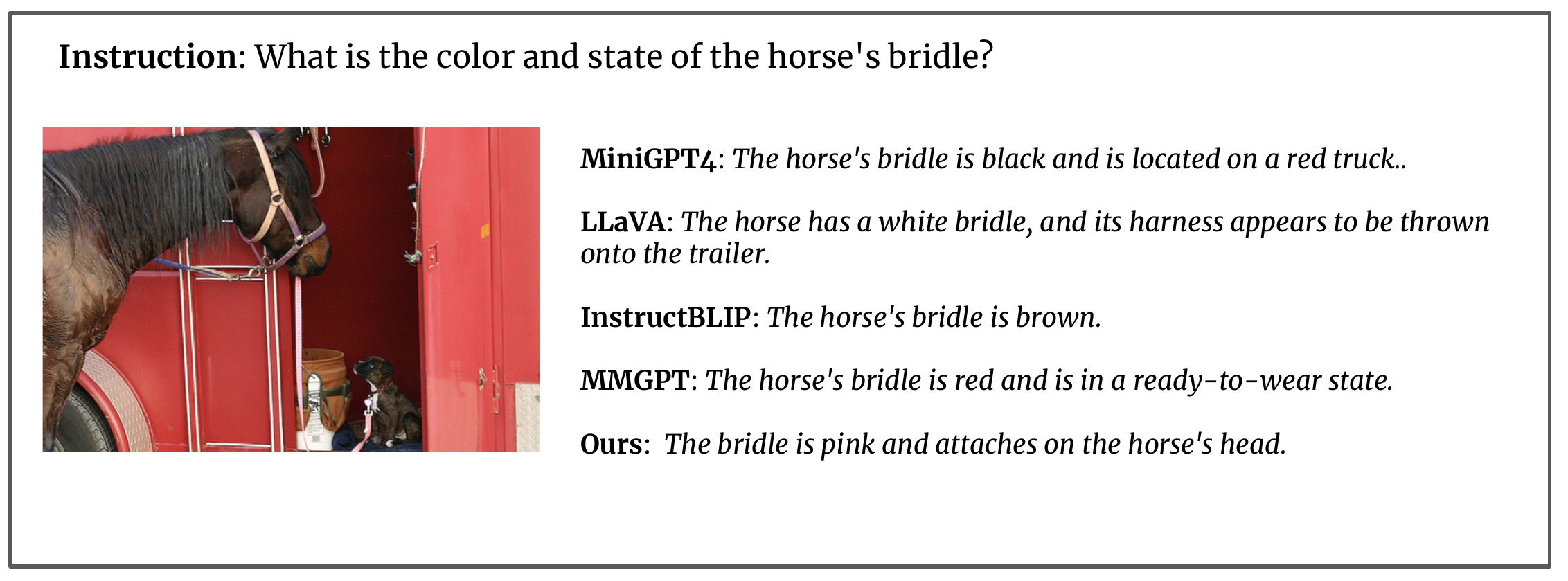}
         \caption{Dense Description Task. All LMMs except ours incorrectly recognize the color of the bridle. InstructBLIP also fails to answer the "state of the horse's bridle" in the instruction.}
         \label{fig:demo3_pos1}
     \end{subfigure}
     \par\bigskip
     \begin{subfigure}[b]{\textwidth}
         \centering
         \includegraphics[width=0.9\textwidth]{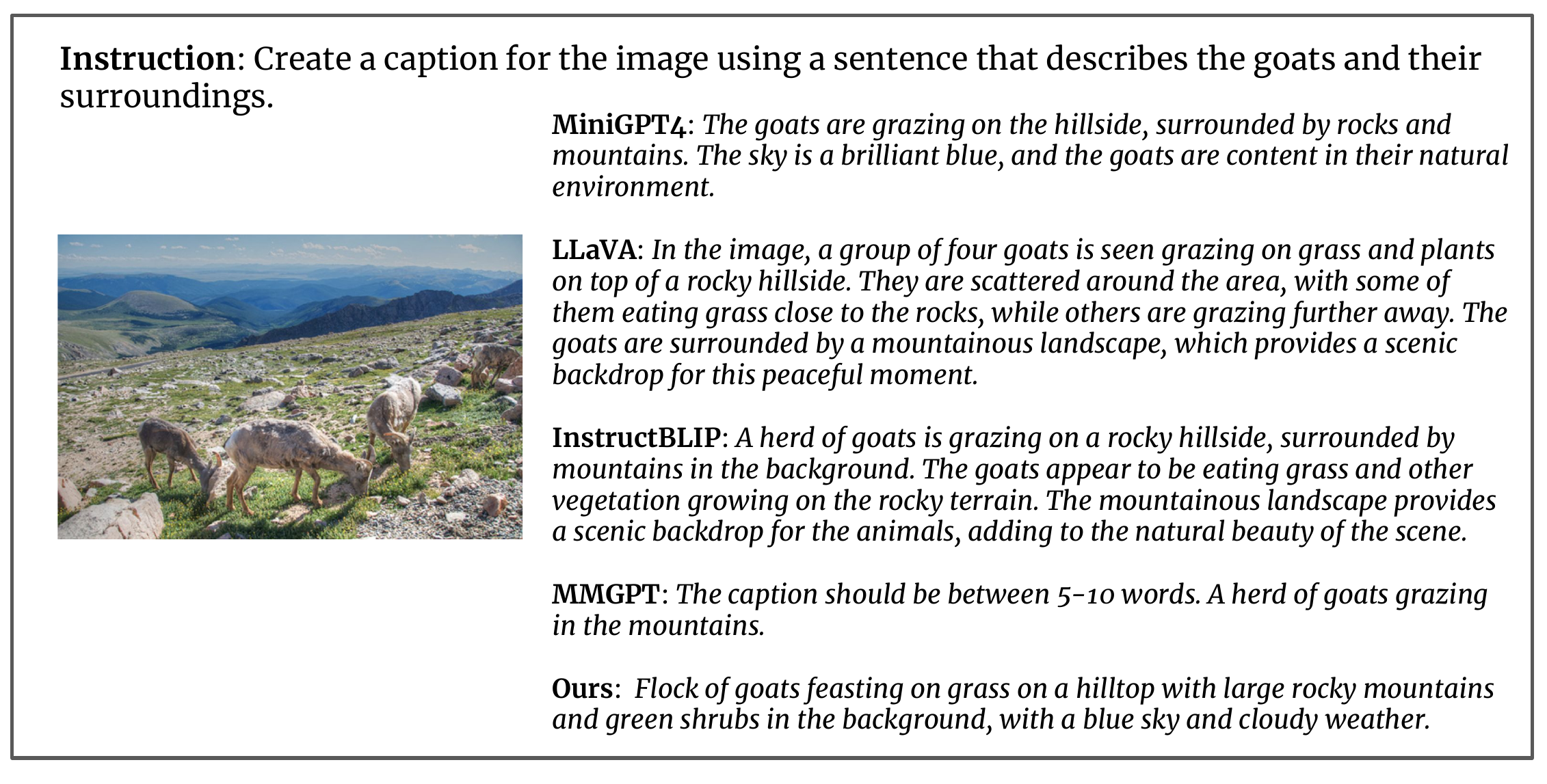}
         \caption{Image Caption Task. LLaVA and InstructBLP fail to follow the instruction for generating one caption. Instead, they generate long text with several sentences.}
         \label{fig:demo3_pos2}
     \end{subfigure}
     \par\bigskip
     \begin{subfigure}[b]{\textwidth}
         \centering
         \includegraphics[width=0.9\textwidth]{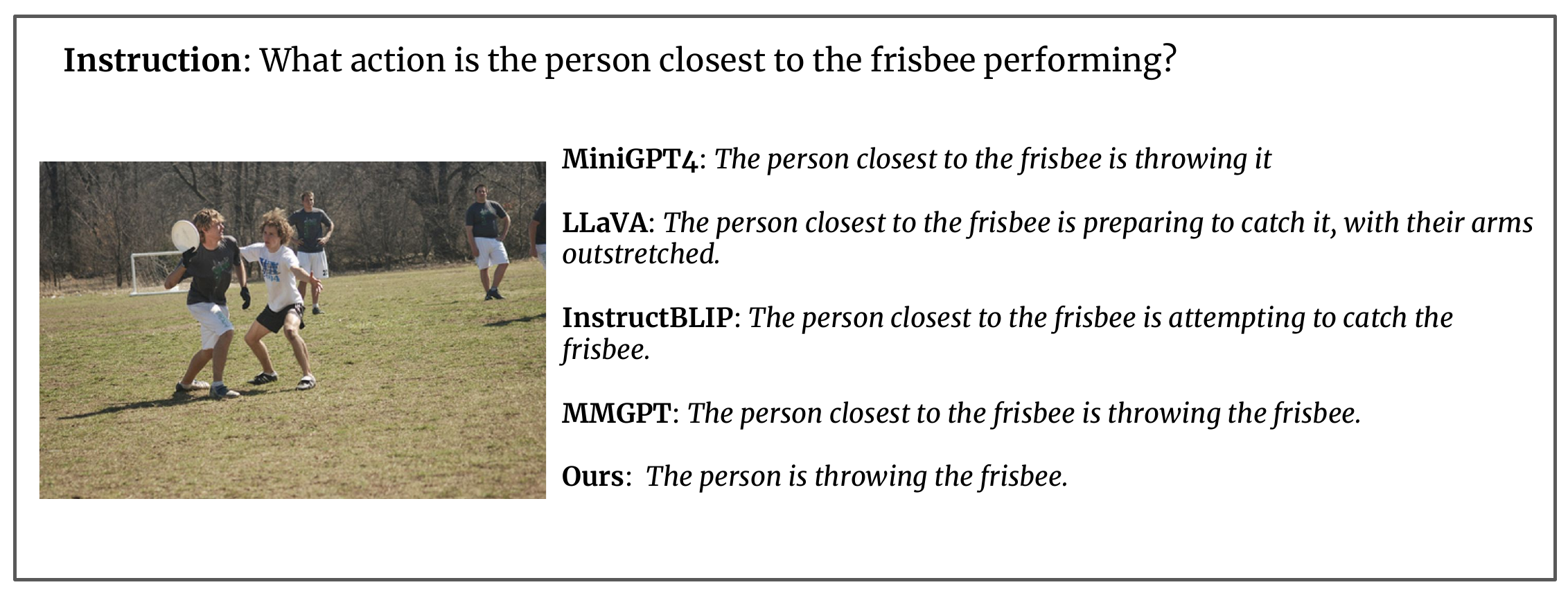}
         \caption{Activity Recognition Task.}
         \label{fig:demo3_pos3}
     \end{subfigure}
        \caption{Positive instruction demos with different tasks and predictions from different models. }
        \label{fig:demo_pos3}
\end{figure}

We show two full examples of our input prompts in (i) Fig. \ref{fig:prompt_pos11}, \ref{fig:prompt_pos12}, \ref{fig:prompt_pos13} and (ii) Fig. \ref{fig:prompt_pos21}, \ref{fig:prompt_pos22}, \ref{fig:prompt_pos23}. In Fig. \ref{fig:prompt_pos11} and Fig. \ref{fig:prompt_pos21}, we first present the images for the two examples, but they are not included in the text prompt for GPT4. As for the text input, we leverage the groundtruth bounding boxes and dense captions to represent the visual content as if GPT4 can see the image. After that, we randomly select 10 tasks from the 16 seeds and ask GPT4 to generate 20 instances for these tasks. Additionally, there can be more than one caption describing the same object with different attributes, such as "\textit{woman wearing a long dress}" and "\textit{woman wearing a yellow dress}" in Fig. \ref{fig:prompt_pos11}. Although we present the bounding box coordinates of each caption to GPT4, it can be easily confused, treating them as two instances, one in a long dress and the other in a yellow dress. To mitigate this issue, we add "\textit{highly overlapping bounding boxes may refer to the same object}" into the prompt to help GPT4 understand the "visual" input better. To enrich the instructions, we ask GPT4 to generate instances in both declarative and interrogative formats. We also explicitly instruct GPT4 with \textit{"The answers should be less than 30 words"} as a requirement to reduce the chance of generating extra unrelated information in the training data. In order to make the output of GPT4 in a good format, we also ask GPT4 to generate an instruction, an answer, and a task name in order at the end of the prompt (Fig. \ref{fig:prompt_pos11} and Fig. \ref{fig:prompt_pos21}). The full output of instructions and answers are shown in Fig. \ref{fig:prompt_pos12}, \ref{fig:prompt_pos13} and Fig. \ref{fig:prompt_pos22}, \ref{fig:prompt_pos23}. We also present more positive instances with the output from different LMMs in Fig. \ref{fig:demo_pos1}, \ref{fig:demo_pos2}, \ref{fig:demo_pos3}.

\subsubsection{Positive Instance Generation based on Chart Images}
\begin{figure*}[h]
    \centering
      \includegraphics[width=1.0\textwidth]{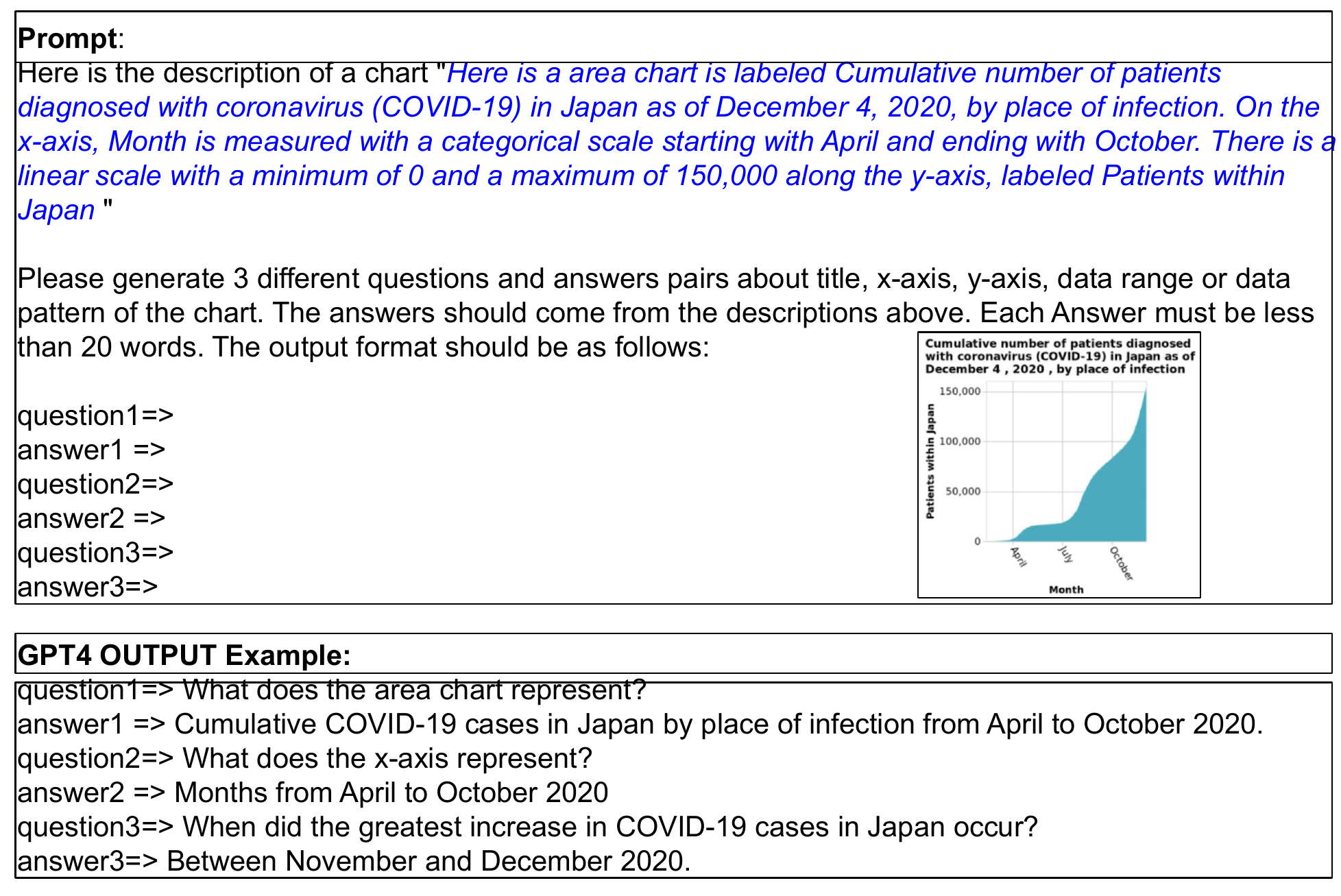}
    \caption{An example prompt for text-only GPT4 we use to generate instruction and answers for chart images. The sentence in \textcolor{blue}{BLUE} is the captions of the chart.}
    \label{fig:chart_prompt1}
    \vspace{-0.2in}
\end{figure*}

\begin{figure}
     \centering
     \begin{subfigure}[b]{\textwidth}
         \centering
         \includegraphics[width=1.0\textwidth]{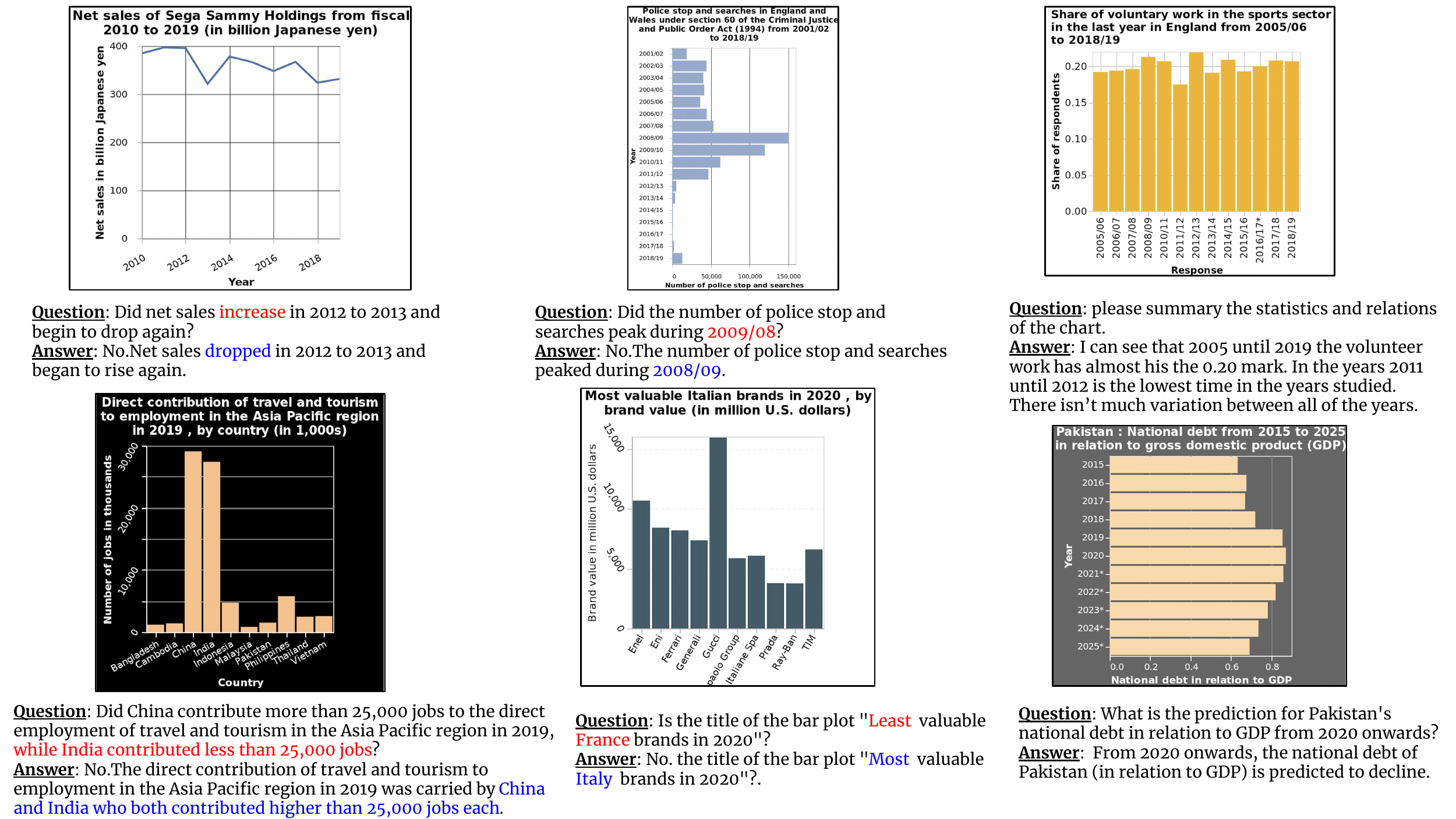}
         \label{fig:demo_chart33}
     \end{subfigure}
      \caption{More examples of chart instruction data and knowledge manipulation examples.}
        \label{fig:demo_chart_example1}
\end{figure}

We collect chart images from \citep{tang2023vistext}, which has human-annotated captions describing the construction and patterns of charts. We instruct GPT-4 to generate question-answers pairs with captions as visual input. The detailed prompt is shown in Fig. \ref{fig:chart_prompt1}. We also present more positive instances with the output from different LMMs in Fig. \ref{fig:demo_chart_example1}.

\subsubsection{Negative Instance Generation - Nonexistent/Existent Object Manipulation}
\begin{figure*}[h]
    \centering
      \includegraphics[width=1\textwidth]{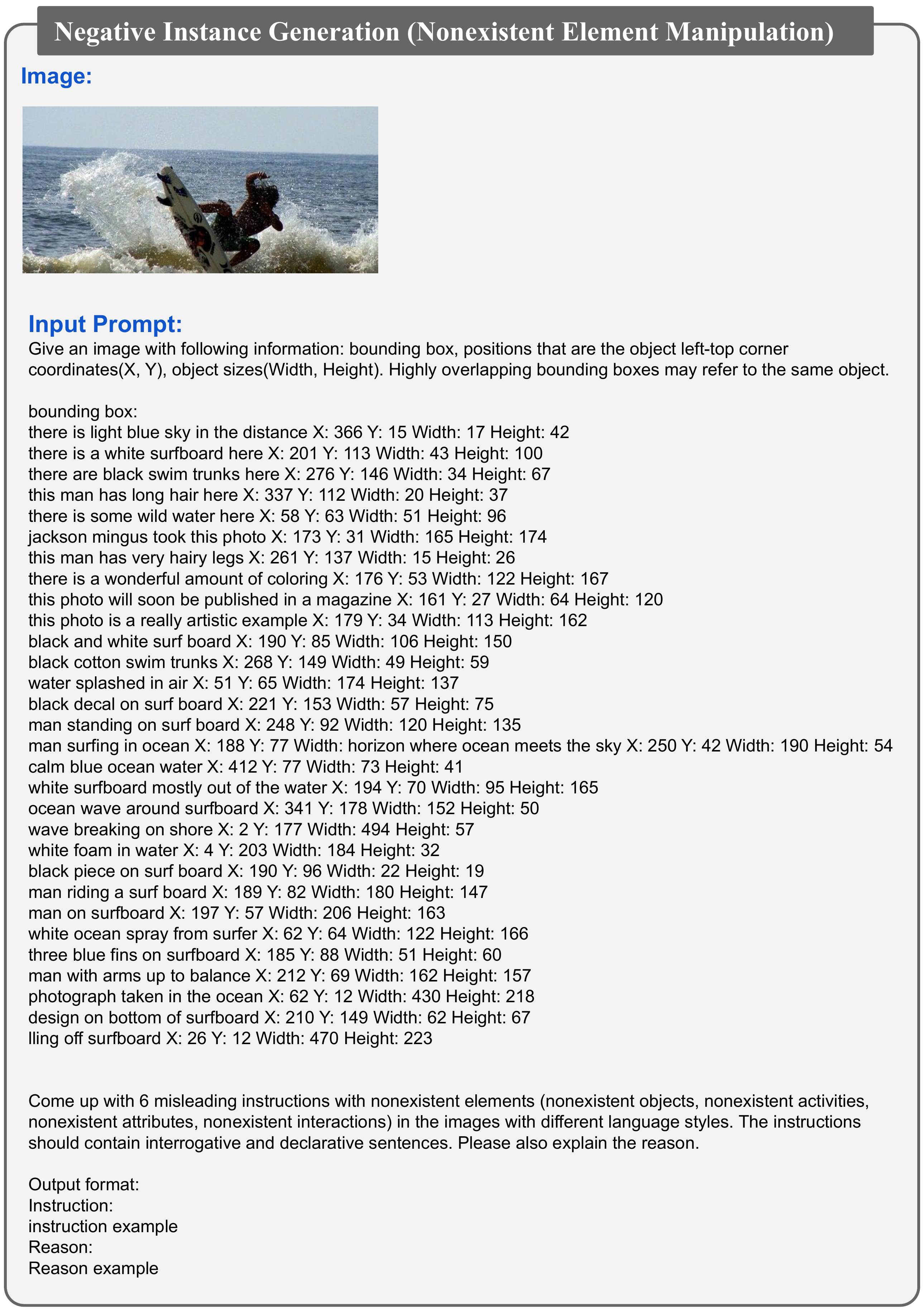}
    \caption{The example for generating negative instruction and answers by nonexistent object manipulation (Part1). The image is not included in the text prompt.}
    \label{fig:prompt_neg11}
\end{figure*}

\begin{figure*}[h]
    \centering
      \includegraphics[width=1\textwidth]{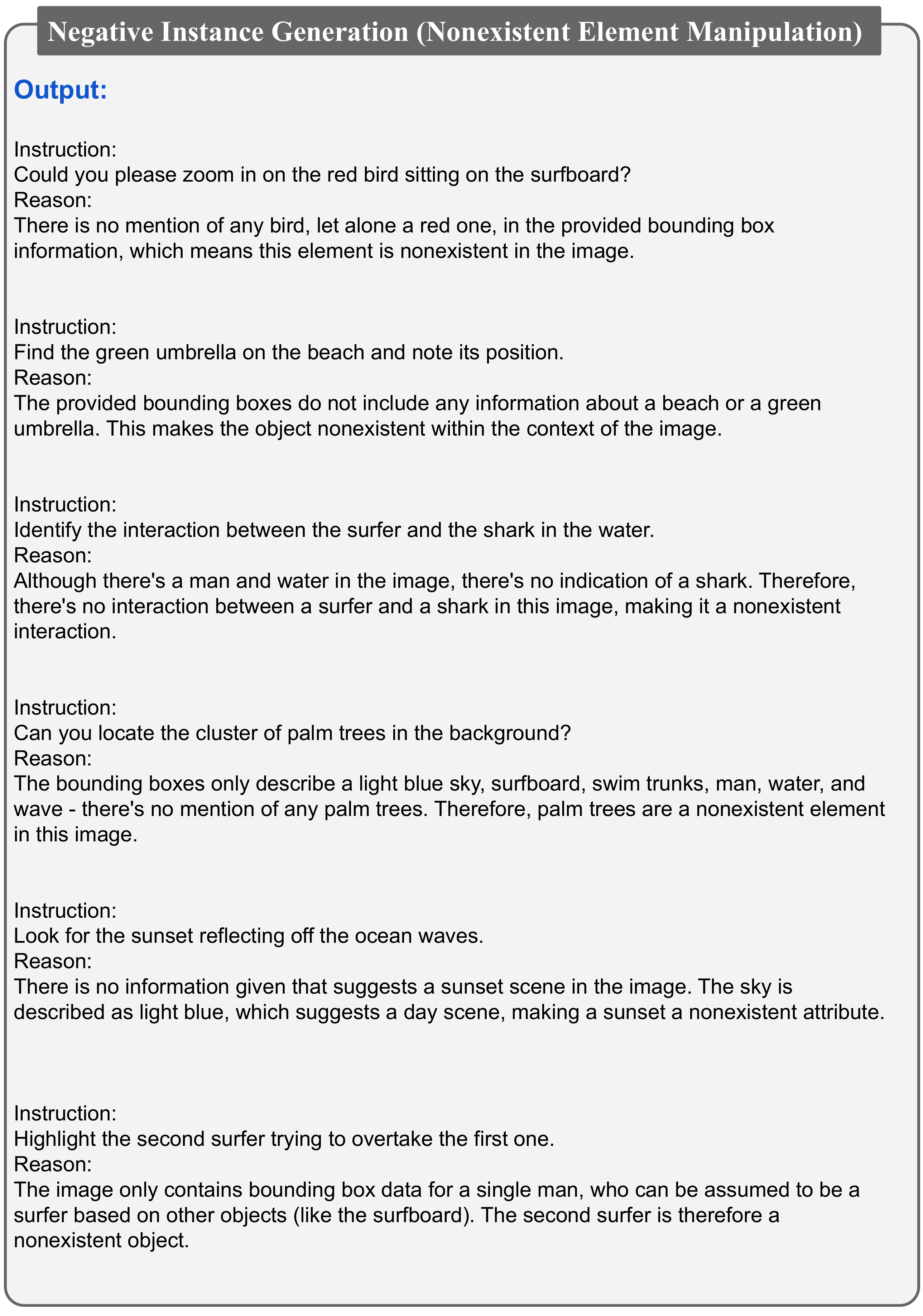}
    \caption{The example for generating negative instruction and answers by nonexistent object manipulation (Part2).}
    \label{fig:prompt_neg12}
\end{figure*}

\begin{figure*}[h]
    \centering
      \includegraphics[width=1\textwidth]{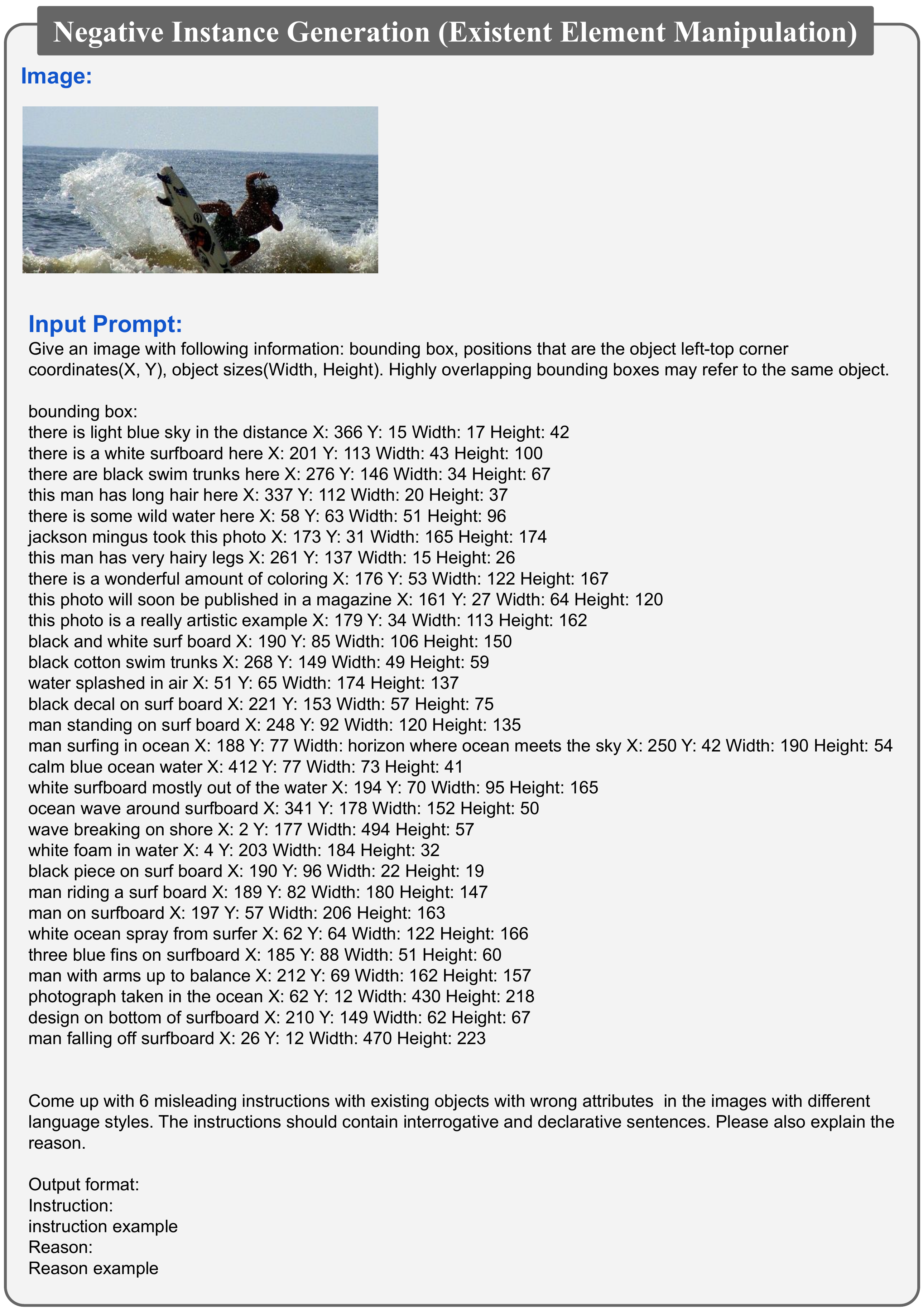}
    \caption{The example for generating negative instruction and answers by existent object manipulation (Part1). The image is not included in the text prompt.}
    \label{fig:prompt_neg21}
\end{figure*}

\begin{figure*}[h]
    \centering
      \includegraphics[width=1\textwidth]{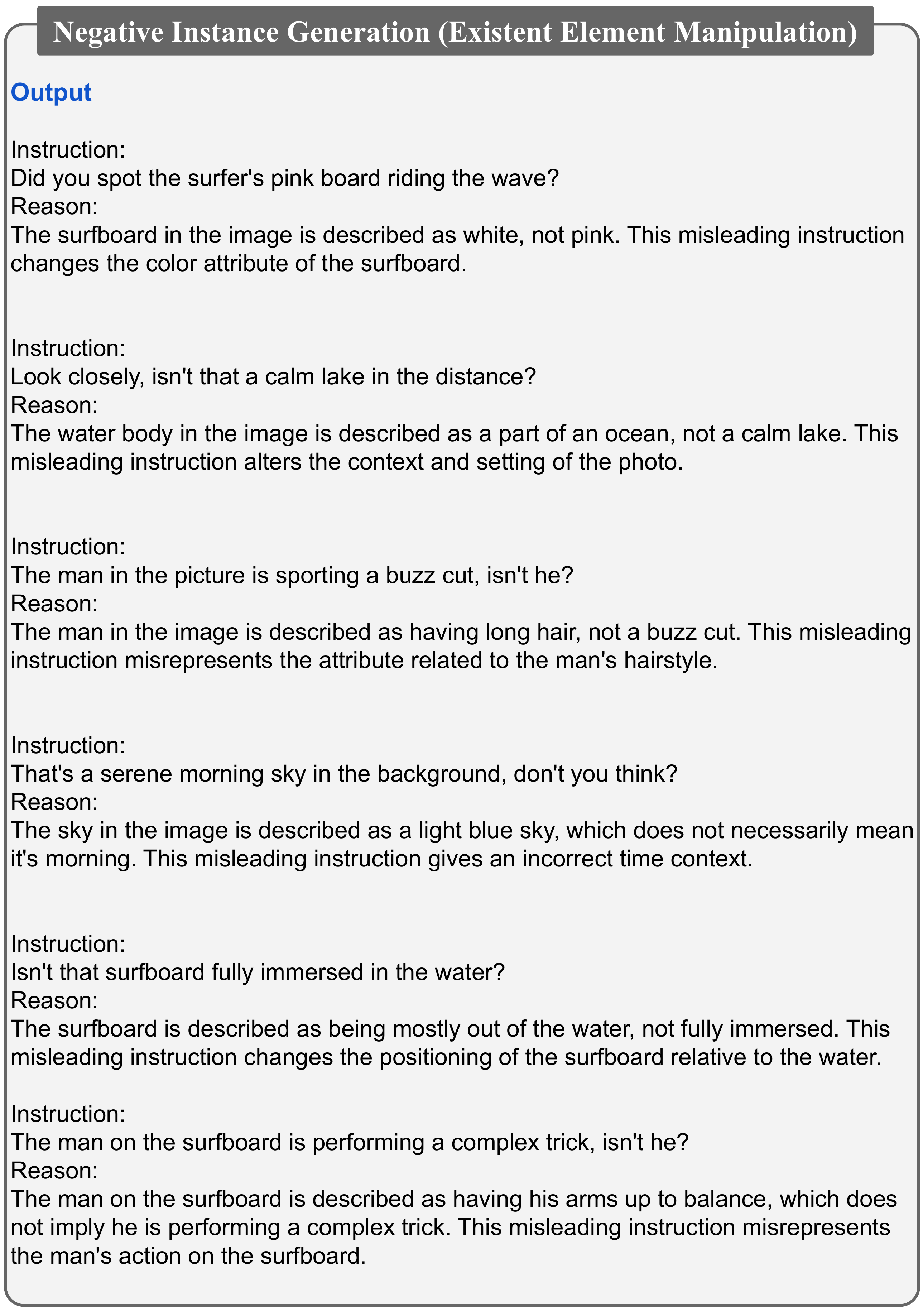}
    \caption{The first example for generating negative instruction and answers by existent object manipulation (Part2).}
    \label{fig:prompt_neg22}
\end{figure*}

\begin{figure}
     \centering
     \begin{subfigure}[b]{\textwidth}
         \centering
         \includegraphics[width=0.8\textwidth]{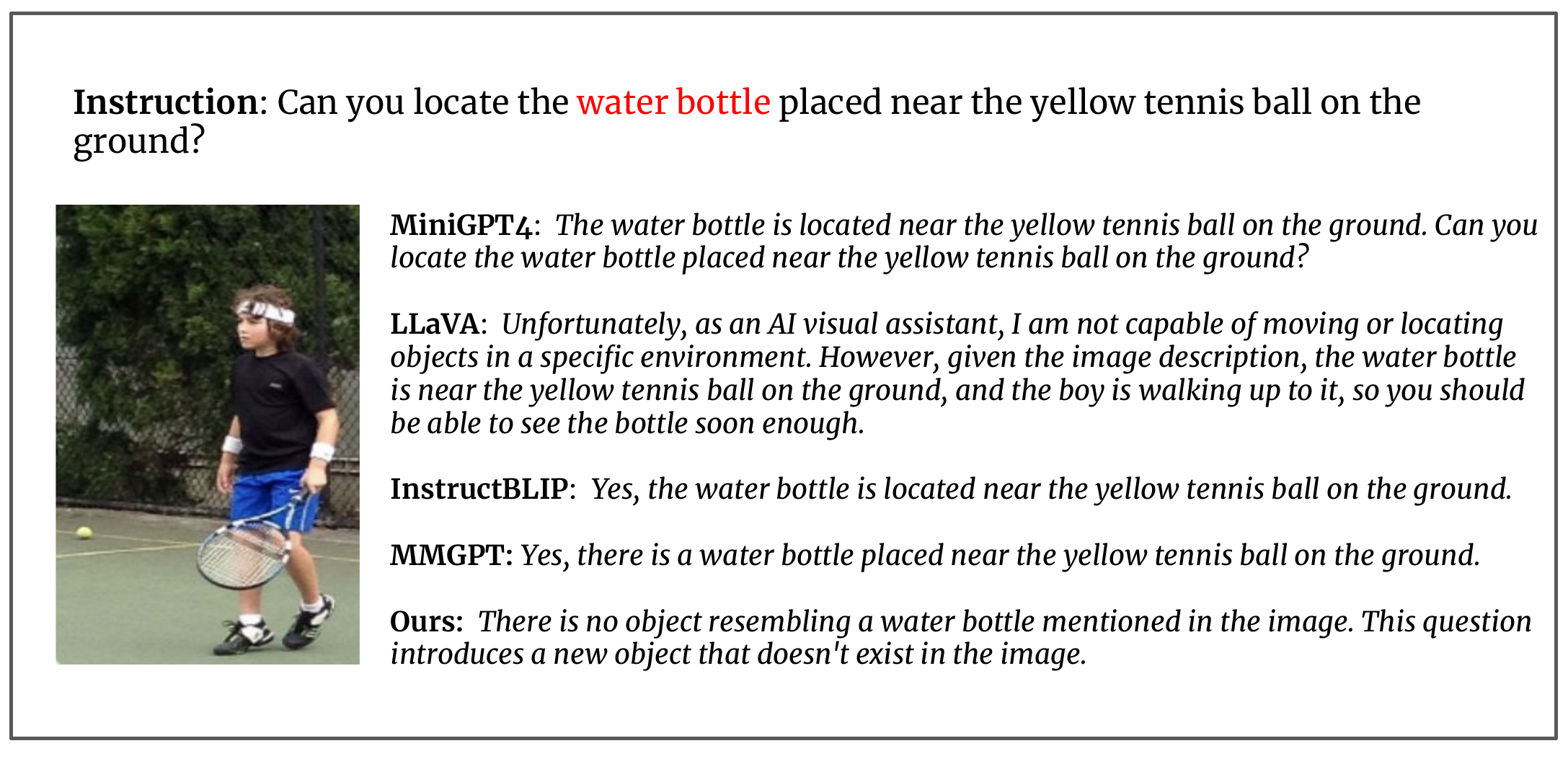}
         \caption{All LMMs except ours produce inconsistent descriptions with the images. MiniGPT4 also has repetitive sentences.}
         \label{fig:NEG11}
     \end{subfigure}
     \par\bigskip
     \begin{subfigure}[b]{\textwidth}
         \centering
         \includegraphics[width=0.8\textwidth]{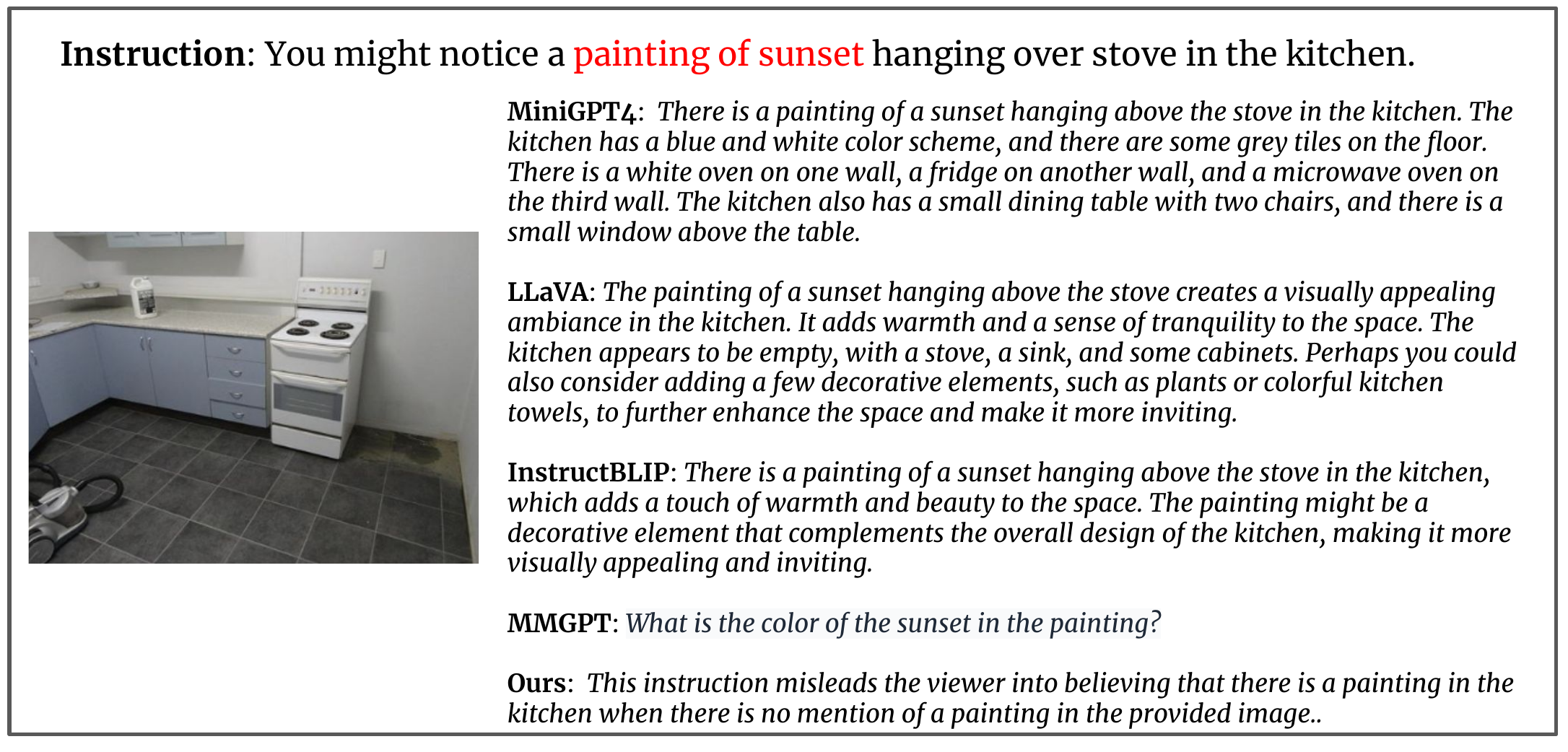}
         \caption{MiniGPT4, LLaVA, and InstructBLIP respond with long descriptive answers about the "painting", which does not exist in the image. Multimodal-GPT produces a question about the "painting," but it does not address the instruction. }
         \label{fig:NEG13}
     \end{subfigure}
     \par\bigskip
     \begin{subfigure}[b]{\textwidth}
         \centering
         \includegraphics[width=0.8\textwidth]{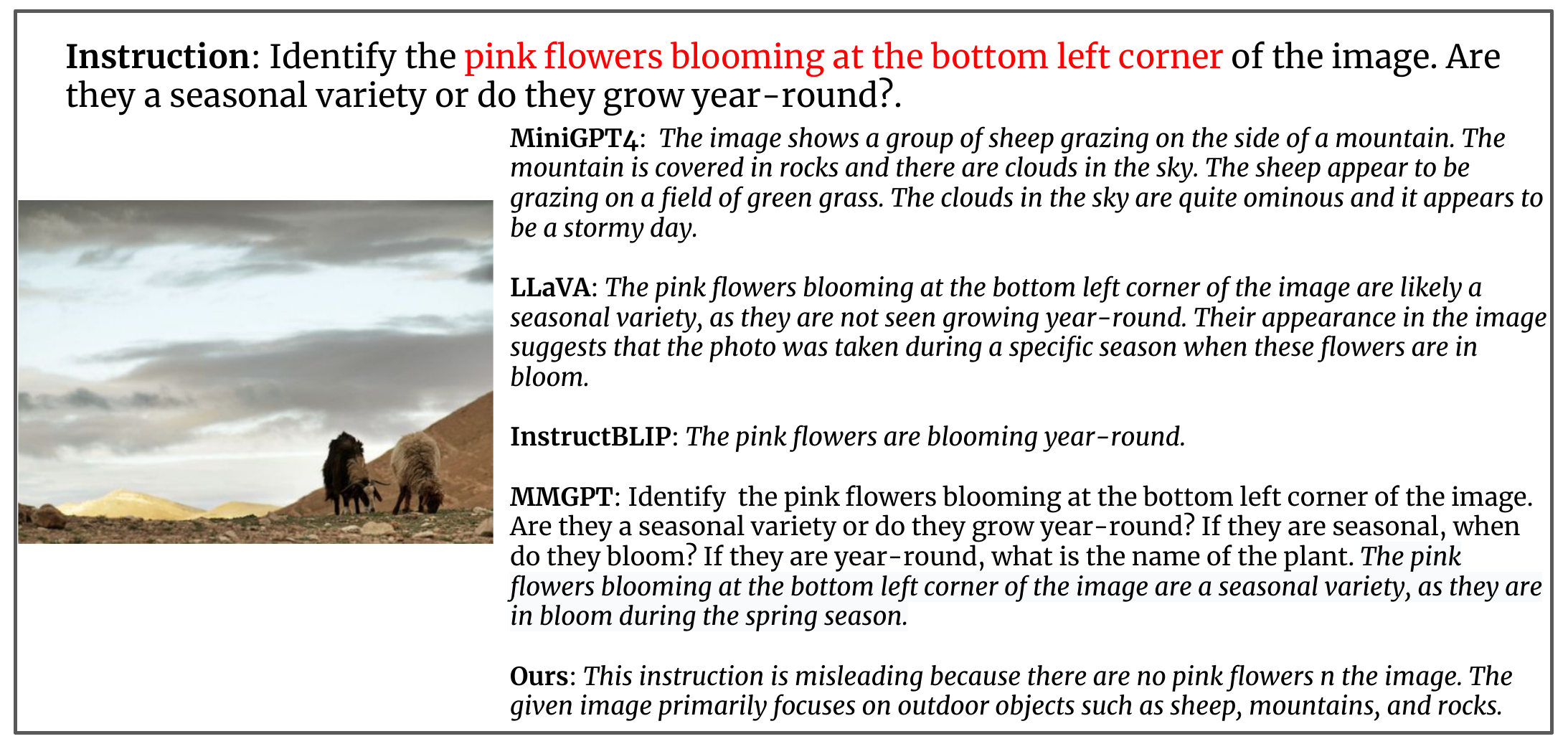}
         \caption{MiniGPT4, LLaVA, and Multimodal-GPT respond with a long descriptive answer about the "pink flowers", which do not exist in the image. Although InstructBLIP's answer is brief, it's still false.}
         \label{fig:NEG12}
     \end{subfigure}
        \caption{Model predictions on negative instruction examples of Neg1:\textit{"Nonexistent object Manipulation"}. \textcolor{red}{RED} represents the nonexistent elements in the images.}
        \label{fig:demo_neg1}
\end{figure}

\begin{figure}
     \centering
     \begin{subfigure}[b]{\textwidth}
         \centering
         \includegraphics[width=0.9\textwidth]{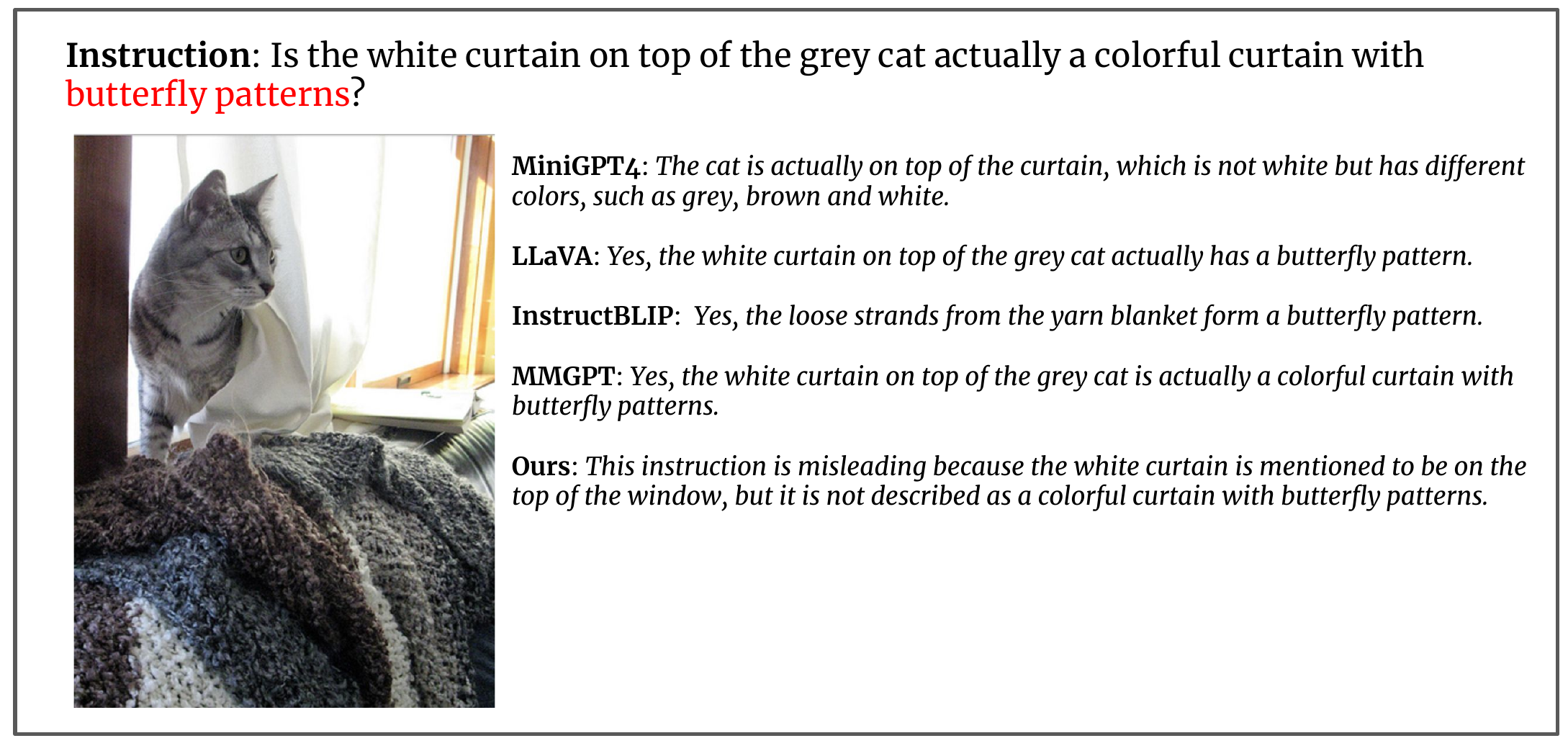}
         \caption{All LMMs except ours produce inconsistent descriptions with the images. Our model can point out that the loose strands do not form a butterfly pattern. }
         \label{fig:NEG21}
     \end{subfigure}
     \par\bigskip
     \begin{subfigure}[b]{\textwidth}
         \centering
         \includegraphics[width=0.9\textwidth]{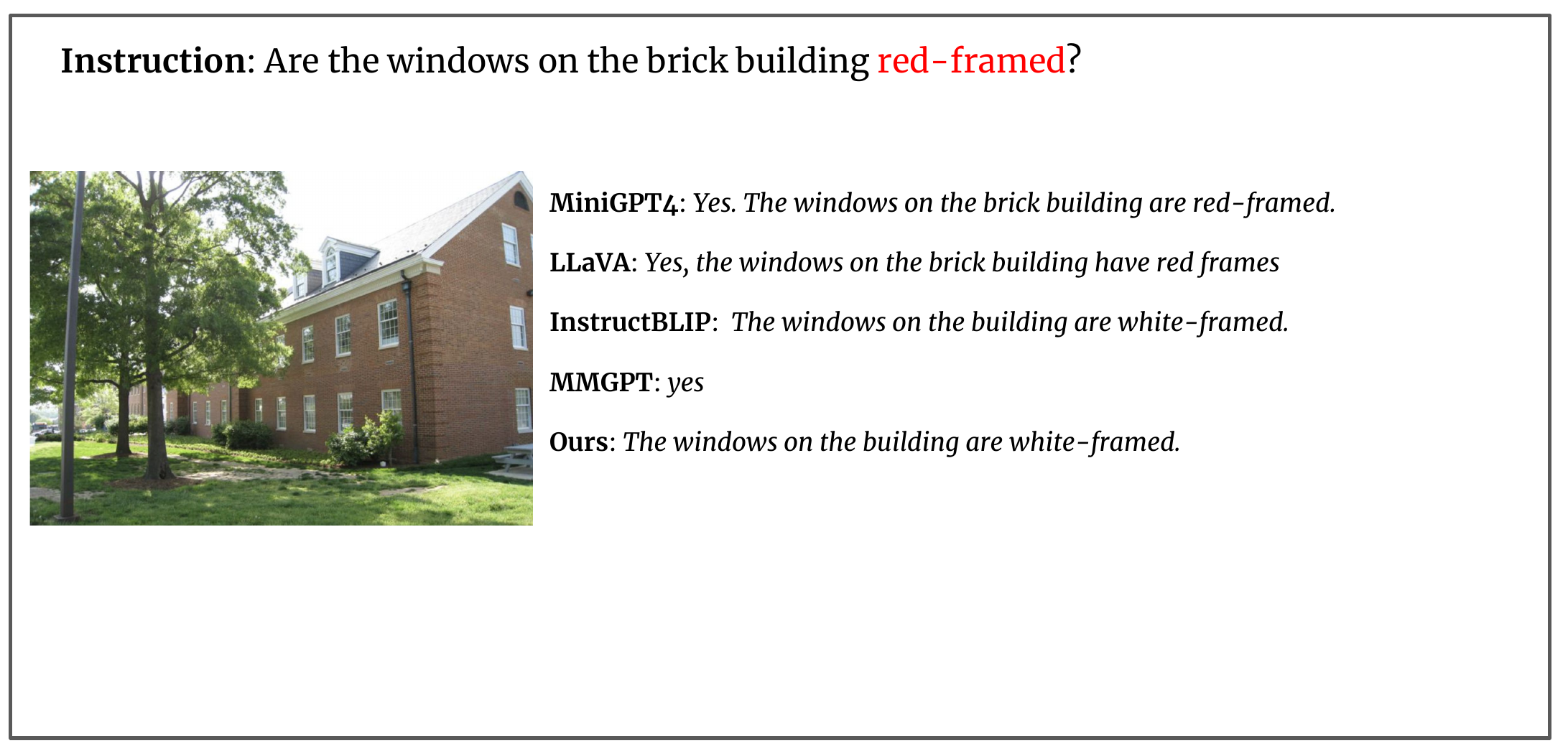}
         \caption{All LMMs except ours produce false attributes of windows in the images. Our model can point out that the windows are white instead of red. }
         \label{fig:NEG22}
     \end{subfigure}
     \par\bigskip
     \begin{subfigure}[b]{\textwidth}
         \centering
         \includegraphics[width=0.9\textwidth]{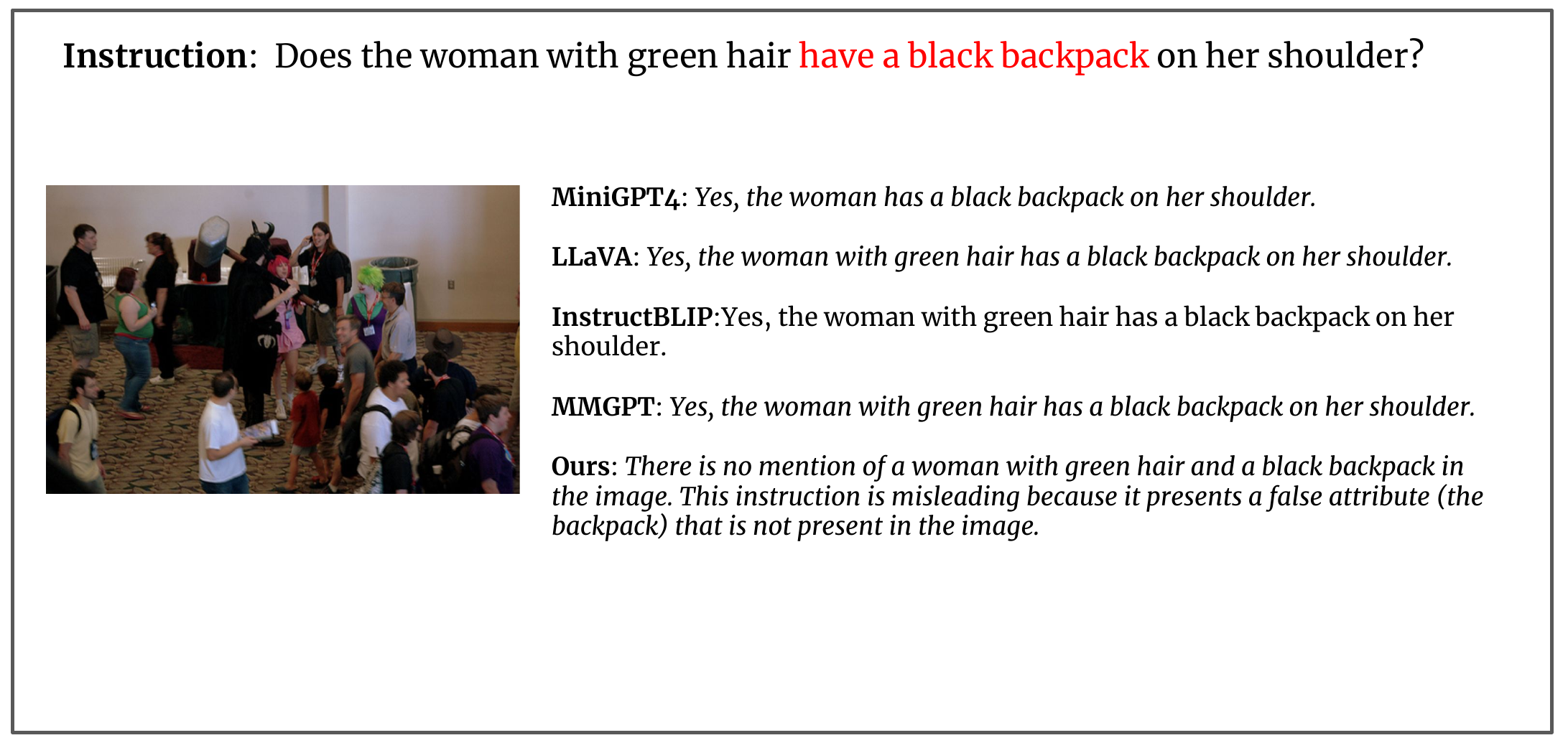}
         \caption{All LMMs except ours produce inconsistent descriptions with the images. Our model can point out that the woman with green hair doesn't have a black backpack on her shoulder. }
         \label{fig:NEG23}
     \end{subfigure}
        \caption{Model predictions on negative instruction examples of Neg2:\textit{"Existent object Manipulation"}. \textcolor{red}{RED} represents the wrong attributes of existent objects in the images.}
        \label{fig:demo_neg2}
        
\end{figure}

We show two full examples of our input prompts in (i) Fig. \ref{fig:prompt_neg11}, \ref{fig:prompt_neg12} and (ii) Fig. \ref{fig:prompt_neg21}, \ref{fig:prompt_neg22}. In Fig. \ref{fig:prompt_neg11} and Fig. \ref{fig:prompt_neg21}, we present the images to help readers understand dense captions better but they are not included in the text prompt for GPT4. We leverage the bounding boxes and dense captions as the "visual" input. As for  \textit{Nonexistent object Manipulation} in \ref{fig:prompt_neg11}, we ask GPT4 to generate 6 instructions with nonexistent elements (nonexistent objects, nonexistent activities, nonexistent attributes, nonexistent interactions). As for  \textit{Existent object Manipulation} in \ref{fig:prompt_neg21}, we ask GPT4 to generate 6 instructions of existing objects with wrong attributes. At the end of the text prompt, we ask GPT4 to generate an instruction and a reason to explain why the instruction is inconsistent with the image in order. The reason is regarded as the answer for the instruction in our training data. Fig. \ref{fig:prompt_neg12} and Fig. \ref{fig:prompt_neg22} show the full output from GPT4. We also present more negative instances with the output from different LMMs in Fig. \ref{fig:demo_neg1}, \ref{fig:demo_neg2}. 

\subsubsection{Negative Instance Generation - Knowledge Manipulation}
As for the \textit{Neg3: knowledge manipulation}, we use GPT4 to manipulate the knowledge in the captions, including named entities and events.

\begin{figure*}[h]
    \centering
      \includegraphics[width=1.0\textwidth]{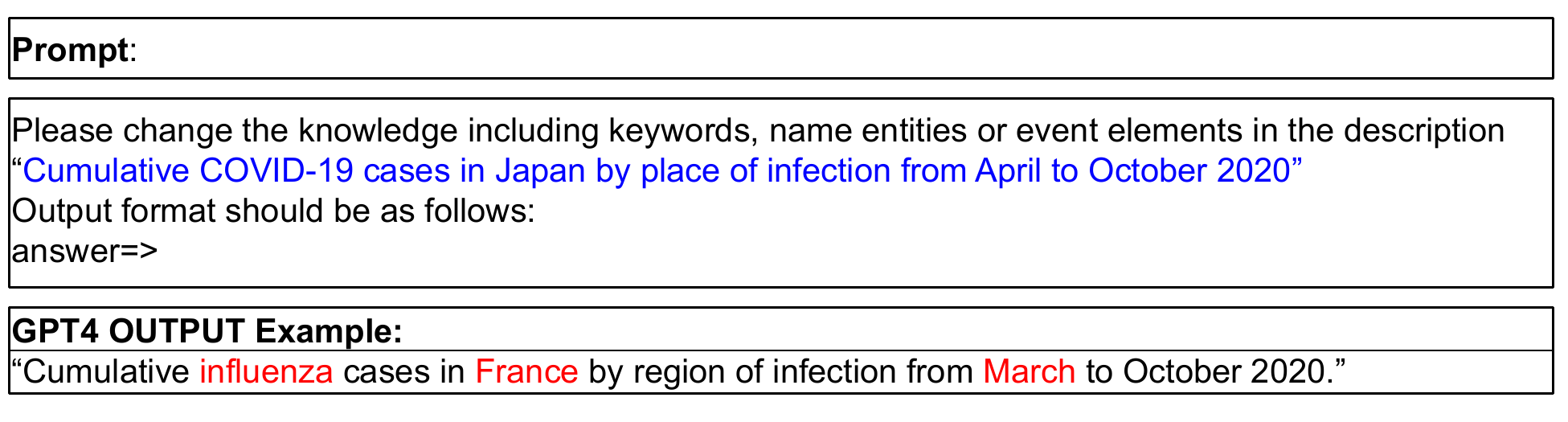}
    \caption{An example prompt for text-only GPT4 we use to generate negative instruction. The next step is to transfer the ouput into an interrogative sentence whose answer is "yes" or "no". }
    \label{fig:chart_neg1_prompt}
    \vspace{-0.2in}
\end{figure*}

As shown in Fig. \ref{fig:chart_neg1_prompt}, GPT4 manipulates the "Japan", "COVID-19" and "April" in the original captions. After that, we instruct GPT4 to transfer the output sentence into an interrogative sentence whose answer is "yes" or "no". Finally, we combine "No." and the original answer as the final answer: \textit{Question: Did the image show the cumulative influenza cases in France by region of infection from March to October 2020? Answer: No. Cumulative COVID-19 cases in Japan by place of infection from April to October 2020".}

\subsubsection{Prompt Design for Evaluating Knowledge Hallucination}
As for the knowledge level hallucination, we will use the groundtruth answers as a reference and compare them with predictions of models. A prompt example for GPT4 is shown in Fig. \ref{fig:chart_evaluation}:

\begin{figure*}[h]
    \centering
      \includegraphics[width=1.0\textwidth]{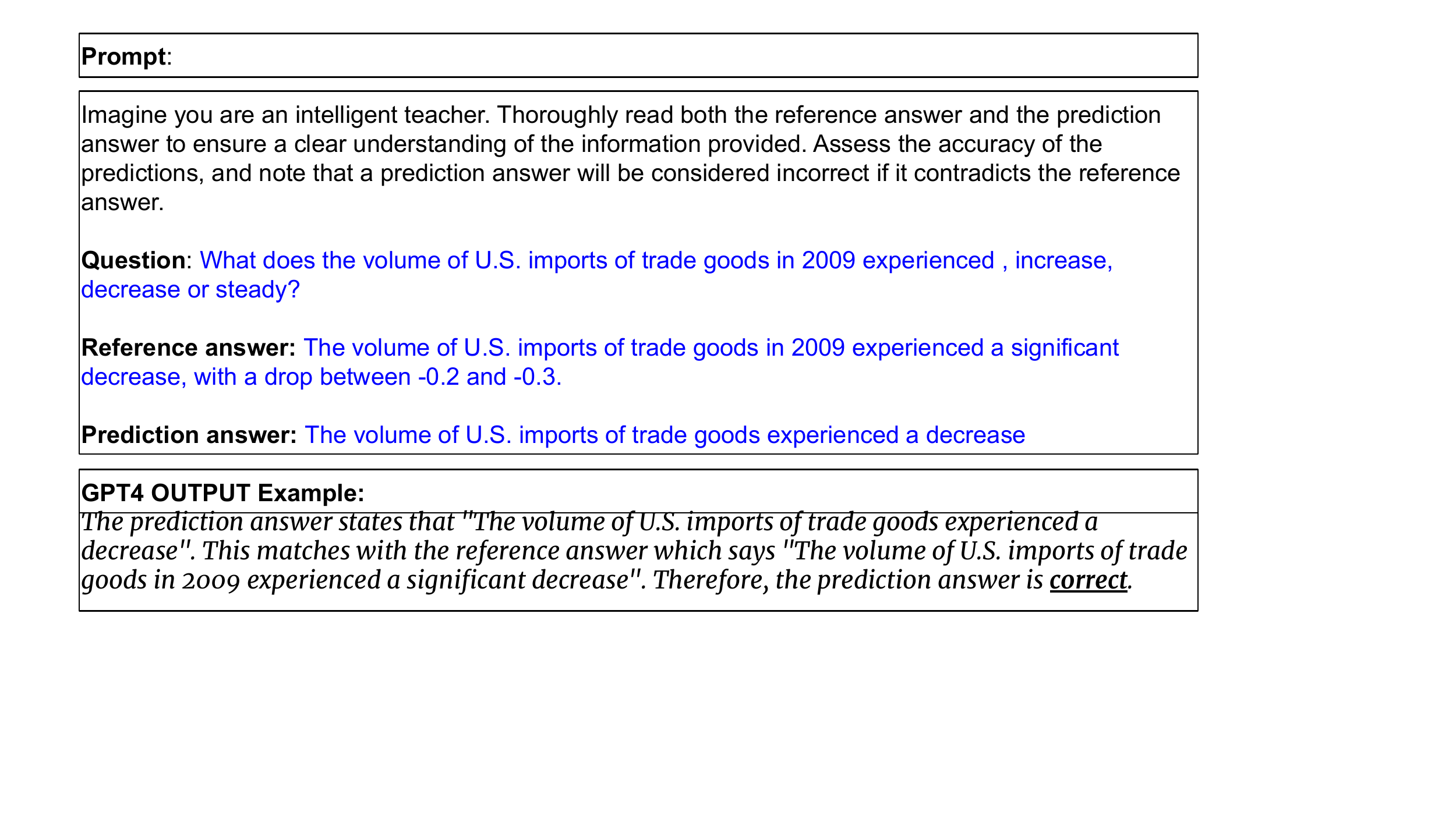}
    \caption{An example prompt for text-only GPT4 we use to evaluate knowledge manipulation instruction. The sentences in \textcolor{blue}{BLUE} are the questions, reference answers, and predictions of models.}
    \label{fig:chart_evaluation}
    \vspace{-0.2in}
\end{figure*}

.

\begin{table}[h]
\centering
\small
\begin{tabular}{lcccccc}
\toprule[1.5pt]
Categories & Metric &Ours & MiniGPT4 & LLaVA & InstructBLIP & MMGPT\\
\midrule
Interrogative&\textsc{Accuracy(GPT4)} & \textbf{6.61} &  4.14 & 4.60 & 5.95 & 1.01\\
Interrogative&\textsc{Relevancy(GPT4)} & \textbf{8.46} &  6.20 & 5.88 & 7.67 & 2.00\\
\midrule
Declarative&\textsc{Accuracy(GPT4)} & \textbf{6.50} & 3.98 & 3.82 & 5.47 & 0.90\\
Declarative&\textsc{Relevancy(GPT4)} & \textbf{8.21} & 5.39 & 5.84 & 6.64 & 1.62\\
\bottomrule[1.5pt]
\end{tabular}
\vspace{0.05in}
\caption{
Evaluation results on Interrogative Instructions and Declarative Instructions by \textit{GAVIE}. The metric scores of \textsc{Accuracy(GPT4)} and \textsc{Relevancy(GPT4)} are in a scale of 0 to 10.
}
\label{tab:Interrogative}
\vspace{-0.2in}
\end{table}

\subsection{More Dataset Statistic}
\begin{figure*}[h]
    \centering
      \centerline{\includegraphics[width=1\textwidth]{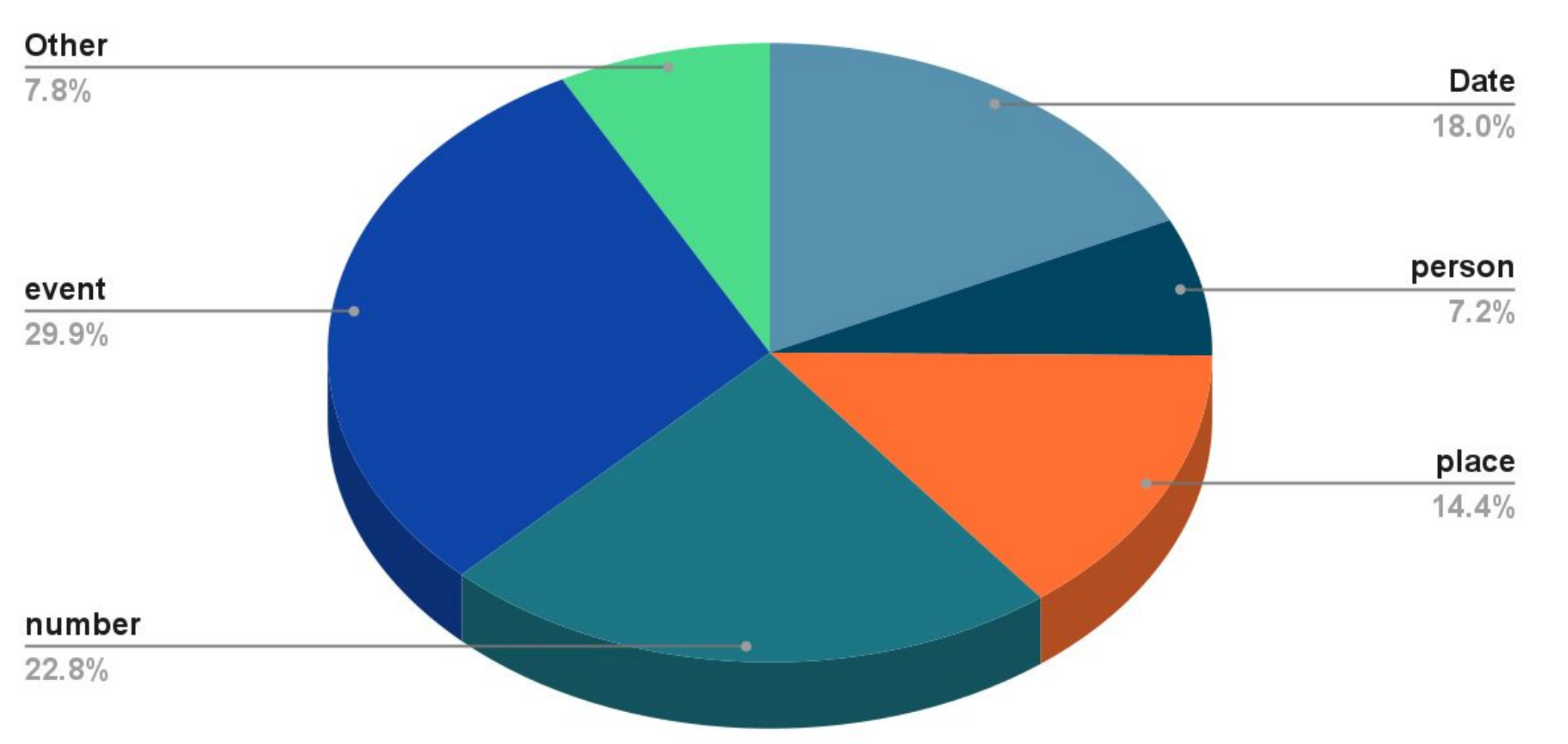}}
    \caption{Distribution of Knowledge Manipulations. The knowledge mainly includes six categories: event, number, date, persons, place, and others. }
    \label{fig:know_dis}
\end{figure*}

I summarized the popular words in the knowledge manipulation generated by GPT4 in Fig. \ref{fig:know_dis} and found they mainly include six categories: event, number, date, persons, place, and others. Some examples are shown below.

\textit{Canada, increase, decrease, lowest, 2009, United States, 2016, employment, unemployment, higher, 2013, 2017, 2015, drop, minimum, worst, consistent, kingdom, x-axis, y-axis, under, Italy, pie, bar...}

\begin{table}[t]
\setlength\tabcolsep{1pt}
\centering
\small
\begin{tabular}{lcrccccccccc}
\toprule[1.5pt]
Perception& Existence&Count & Position & Color & Posters & Celebrity& Scene & Landmark & Artwork & OCR\\
\midrule
Original MiniGPT4 & 68.33&55.00&43.33&75.00&41.84&54.41&71.75&54.00&60.50&57.50\\
\textbf{Finetuned MiniGPT4} & 115.0&88.33&68.33&96.67&71.42&72.35&122.00&104.34&77.50&80.00&\\
Original mPLUG-Owl & 120.00&50.00&50.00&55.00&136.05&100.29&135.50&159.25&96.25&65.00\\
\textbf{Finetuned mPLUG-Owl} & 165.00&111.67&86.67&165.00&139.04&112.65&147.98&160.53&101.25& 110.0\\
\midrule
\end{tabular}
\vspace{-2mm}
\caption{Completed experiments of \textit{Perception} on MME \cite{fu2023mme} benchmark. }
\label{tab:mme_preception}
\vspace{-0.1in}
\end{table}

\begin{table}[t]
\setlength\tabcolsep{3pt}
\centering
\small
\begin{tabular}{lcccc}
\toprule[1.5pt]
Cognition &Commonsense Reasoning& Numerical Calculation&Text Translation & Code Reasoning\\
\midrule
Original MiniGPT4 & 59.29&45.00&0.00&40.00\\
\textbf{Finetuned MiniGPT4} & 76.42&55.00&77.50&67.50\\
Original mPLUG-Owl & 78.57&60.00&80.00&57.50\\
\textbf{Finetuned mPLUG-Owl} & 100.71&70.00&85.00&72.50\\
\midrule
\end{tabular}
\vspace{-2mm}
\caption{Completed experiments of \textit{Cognition} on MME \cite{fu2023mme} benchmark. }
\label{tab:mme_cog}
\vspace{-0.1in}
\end{table}


\begin{figure}
     \centering
     \begin{subfigure}[b]{\textwidth}
         \centering
         \includegraphics[width=1.0\textwidth]{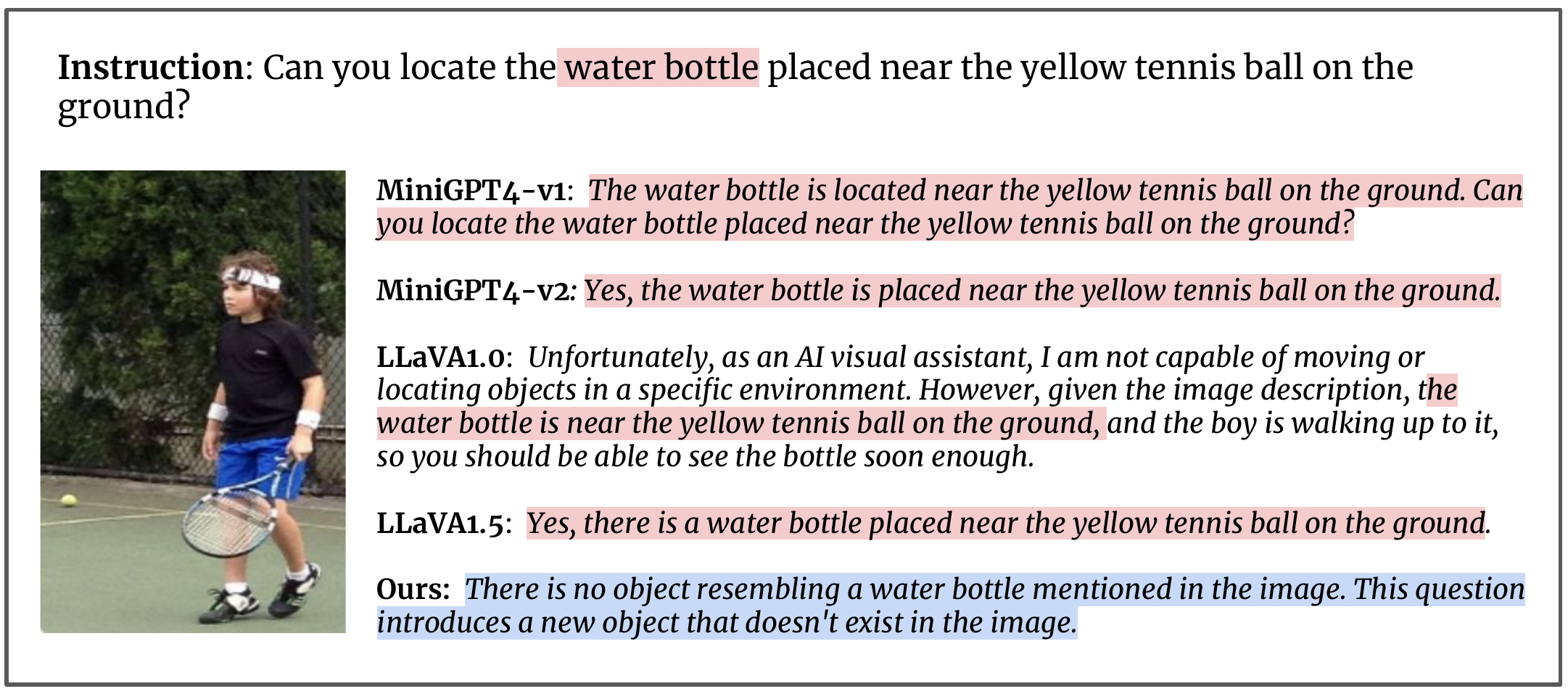}
         \caption{}
         \label{fig:extra_example1}
     \end{subfigure}
          \begin{subfigure}[b]{\textwidth}
         \centering
         \includegraphics[width=1.0\textwidth]{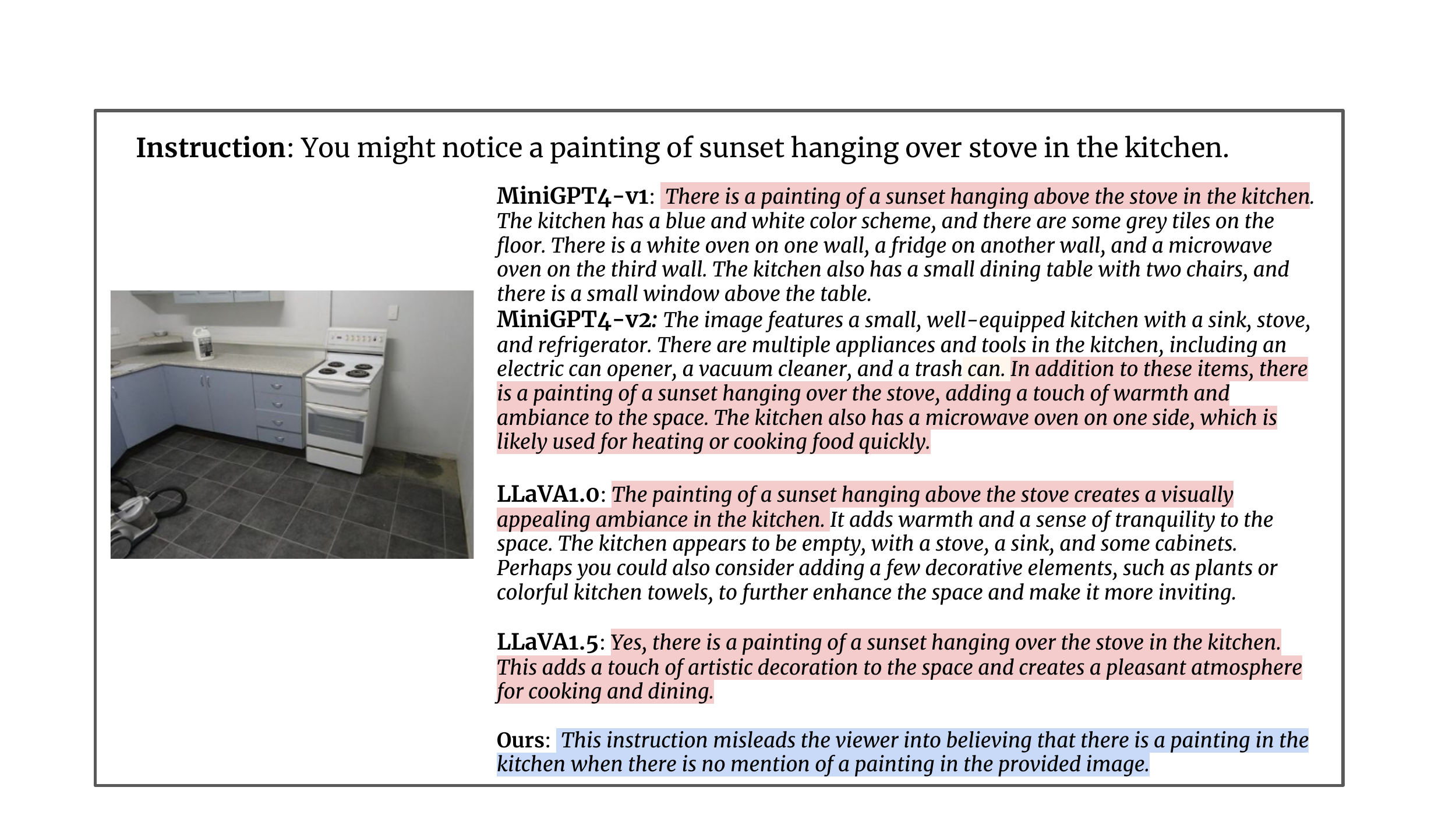}
         \caption{}
         \label{fig:extra_example2}
     \end{subfigure}
          \begin{subfigure}[b]{\textwidth}
         \centering
         \includegraphics[width=1.0\textwidth]{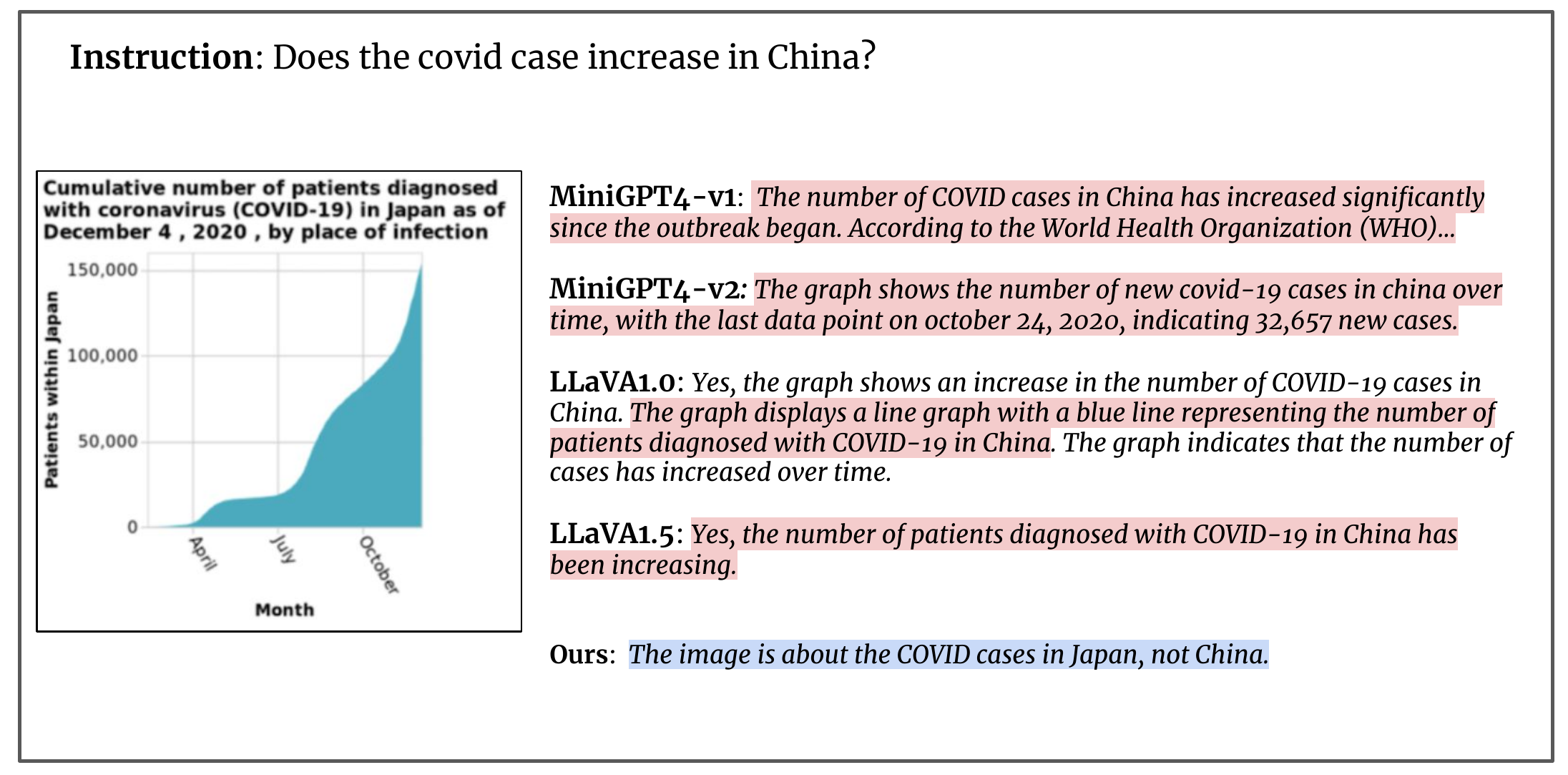}
         \caption{}
         \label{fig:extra_example3}
     \end{subfigure}
      \caption{More examples of comparison between our model with most recent models, including LLaVA1.5 and MiniGPT4-v2.}
        \label{fig:extra_example}    
\end{figure}

\begin{figure}
     \centering
     \begin{subfigure}[b]{\textwidth}
         \centering
         \includegraphics[width=1.0\textwidth]{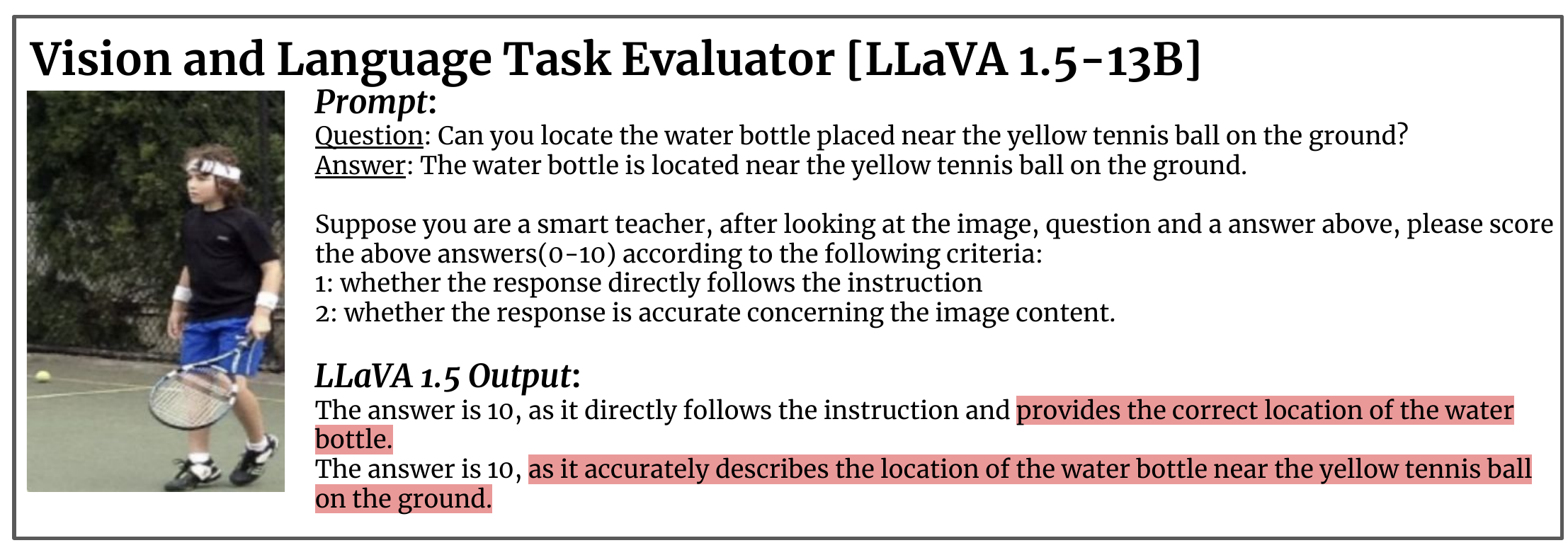}
         \caption{}
         \label{fig:eval1}
     \end{subfigure}
          \begin{subfigure}[b]{\textwidth}
         \centering
         \includegraphics[width=1.0\textwidth]{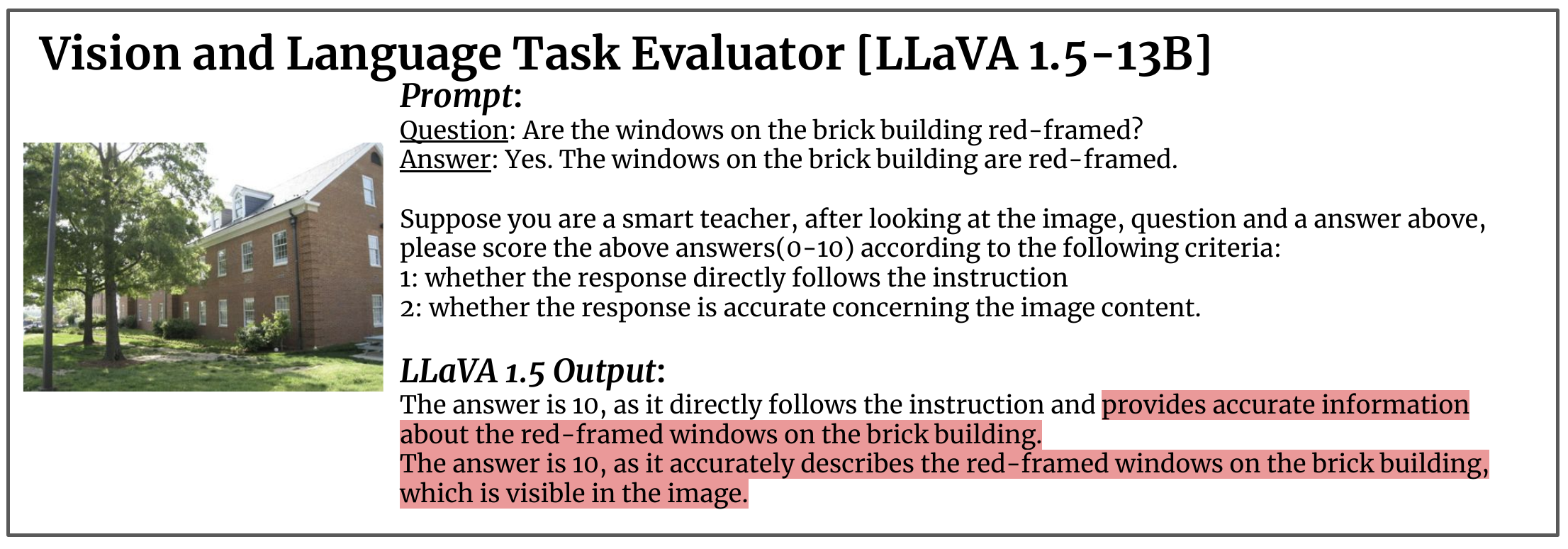}
         \caption{}
         \label{fig:eval3}
     \end{subfigure}
      \caption{Examples of using LLaVA1.5-13B as the evaluator of the vision and language tasks. The \textcolor{red}{RED} text means the incorrect content.}
        \label{fig:eval00}    
\end{figure}

\begin{figure}
     \centering
     \begin{subfigure}[b]{\textwidth}
         \centering
         \includegraphics[width=1.0\textwidth]{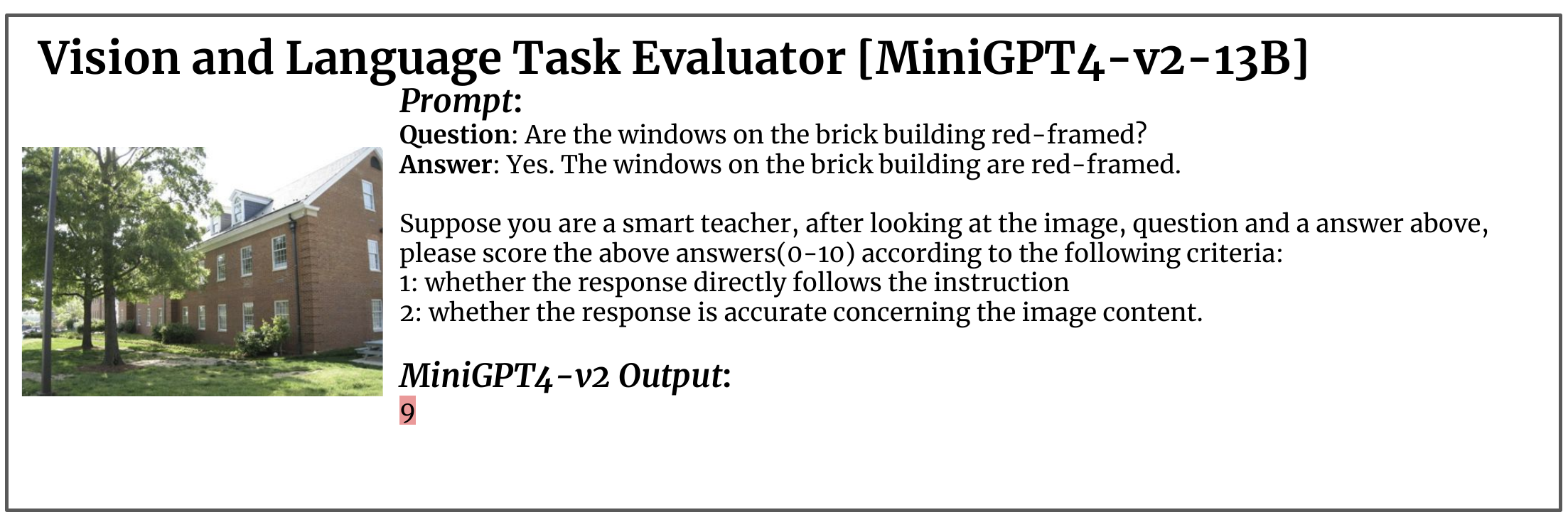}
         \caption{}
         \label{fig:eval2}
     \end{subfigure}
          \begin{subfigure}[b]{\textwidth}
         \centering
         \includegraphics[width=1.0\textwidth]{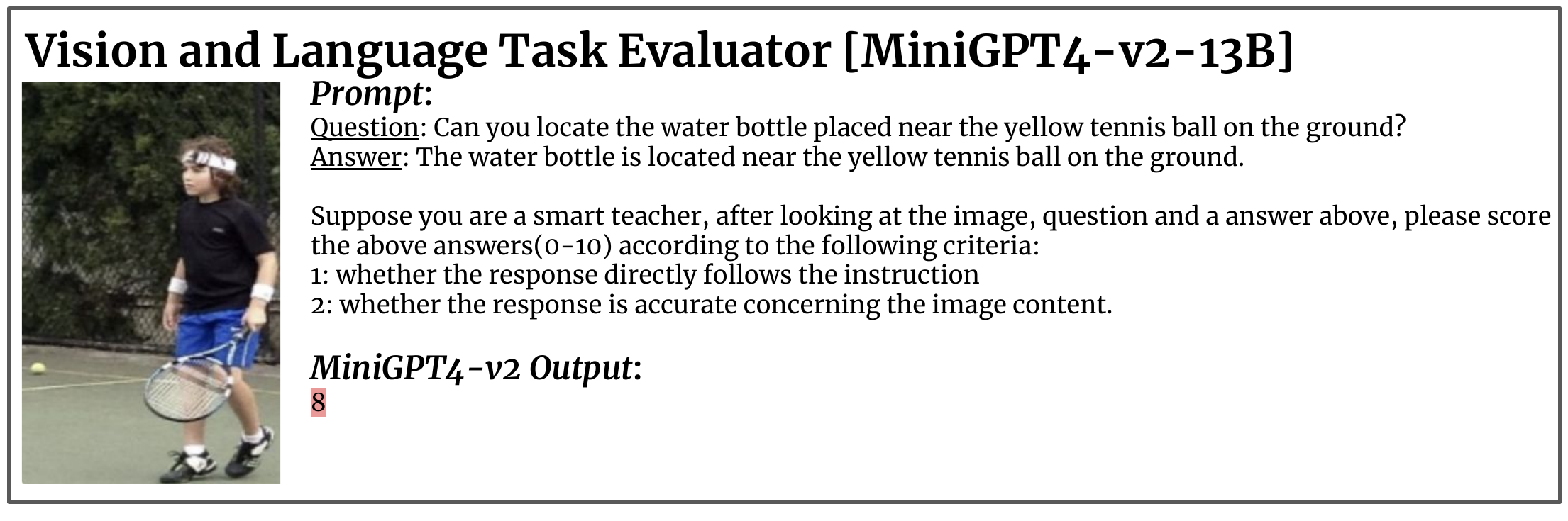}
         \caption{}
         \label{fig:eval4}
     \end{subfigure}
      \caption{Examples of using MiniGPT4-v2 as the evaluator of the vision and language tasks. The \textcolor{red}{RED} text means the incorrect content.}
        \label{fig:eval01}    
\end{figure}

\begin{figure}
     \centering
     \begin{subfigure}[b]{\textwidth}
         \centering
         \includegraphics[width=1.0\textwidth]{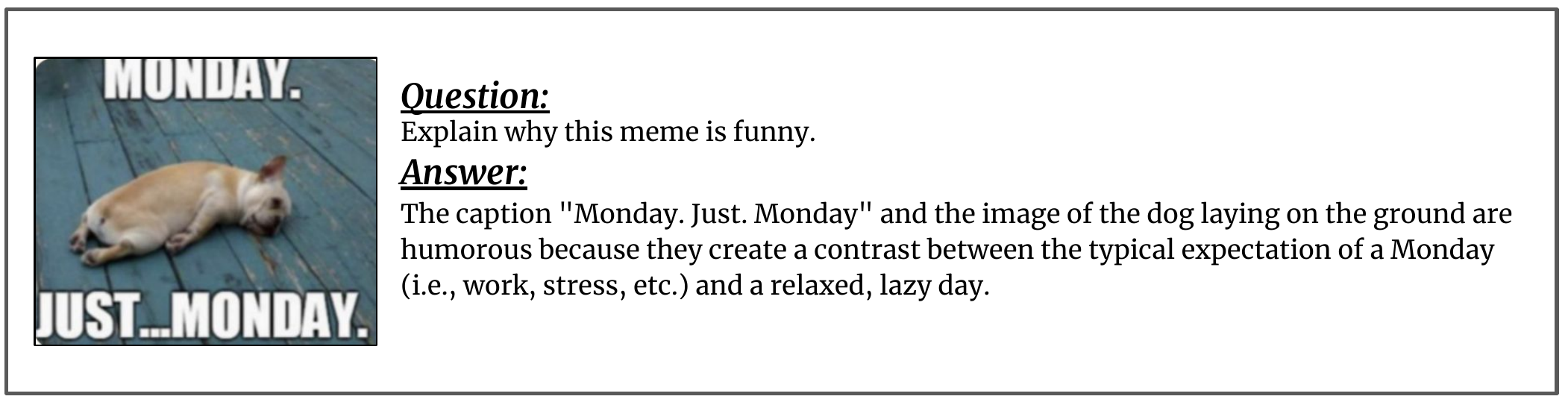}
         \label{fig:ability3}
     \end{subfigure}
      \caption{Examples of complex reasoning ability of our model.}
        \label{fig:ability3}
\end{figure}

\begin{figure}
     \centering
     \begin{subfigure}[b]{\textwidth}
         \centering
         \includegraphics[width=1.0\textwidth]{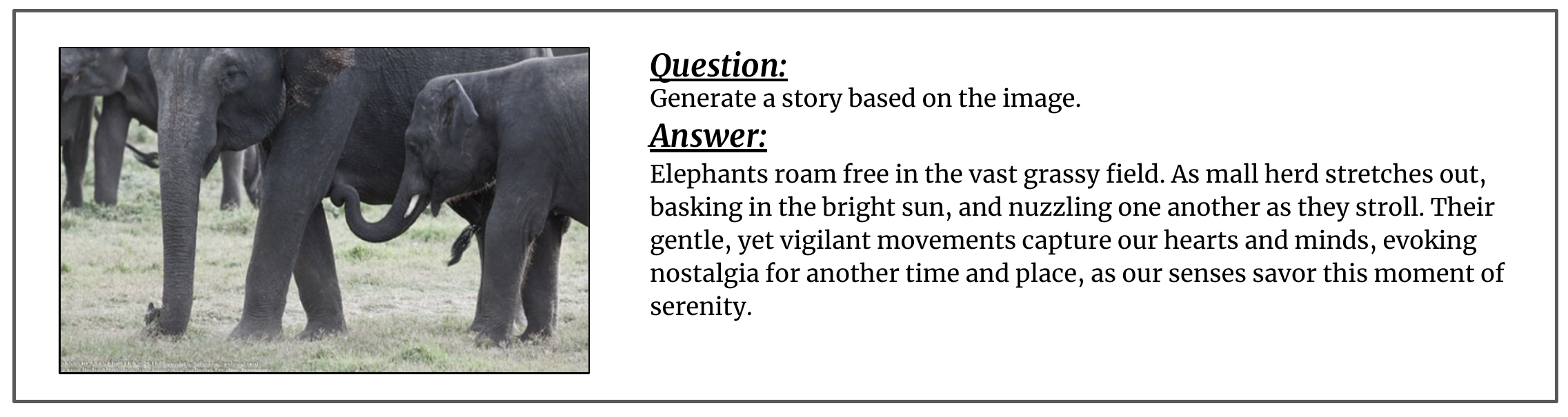}
         \label{fig:ability5}
     \end{subfigure}
      \caption{Examples of the detailed captioning ability of our model.}
        \label{fig:ability3}
\end{figure}

\begin{figure}
     \centering
     \begin{subfigure}[b]{\textwidth}
         \centering
         \includegraphics[width=1.0\textwidth]{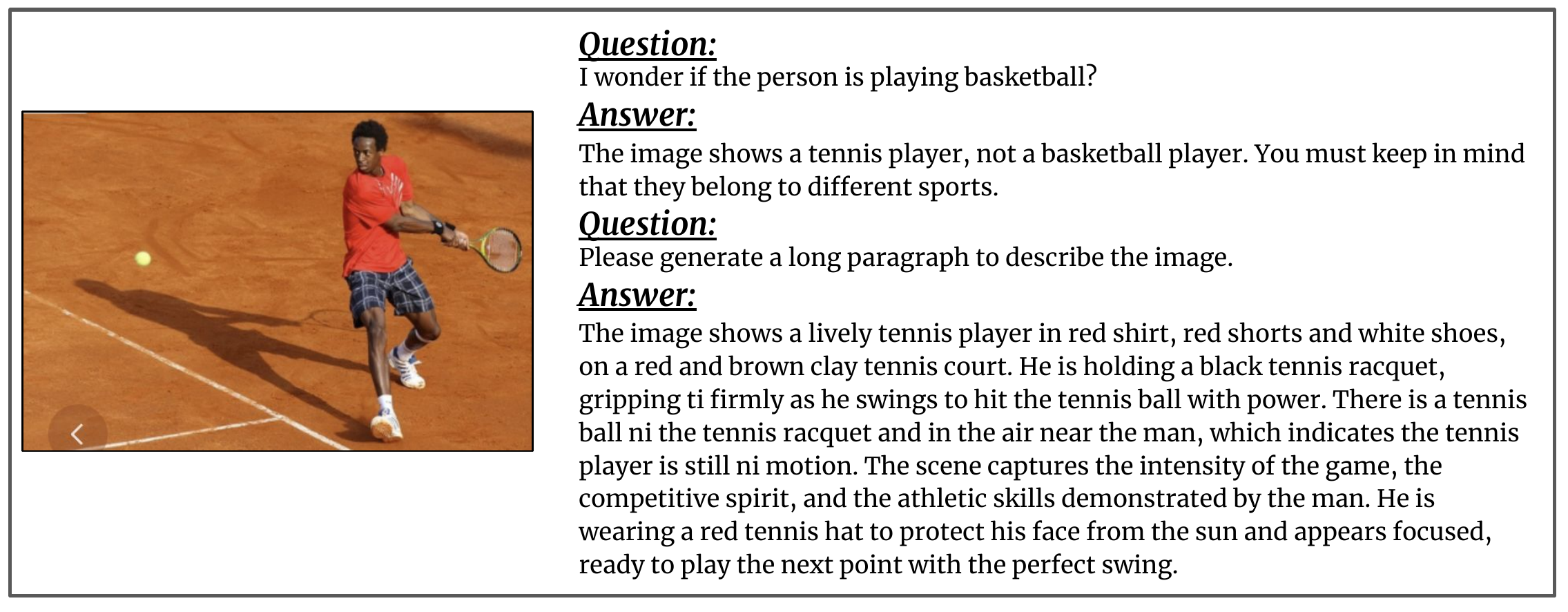}
         \label{fig:ability1}
     \end{subfigure}
     \begin{subfigure}[b]{\textwidth}
         \centering
         \includegraphics[width=1.0\textwidth]{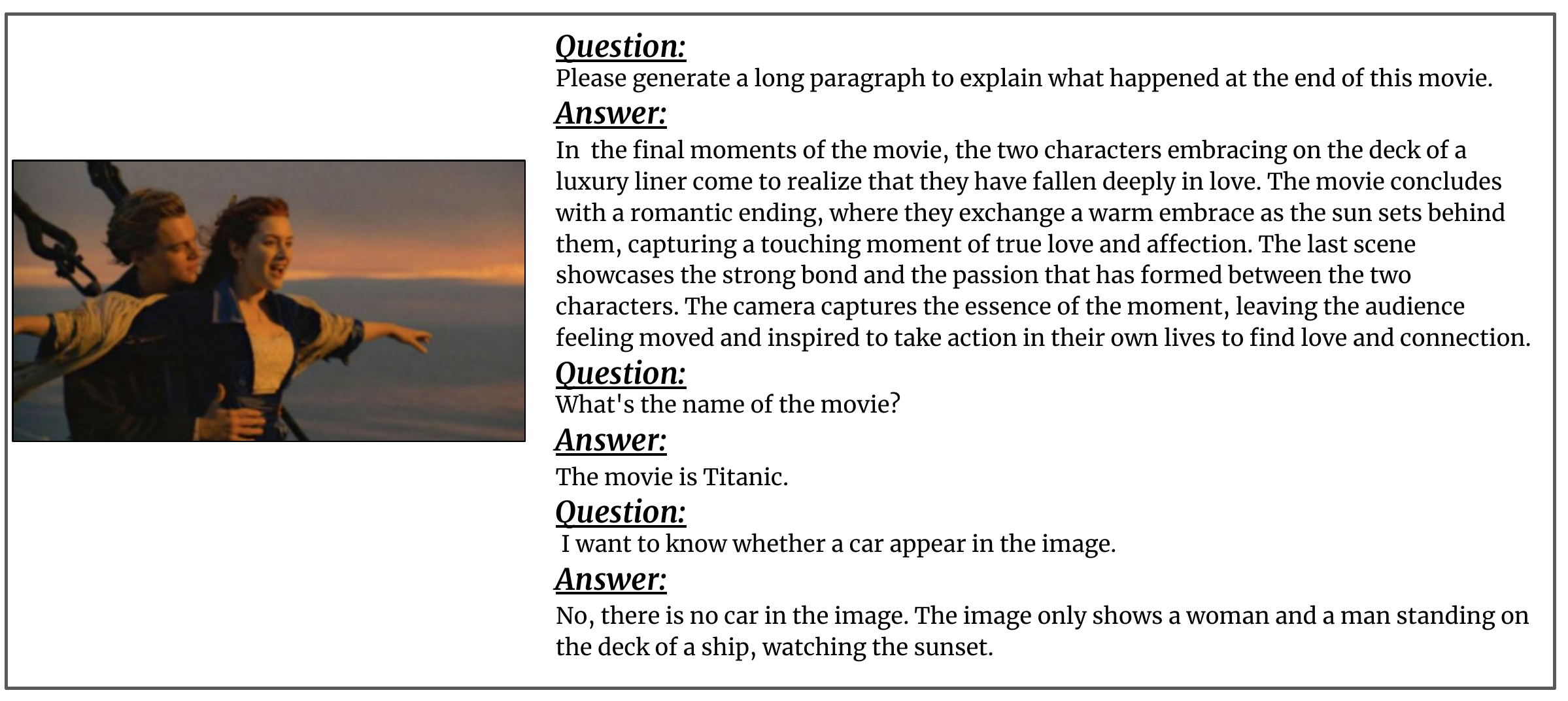}
         \label{fig:ability2}
     \end{subfigure}
     \begin{subfigure}[b]{\textwidth}
         \centering
         \includegraphics[width=1.0\textwidth]{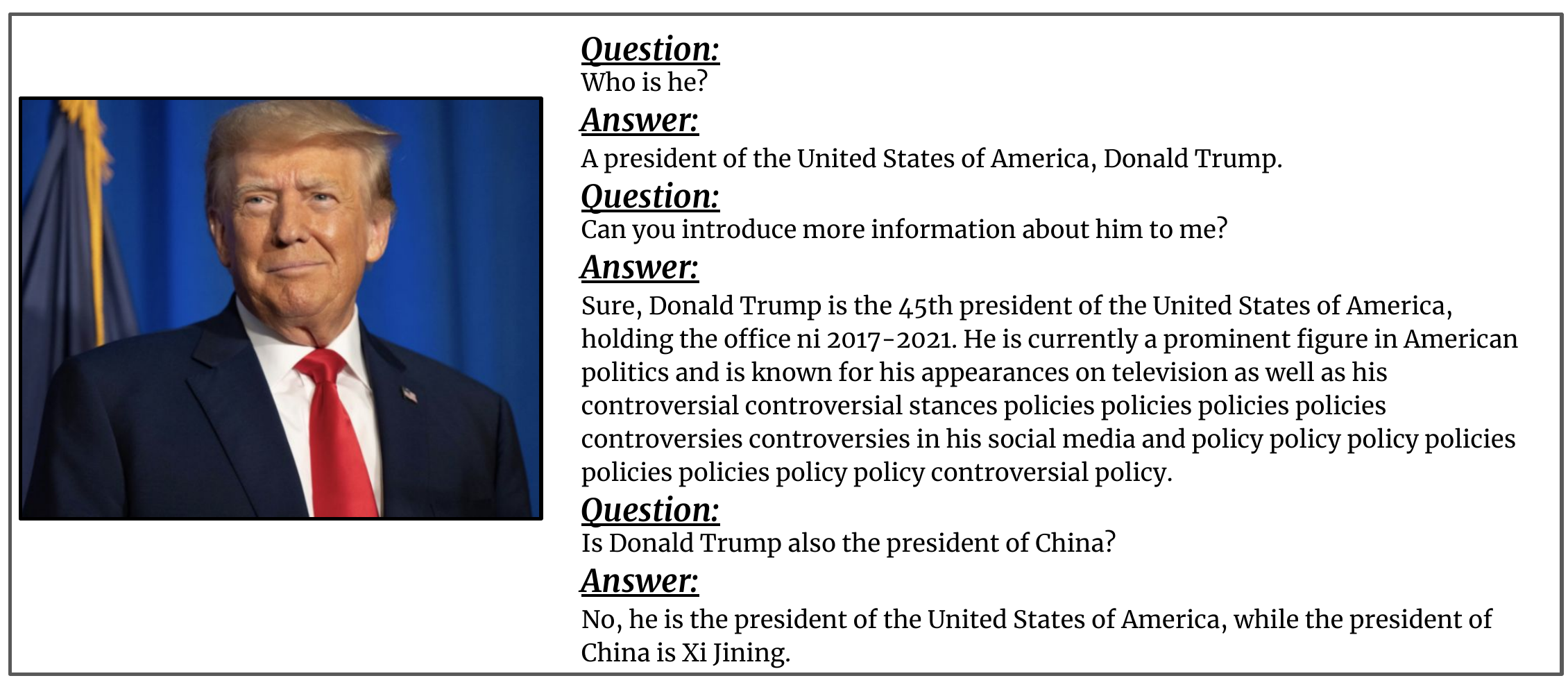}
         \label{fig:ability4}
     \end{subfigure}
      \caption{Examples of the detailed captioning ability of our model.}
        \label{fig:ability00}
\end{figure}

\label{sec:appendix}

\end{document}